\documentclass[12pt,twoside]{report}
\usepackage{pbox}
\usepackage[table, dvipsnames]{xcolor}
\usepackage{plain}
\usepackage{multicol}
\usepackage{changepage}
\usepackage{pdfpages}

\newcommand{\reporttitle}{Interpretable Deep Neural Networks for Facial Expression and Dimensional Emotion Recognition in-the-wild}
\newcommand{\reportauthor}{Valentin RICHER}
\newcommand{\supervisor}{Dimitrios KOLLIAS}
\newcommand{\degreetype}{Advanced Computing}

\usepackage[a4paper,hmargin=2.8cm,vmargin=2.0cm,includeheadfoot]{geometry}
\usepackage{textpos}
\usepackage{tabularx,longtable,multirow,subfigure,caption}
\usepackage{fncylab} 
\usepackage{fancyhdr} 
\usepackage{url} 
\usepackage[english]{babel}
\usepackage{amsmath}
\usepackage{graphicx}
\usepackage{dsfont}
\usepackage{epstopdf} 
\usepackage{backref} 
\usepackage{array}
\usepackage{latexsym}
\usepackage[pdftex,pagebackref,hypertexnames=false,colorlinks]{hyperref} 

\hypersetup{pdftitle={},
  pdfsubject={}, 
  pdfauthor={},
  pdfkeywords={}, 
  pdfstartview=FitH,
  pdfpagemode={UseOutlines},
  bookmarksnumbered=true, bookmarksopen=true, colorlinks,
    citecolor=black,%
    filecolor=black,%
    linkcolor=black,%
    urlcolor=black}

\usepackage[all]{hypcap}

\def\@makechapterhead#1{%
  \vspace*{10\p@}%
  {\parindent \z@ \raggedright \sffamily
    \interlinepenalty\@M
    \Huge\bfseries \thechapter \space\space #1\par\nobreak
    \vskip 30\p@
  }}

\def\@makeschapterhead#1{%
  \vspace*{10\p@}%
  {\parindent \z@ \raggedright
    \sffamily
    \interlinepenalty\@M
    \Huge \bfseries  #1\par\nobreak
    \vskip 30\p@
  }}

\allowdisplaybreaks

\date{September 2018}

\begin{document}

\begin{titlepage}

\newcommand{\HRule}{\rule{\linewidth}{0.5mm}} 


\includegraphics[width = 4cm]{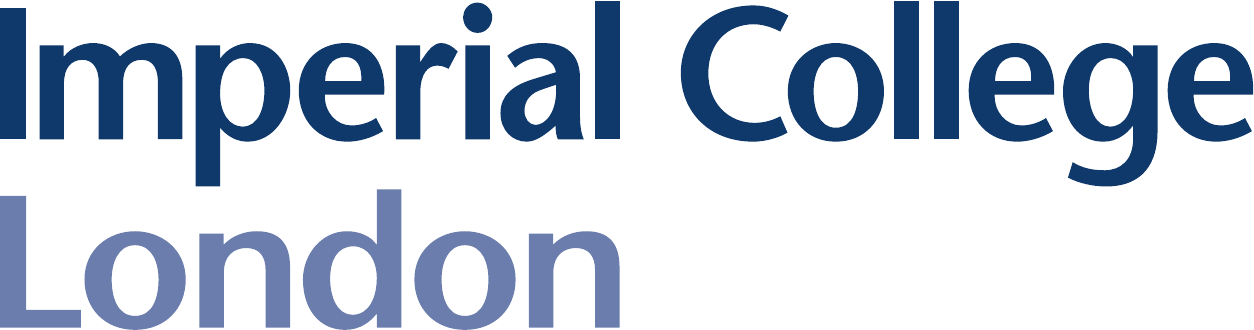}\\[0.5cm] 

\center 


\textsc{\Large Imperial College London}\\[0.5cm] 
\textsc{\large Department of Computing}\\[0.5cm] 


\HRule \\[0.4cm]
{ \huge \bfseries \reporttitle}\\ 
\HRule \\[1.0cm]


\begin{minipage}{0.4\textwidth}
\begin{flushleft} \large
\emph{Author:}\\
\reportauthor 
\end{flushleft}
\end{minipage}
~
\begin{minipage}{0.4\textwidth}
\begin{flushright} \large
\emph{Supervisor:} \\
\supervisor 
\end{flushright}
\end{minipage}\\[4cm]

\vfill 
Submitted in partial fulfillment of the requirements for the MSc degree in
\degreetype~of Imperial College London\\[0.5cm]

\makeatletter
\@date 
\makeatother

\end{titlepage}

\pagenumbering{roman}
\clearpage{\pagestyle{empty}\cleardoublepage}
\setcounter{page}{1}
\pagestyle{fancy}

\begin{abstract}
In this project, we created a database with two types of annotations used in the emotion recognition domain : Action Units and Valence Arousal to try to achieve better results than with only one model. The originality of the approach is also based on the type of architecture used to perform the prediction of the emotions : a categorical Generative Adversarial Network. This kind of dual network can generate images based on the pictures from the new dataset thanks to its generative network and decide if an image is fake or real thanks to its discriminative network as well as help to predict the annotations for Action Units and Valence Arousal due to its categorical nature. GANs were trained on the Action Units model only, then the Valence Arousal model only and then on both the Action Units model and Valence Arousal model in order to test different parameters and understand their influence. The generative and discriminative aspects of the GANs have performed interesting results.
\end{abstract}

\cleardoublepage
\section*{Acknowledgments}
I would like to sincerely thank my supervisor, Dimitrios Kollias, who was always ready to guide and help me throughout this project. His ideas were always inspiring and practical. During the numerous discussions we had, he was very helpful, willing to answer my questions and encourage me. Without him, I could never have completed this project successfully. 

\clearpage{\pagestyle{empty}\cleardoublepage}

\fancyhead[RE,LO]{\sffamily {Table of Contents}}
\tableofcontents

\clearpage{\pagestyle{empty}\cleardoublepage}
\pagenumbering{arabic}
\setcounter{page}{1}
\fancyhead[LE,RO]{\slshape \rightmark}
\fancyhead[LO,RE]{\slshape \leftmark}


\chapter{Introduction}

\begin{figure}[!ht]
    \centering
    \includegraphics[scale=0.3]{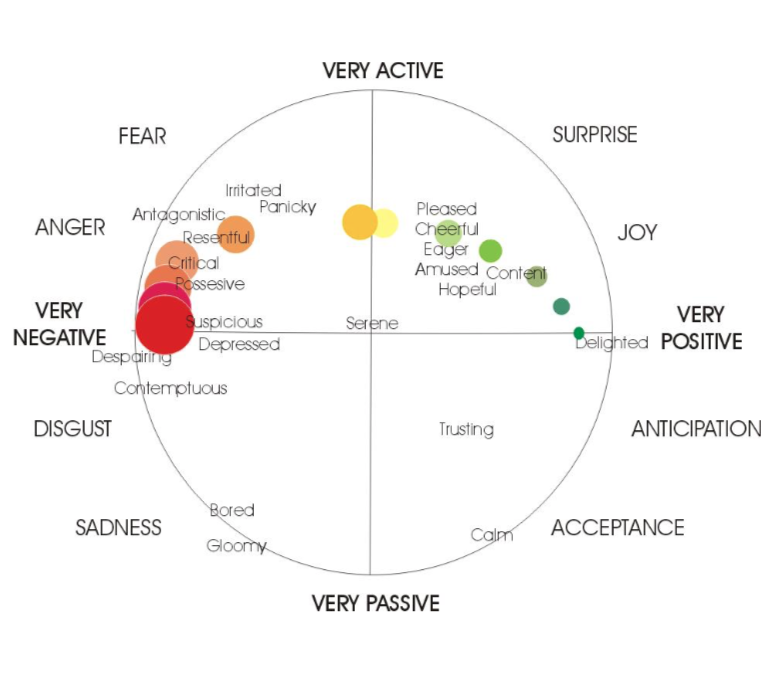}
    \caption{The 2-D Emotion Wheel \cite{kollias1}}
    \label{fig:emotion_wheel}
\end{figure}

Emotion recognition is the subject of numerous studies because of the various applications \cite{goudelis2013exploring,mylonas2009using} but also due to the complexity and diversity of human faces. Facial emotions have long been described by the six basic emotions (anger, disgust, fear, happiness, sadness and surprise).
\newline
In this paper, we focus on two different ways of describing human emotions. Valence and Arousal is one of them. It is a continuous 2D scale where Valence represents how much the subject is feeling a positive or negative emotion and Arousal describes how active or passive the subject is. The Valence and Arousal description is more subtle than the six basic emotions used to describe how a person feels, as the Figure \ref{fig:emotion_wheel} shows. 
\newline
On the other hand, Action Units are the fundamental movements of muscles on a human face from the combination of which emotions can be assumed. The Facial Action Coding System, gathering all the Action Units, has first been described by Carl-Herman Hjortsj\"o. Then, Ekman and Friesen have adopted the system in 1978 \cite{RefWorks:doc:5b0ec7f2e4b02f452d8e78c0} and updated it in 2002 \cite{RefWorks:doc:5b0ec87be4b0ca9da8b4fa87}. Action Units can be described with 5 levels of intensity or just with the presence or the absence (as used in this paper).
\\
This project aims at assembling Facial Action Units and Valence Arousal models so as to perform better results than those achieved before. The approach chosen is a new one in emotion recognition : a categorical Generative Adversarial Network will be trained to predict the values of these two models and generate images of faces. \\
First, a new dataset was created to assemble the two types of annotations (Action Units and Valence Arousal). This dataset was created by selecting videos from the Aff-Wild dataset (described in the following section \ref{sec:database}) which gathers YouTube videos of people reacting, i.e. videos in-the-wild. This dataset was already annotated for Valence Arousal and the main part was to annotate it for the 8 Action Units chosen : Action Unit 1 (Inner brow raiser), Action Unit 2 (Outer brow raiser), Action Unit 4 (Brow lowerer), Action Unit 6 (Cheek raiser), Action Unit 12 (Lip corner puller), Action Unit 15 (Lip corner depressor), Action Unit 20 (Lip stretcher) and Action Unit 25 (Lips part). \\
After the creation of the dataset, three different models of GANs were trained on this new dataset : the first dedicated to Action Units only, the second one focused on Valence Arousal and the third one assembling the Action Units and the Valence Arousal models. This enabled us to achieve interesting performances when the GAN was dedicated to one model and more disappointing ones when the two models were gathered.


\chapter{Related work}

\section{Presentation}

Scientific work on emotion recognition and facial expression has started more than 30 years ago \cite{simou2008image,simou2007fire}. At the beginning, the studies were focused on computer vision techniques as shown in these two surveys \cite{RefWorks:doc:5af42217e4b0cebb7adfb215, RefWorks:doc:5af421b6e4b02dfcb38d52a0}. Moreover, the databases evolved from controlled environments to completely free ones, i.e. in-the-wild as shown in the following Tables \ref{tab:datasetsVA} and \ref{tab:datasetsAU}. 
\newline
In this paper, the focus will be more on deep learning techniques like end-to-end neural networks with minimal computer vision methods added trained on a in-the-wild database. Indeed, the best results in facial expression recognition is now reached with deep neural networks \cite{RefWorks:doc:5af42248e4b0f7b951da62fe}; deep neural networks also achieve state-of-the-art results for other fields such as medical \cite{kollias13,tagaris1,tagaris2}
, robotics, marketing.

\section{Databases} \label{sec:database}

\subsection{Databases for Valence Arousal} \label{sec:va_database}

For Valence and Arousal estimation, the main databases are :
\begin{itemize}
    \item The SEMAINE (Sustained Emotionally coloured Machine-human Interaction using Nonverbal Expression) dataset. It is audiovisual interactions between a human and a machine \cite{kollias10}, using the Sensitive Artificial Listener method (SAL) \cite{RefWorks:doc:5b07ebbce4b0f7c767179557}. SAL is a technique of exchange between a human and a machine based on the principle that the machine can have a superficial approach of the meaning of the conversation \cite{RefWorks:doc:5b0be2a4e4b02f452d8e2352}. The dataset is annotated for 5 dimensions : Valence, Activation, Power, Anticipation/Expectation and Intensity. Other labels have also been optionnally added like FACS (Facial Action Coding System). 
    \item The RECOLA (REmote COLlaborative and Affective) database that contains audio, visual and psychological (electrocardiogram and electrodermal activity) reactions of 2 persons performing a collaborative task amongst 46 French-speaking participants (34 gave their consent to give the data outside the consortium) \cite{RefWorks:doc:5b07f529e4b0407f82f771b2}.
    \item The AVEC datasets that use the Solid-SAL partition of the SEMAINE dataset (140 operator-user interaction). The AVEC challenges are divided into three different challenges : Audio, Visual and AudioVisual. These challenges also contain sub-challenges like Depression Analysis \cite{RefWorks:doc:5b0c04c9e4b08a9fe9335675, RefWorks:doc:5b0811f6e4b009b94b6ca272}.
    \item Aff-Wild (Affect in-the-wild) Valence and Arousal database which is a set of 298 YouTube videos capturing people reactions filmed by their webcams in an uncontrolled environment, i.e. in-the-wild. This database has been annotated by 6-8 experts \cite{kollias1,kollias2,kollias3}.
    \item AFEW-VA that brings together 600 videos. In addition to the Valence and Arousal annotations, 68 facial landmarks were added \cite{RefWorks:doc:5b0c082ee4b0562e289ce43d}.   
\end{itemize}

\begin{table}[!ht]
     
        \centering
        \begin{adjustwidth}{-0.15\textwidth}{-0.0\textwidth}
        \renewcommand{\arraystretch}{3}

    \resizebox{570pt}{!}{\begin{tabular}{|c|c|c|c|c|c|c|}
        \hline
        \rowcolor{gray!50}
        {Database} & {Subjects Demography} & {Annotation Type} & Amount of data & Elicitation method & Environment & Launch Date \\
        \hline
        \rowcolor{gray!20}
        SEMAINE \cite{RefWorks:doc:5b07ebbce4b0f7c767179557} & 150 & continuous traces & \pbox{5cm}{959 conversations \\ lasting 5 minutes each} & SAL & controlled & 2007  \\
        RECOLA \cite{RefWorks:doc:5b07f529e4b0407f82f771b2} & 46 French-speaking & time and value continuous & 9.5 hours & online dyadic interactions & controlled & 2013 \\
        \rowcolor{gray!20}
        AVEC 2013 \cite{RefWorks:doc:5b0c04c9e4b08a9fe9335675} & 292 & continuous & \pbox{5cm}{340 video clips of 25 minutes on average \\ total duration of 240 hours} & human-computer interactions & in-the-wild & 2013 \\
        AVEC 2014 \cite{RefWorks:doc:5b0811f6e4b009b94b6ca272} & 84 German-speaking & continuous & 300 videos from 6s to 3 min 50 s & human-computer interactions & in-the-wild & 2014 \\  
        \rowcolor{gray!20}
        Aff-Wild VA \cite{kollias4,kollias5,kollias15} & \pbox{5cm}{130 males \\ 70 females} & frame-by-frame & \pbox{5cm}{298 YouTube videos \\ +30 hours \\ 1,180,000 frames} & videos & in-the-wild & 2016 \\
        AFEW-VA \cite{RefWorks:doc:5b0c082ee4b0562e289ce43d} & 240 & frame-by-frame & \pbox{5cm}{600 videos from 10 frames to 120 frames \\ 30,000 frames in total} & movie actors & in-the-wild & 2016 \\
        \hline
    \end{tabular}}
    \end{adjustwidth}
    \caption{Datasets for Valence and Arousal estimation}
    \label{tab:datasetsVA}
\end{table}

\subsection{Databases for Action Units}

For Facial Action Units, the datasets are :
\begin{itemize}
    \item The MMI dataset which consists of 2900 videos and images of 75 subjects in controlled conditions (frontal and profile conditions). It is an incremental database with 5 parts now \cite{RefWorks:doc:5b0c0ca2e4b0542f750169d4, RefWorks:doc:5b081f42e4b085f70ca5255f}.
    \item The CK+ database (Extended Cohn-Kanade) which is the extension of the CK database including 486 FACS-coded sequences from 97 subjects (launched in 2000). The CK+ augments the previous dataset to reach 593 sequences from 123 subjects (for posed expression) \cite{RefWorks:doc:5b07d52ce4b0262a90ba52d7}.
    \item The GEMEP (GEneva Multimodal Emotion Portrayals) that collects audio and video of 10 actors playing 18 affective states \cite{RefWorks:doc:5b082aaae4b0f72b00d4ca73}. The  difference with CK+ and MMI is that  in GEMEP the actors were  allowed to act  freely \cite{RefWorks:doc:5af1aedee4b04597bddec2c7}.
    \item The DISFA (Denver Intensity of Spontaneous Facial Action) database. It consists of 4845 frames of 27 subjects (12 females and 15 males). The reactions are spontaneous to the viewing of a 4-minute long video in a controlled environment. 12 AUs are recorded in intensity frame-by-frame \cite{RefWorks:doc:5b08331be4b0353e399a5f01}. 
    \item The RU-FACS-1 corpus. It is a collection of spontaneous facial expressions from multiple views. The FACS codes are provided by two experts. The recordings are done in a controlled environment without the subject being aware of the cameras \cite{RefWorks:doc:5b0c139ae4b05e315bed2039}.
    \item The UNBC-McMaster that contains 200 videos of spontaneous facial expressions which represents 48,398 FACS coded frames. (Moreover, this dataset contains the frame-by-frame associated score and 66 facial landmarks) \cite{RefWorks:doc:5b083834e4b0357468f0b292}.
    \item The AM-FED (Affectiva-MIT Facial Expression Dataset) which brings together 242 facial videos (168,359 frames) recorded in-the-wild : the subjects were watching one of the three intentionally amusing Superbowl commercials. 14 AUs are labelled frame-by-frame. (This dataset also contains other labels like facial landmarks) \cite{RefWorks:doc:5b0c15b9e4b08a9fe9335907}.
    \item Aff-Wild FAU database consists of 10,000+ facial images of more than 2000 people in-the-wild. 16 AUs have been annotated \cite{RefWorks:doc:5af1aedee4b04597bddec2c7}.
\end{itemize}

\begin{table}[!ht]

    \centering
    \begin{adjustwidth}{-0.15\textwidth}{-0.0\textwidth}
    \renewcommand{\arraystretch}{2.7}

    \resizebox{570pt}{!}{\begin{tabular}{|c|c|c|c|c|c|c|c|}
        \hline
        \rowcolor{gray!50}
        {Database} & {Subjects Demography} & {Annotation Type} & Amount of data & Number of AUs & Elicitation method & Environment & Launch Date \\
        \hline
        \rowcolor{gray!20}
        MMI \cite{RefWorks:doc:5b081f42e4b085f70ca5255f} & 75 & event-coded and partially frame-coded & 2,900 videos & 31 & video reactions & controlled & 2002 \\
        CK+ \cite{RefWorks:doc:5b07d52ce4b0262a90ba52d7} & 123 & action units coded and frame-by-frame coded & 593 sequences & 30 & videos & controlled (posed and spontaneous) & 2010 \\
        \rowcolor{gray!20}
        GMEP \cite{RefWorks:doc:5b082aaae4b0f72b00d4ca73} & 10 & continuous & 1,260 portrayals & 40 & acting & controlled & 2007 \\
        DISFA \cite{RefWorks:doc:5b08331be4b0353e399a5f01} & 27 & frame-by-frame & \pbox{5cm}{130,000 video frames \\ 4-minute long videos} & 12 & 4-minute video viewing & controlled (spontaneous) & 2013 \\
        \rowcolor{gray!20}
        RU-FACS \cite{RefWorks:doc:5b0c139ae4b05e315bed2039} & 100 & action unit coded & 2.5 minutes for each subject & 20 & truth or lie situation & controlled (spontaneous) & 2004 \\
        UNBC-McMaster \cite{RefWorks:doc:5b083834e4b0357468f0b292} & 129 & frame-by-frame & 48,398 FACS coded frames & 44 & pain & controlled & 2011 \\
        \rowcolor{gray!20}
        AM-FED \cite{RefWorks:doc:5b0c15b9e4b08a9fe9335907} & 242 & frame-by-frame & 168,359 frames & 14 & watching a commercial & in-the-wild & 2013 \\
        Aff-Wild \cite{RefWorks:doc:5af1aedee4b04597bddec2c7} & 2000 & frame-by-frame & 10,000+ facial images & 16 & videos & in-the-wild & 2016 \\
        \hline
    \end{tabular}}
    \end{adjustwidth}
    \caption{Datasets for Action Units estimation}
    \label{tab:datasetsAU}
\end{table}

\subsection{Database to be used for the project}

The database that will be used for this project is a subset of the Aff-Wild database on which 8 Action Units annotations have been added. These Action Units are :

\begin{table}[!ht]
    \tiny
    \centering
    \renewcommand{\arraystretch}{0.1}
    \resizebox{\textwidth}{!}{\begin{tabular}{|c|c|c|}
        \hline
        \rowcolor{gray!50}
        {Action Unit} & {Description} & {Example}  \\
        \hline
           1 & Inner brow raiser & \begin{minipage}{0.15\textwidth}
            \includegraphics[width=\linewidth]{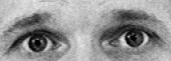}
            \end{minipage} \\
            \hline
        2 & Outer brow raiser & 
        \begin{minipage}{0.15\textwidth}
            \includegraphics[width=\linewidth]{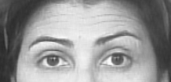}
            \end{minipage} \\
        \hline
        4 & Brow lowerer & 
        \begin{minipage}{0.15\textwidth}
            \includegraphics[width=\linewidth]{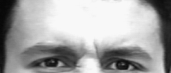}
        \end{minipage} \\
        \hline
        6 & Cheek raiser & 
        \begin{minipage}{0.15\textwidth}
            \includegraphics[width=\linewidth]{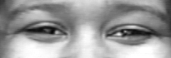}
        \end{minipage} \\
        \hline
        12 & Lip corner puller & 
        \begin{minipage}{0.15\textwidth}
            \includegraphics[width=\linewidth]{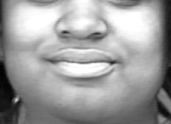}
        \end{minipage} \\
        \hline
        15 & Lip corner depressor & 
        \begin{minipage}{0.15\textwidth}
            \includegraphics[width=\linewidth]{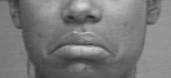}
        \end{minipage} \\
        \hline
        20 & Lip stretcher & 
        \begin{minipage}{0.15\textwidth}
            \includegraphics[width=\linewidth]{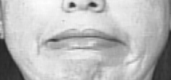}
        \end{minipage} \\
        \hline
        25 & Lips part & 
        \begin{minipage}{0.15\textwidth}
            \includegraphics[width=\linewidth]{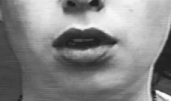}
        \end{minipage} \\
        \hline
    \end{tabular}}
    \caption{Action Units labelled on Aff-Wild}
    \label{tab:AU}
\end{table}

\newpage

\section{How to improve the database}

The database needs to be improved in terms of quantity and quality. Indeed, the number of videos is not optimal as deep neural networks often need a lot of data to learn and generalize well without overfitting. Moreover, there is only one annotator so the possible mistakes cannot be balanced by other annotators with a mean of annotations for example. Generative Adversarial Networks (GANs) seem to be suitable to improve the dataset. GAN is a type of deep learning technique in which two neural networks compete : one is the generative neural network which tries to generate data that can be mistaken as genuine while the other one is the discriminative network that tries to distinguish the real data from the data produced by the generative neural network. It is an iterative process in which the improvement of one network brings about the improvement of the other \cite{RefWorks:doc:5af42188e4b07aa41af12318}.
\newline
\newline
The dataset to be used may suffer from a lack of annotations for Action Units. The idea would be to use a GAN to label the AUs on the unlabelled videos as it is a time-consuming task. The paper \cite{RefWorks:doc:5af1a0d1e4b0155360c56aa1} uses GANs with k classes to augment the dataset. In this categorical GAN the discriminator assigns all examples to one of k categories while staying uncertain of class assignments for samples from the generative model k. Moreover, instead of generating samples from the dataset the generator creates samples from one of the k categories. 
\newline
Another method would be to use unsupervised learning to create clusters of Action Units as explained in \cite{RefWorks:doc:5af1a199e4b0b5a912ed9e0a}.

\section{Deep Learning architectures}

Deep Learning is improving the results on emotion recognition \cite{kollias6,kollias11,kollias12} compared to previous computer vision techniques. This is the reason why we focus on these promising techniques in the background. 

\subsection{Different types of neural networks}\label{sec:nn_types}

A \textbf{feedforward neural network} is the simplest type of neural network. Every node is connected to all the previous nodes by a linear combination. An activation function $\sigma$ can be applied to the node. There are many different types but the most common are the sigmoid function and the ReLU (Rectified Linear Unit).
\begin{align}
    z_i^l = \sum_{k=1}^{n} w_k^l a_k^{l-1} + b_i^l \\
    a_i^l = \sigma(z_i^l)
\end{align}
where $z_i^l$ is the value of the node $i$ on layer $l$, $n$ the number of nodes in the previous layer $l-1$, $a_k^{l-1}$ is the activation value of the node $k$ of the previous layer $l-1$ and $b_i^l$ is the bias for the node $i$ in layer $l$. \newline
The first layer is called the input layer and the last layer is the output layer. The layers between these two layers are called the hidden layers. This type of network is feedforward because the information is propagated from the input layer to the output layer. 
\begin{figure}[!ht]
\centering 
\includegraphics[width = 0.7\hsize]{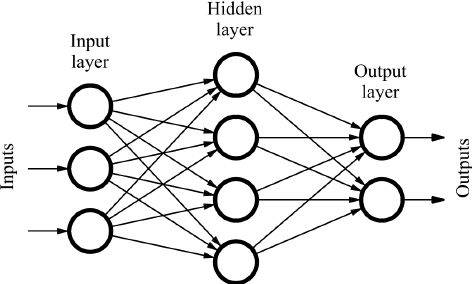} 
\caption{Graph of a Feedforward Neural Network \cite{RefWorks:doc:5b116947e4b006165bc77480}} 
\label{fig:RNN} 
\end{figure}

A \textbf{Convolutional Neural Network} (CNN) is a type of neural network stacking non-linear layers to extract features from an input. In our case the input is a 2D image. The layers can be either filters doing convolutions over the input or pooling layers whose purpose is to reduce the dimensions of the output. Filters are biased matrices that are applied on an input and produce an output on which an activation function is applied given an activation or feature map. The pooling layer chosen is often max-pooling which selects the maximum input as an output. Fully-connected layers are often stacked on top of a CNN and the final layer is the result features from which the loss or cost function is computed and then the backpropagation is fired, then modifying the weights of the filters.
\begin{figure}[!ht]
\centering 
\includegraphics[width = 1\hsize]{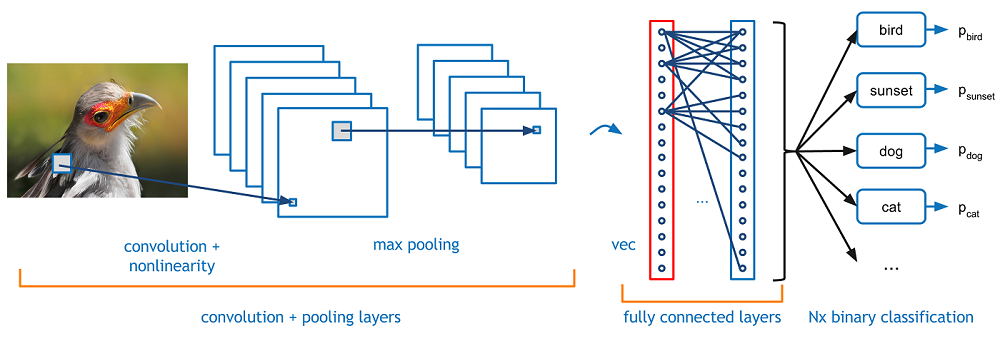} 
\caption{Graph of a CNN \cite{RefWorks:doc:5b154169e4b03a529bc77e3c}} 
\label{fig:RNN} 
\end{figure}
\clearpage

\begin{figure}[!ht]
\begin{tabular}{cccccc}
    \includegraphics[width=2.1cm]{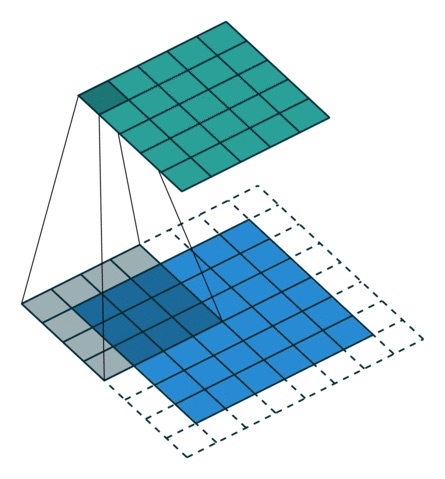} &  \includegraphics[width=2.1cm]{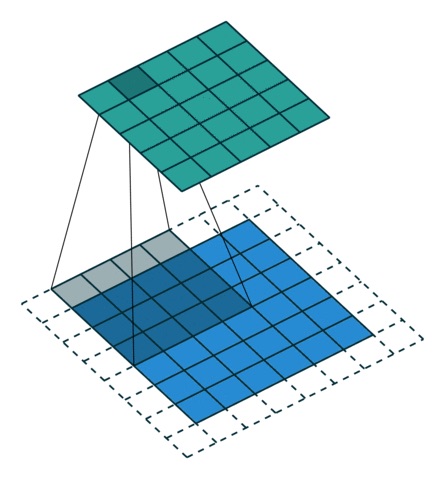} &
    \includegraphics[width=2.1cm]{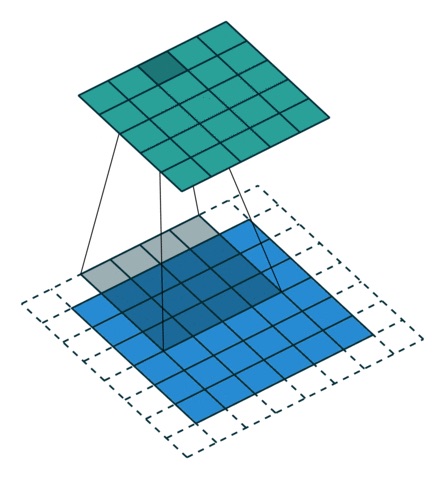} &
    \includegraphics[width=2.1cm]{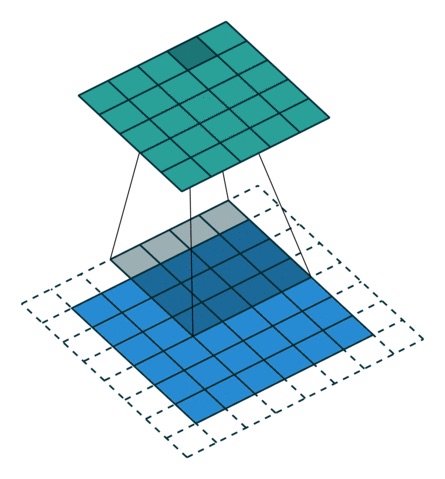} &
    \includegraphics[width=2.1cm]{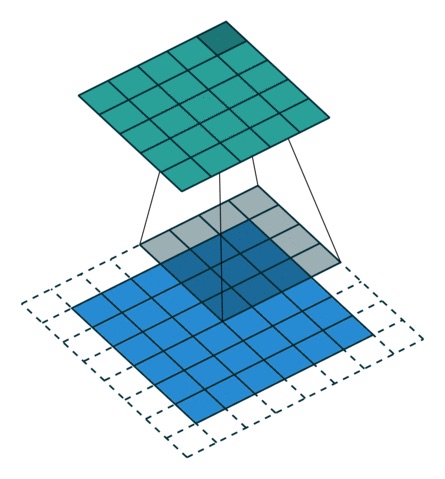} &   \includegraphics[width=2.1cm]{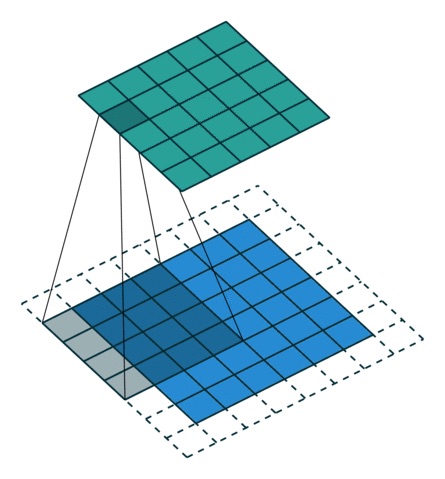} \\
    \footnotesize (a) & \footnotesize (b) & \footnotesize (c) & \footnotesize (d) & \footnotesize (e) & \footnotesize (f) \\
\end{tabular}
\caption{A convolutional layer in action from \cite{RefWorks:doc:5b8c0195e4b07993ca600a55}}
\label{fig:conv_action}
\end{figure}

On the Figure \ref{fig:conv_action}, we can see how a convolutional layer works. The filter in grey scans the image in blue. Each box in the grey matrix and blue matrix contains a number called weight. During one iteration (a, b, c, d, e or f) a multiplication between the weights of the filter and the image is done and summed. This gives the first box \cite{horrocks2011answering,glimm2013using} of the output in green. Then the filter slides through the image with a given horizontal stride like from image (a) to image (b) (horizontal stride is 1) and the operation is repeated. The vertical stride of 1 can be seen between image (e) and (f). Once the filter has scanned the entire image, the output is ready. We can see that a padding of 1 has been added around the original image. 

\begin{figure}[!ht]
\begin{tabular}{cccccc}
    \includegraphics[width=2.1cm]{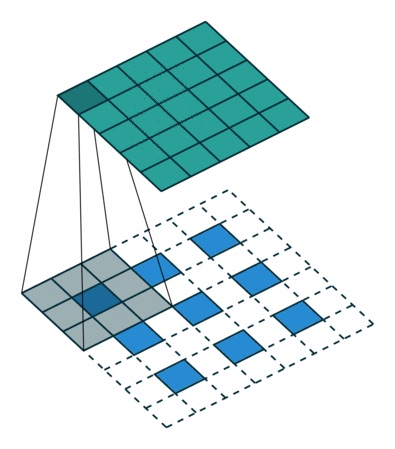} &  \includegraphics[width=2.1cm]{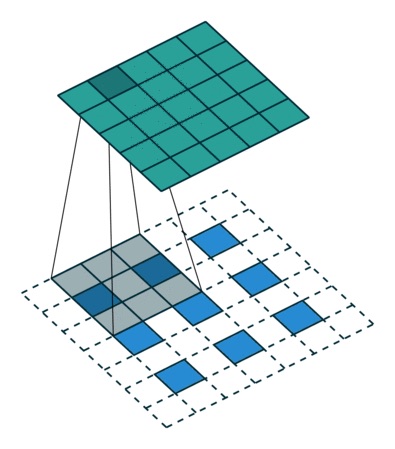} &
    \includegraphics[width=2.1cm]{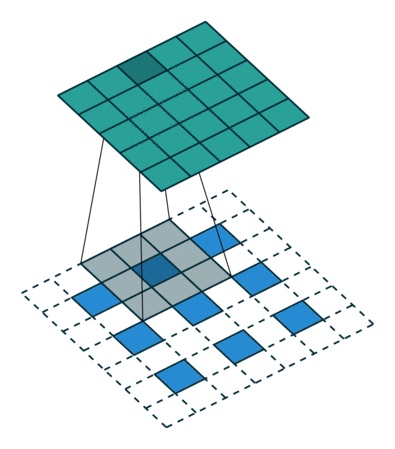} &
    \includegraphics[width=2.1cm]{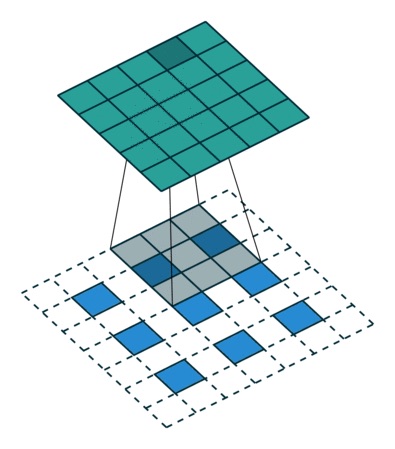} &
    \includegraphics[width=2.1cm]{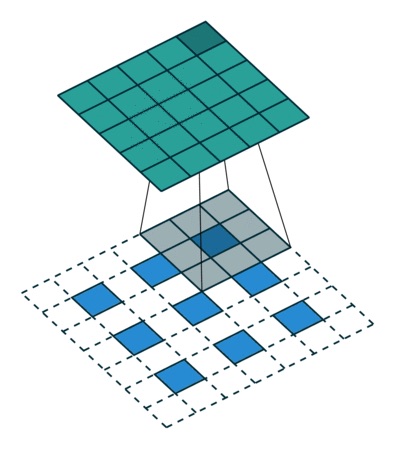} &   \includegraphics[width=2.1cm]{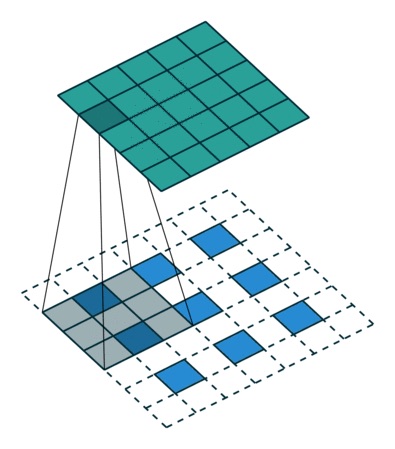} \\
    \footnotesize (a) & \footnotesize (b) & \footnotesize (c) & \footnotesize (d) & \footnotesize (e) & \footnotesize (f) \\
\end{tabular}
\caption{A deconvolutional layer in action from \cite{RefWorks:doc:5b8c0195e4b07993ca600a55}}
\label{fig:deconv_action}
\end{figure}

While a convolutional layer is suitable for extracting features of an image by selecting parts and reducing the dimensions of the image, sometimes it is interesting to do the contrary. This is the goal of deconvolutional layers or more accurately named transposed convolutions. When the need is to recreate an image from features, a deconvolutional layer is used. In Figure \ref{fig:deconv_action}, we can see a filter of size 3*3 in grey sliding through an input in blue with a padding of 1 (the white boxes surrounding the input) and a stride of 2 (which gives the spacing between the blue boxes of the input). This kind of feature is particularly useful for the Generator network in a Generative Adversarial Network (GAN). 

A \textbf{Recurrent Neural Network} (RNN) is a neural network with memory. A node in a RNN has a state determined by a linear combination of the input at time \textit{t} and its previous state at time \textit{t-1}. At each time \textit{t} the node produces an output. RNNs are used for tasks that need to take into account what happened before like predicting the next word in a sentence or translating from one language to another.
\begin{figure}[!ht]
\centering 
\includegraphics[width = 1\hsize]{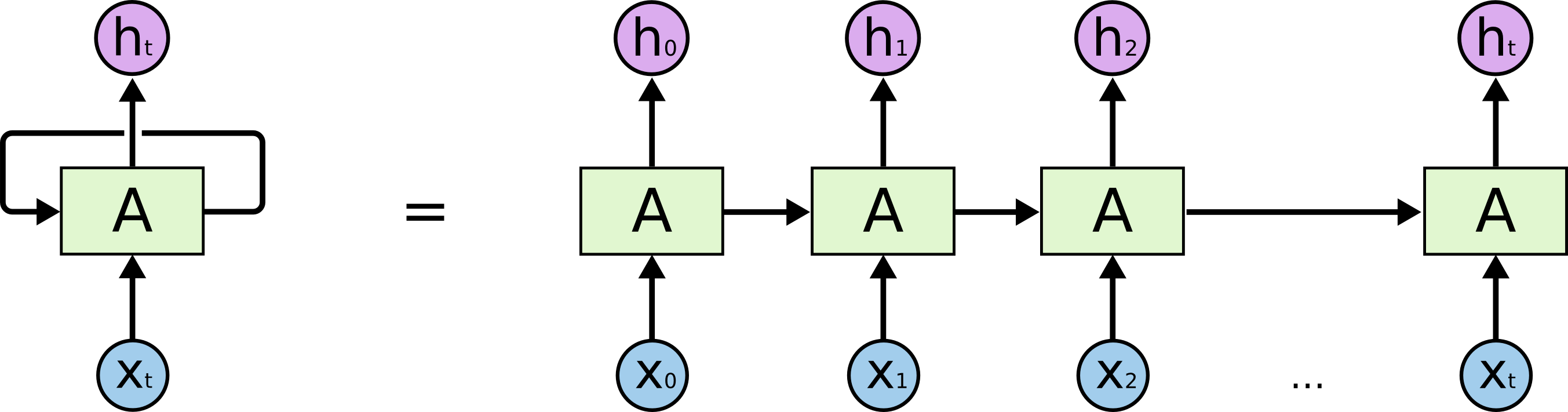} 
\caption{Graph of a RNN \cite{RefWorks:doc:5b154078e4b014266afb5926}} 
\label{fig:RNN} 
\end{figure}

A \textbf{Bidirectional Recurrent Neural Network} (BRNN) is like a RNN but with a supplementary node whose state is determined by a combination of the input at time \textit{t} and the following state at time \textit{t+1}. 
\newline
\newline
To improve the memory of RNNs, \textbf{Long Short-Term Memory} networks have been introduced. This kind of neural network is based on the same principle as the RNN but with a more complicated state node. The cell state and the output are determined by different gates which can be thought as nodes performing linear operations on the input at time \textit{t}, the previous state at time \textit{t-1} and the previous output at time \textit{t-1}. These gates let the information pass through or not and are named forget gate, input gate and output gate. Then the output of the cell and its state is passed to the following time step.
\begin{figure}[!ht]
\centering
\centerline{
   \begin{minipage}{.5\linewidth}
   \centering
      \includegraphics[scale=0.6]{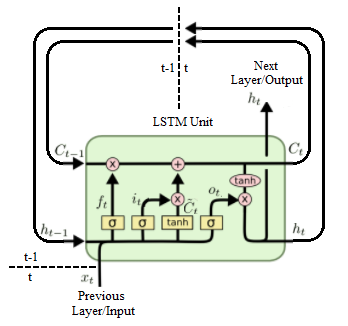}
      \caption{Graph of a LSTM \cite{RefWorks:doc:5b15422ce4b03a529bc77e56, RefWorks:doc:5b154078e4b014266afb5926}}
      \label{LSTM}
   \end{minipage} \hfill
   \begin{minipage}{.5\linewidth}
    \begin{align}
    f_t = \sigma(W_f.[h_{t-1},x_t] + b_f) \\
    i_t = \sigma(W_i.[h_{t-1},x_t] + b_i) \\
    \hat{C_t} = tanh(W_C.[h_{t-1},x_t] +b_C) \\
    C_t = f_t*C_{t-1} + i_t*\hat{C_t} \\
    o_t = \sigma(W_o.[h_{t-1},x_t] + b_o) \\
    h_t = o_t*tanh(C_t)
    \end{align}
   \end{minipage} \hfill
}
\end{figure}

\begin{figure}[!ht]
\centering 
\includegraphics[width = 0.7\hsize]{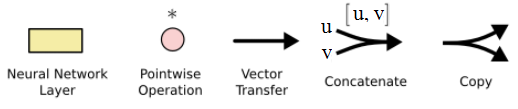} 
\caption{Caption of the LSTM and GRU graphs \cite{RefWorks:doc:5b154078e4b014266afb5926}} 
\label{fig:LSTM_GRU_caption} 
\end{figure}

As LSTMs can get complicated and so take a long time to train, a simpler architecture has been proposed : \textbf{Gated Recurrent Unit}. A GRU is a LSTM without an output gate. It possesses a reset and an update gate \cite{RefWorks:doc:5af47ed9e4b0bb78c0e4a778}. 
\begin{figure}[!ht]
\centering
\centerline{
   \begin{minipage}{.5\linewidth}
   \centering
      \includegraphics[scale=0.3]{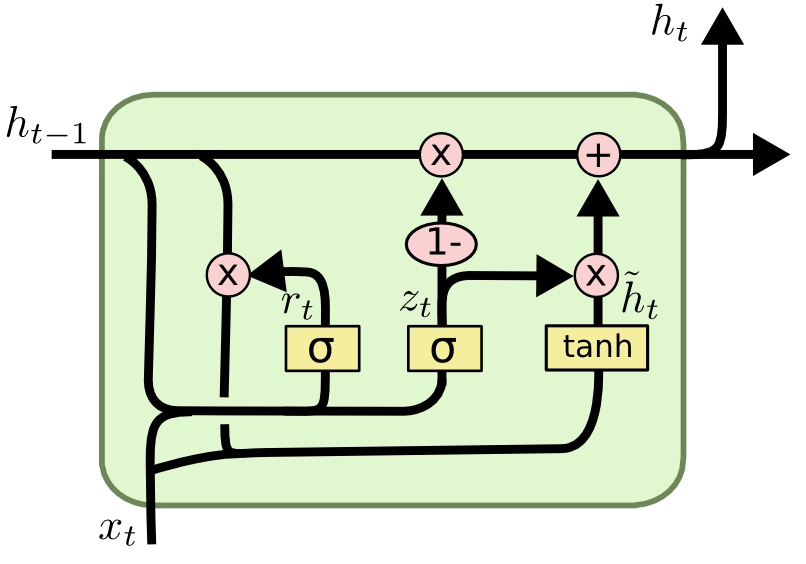}
      \caption{Graph of a GRU \cite{RefWorks:doc:5b154078e4b014266afb5926}}
      \label{GRU}
   \end{minipage} \hfill
   \begin{minipage}{.5\linewidth}
        \begin{align}
            z_t = \sigma(W_z.[h_{t-1},x_t]) \\
            r_t = \sigma(W_r.[h_{t-1},x_t]) \\
            \hat{h_t} = tanh(W.[r_t*h_{t-1},x_t]) \\
            h_t = (1-z_t)*h_{t-1} + z_t*\hat{h_t}
        \end{align}
   \end{minipage} \hfill
}
\end{figure}

\subsection{Cost or Loss functions} \label{sec:cost_fcts}

Different functions are used to determine the error between the prediction made by a neural network and what was really expected : the groundtruth labels. The choice of this loss function is important because the error is then backpropagated in the neural network so as to adjust the parameters during the training phase. 
\begin{itemize}
    \item The \textbf{Mean Squared Error} (MSE) function :
    \begin{align}
        MSE = \frac{1}{n} \sum_{i=1}^{n} (x_i - y_i)^2
    \end{align}
    where $n$ is the number of samples, $x_i$ is the prediction and $y_i$ is the observation or label.
    \item The \textbf{1 - Concordance Correlation Coefficient} (1-CCC) :
    \begin{align}
        1 - \rho_c = 1 - \frac{2 \rho \sigma_x \sigma_y}{\sigma_x^2 + \sigma_y^2 + (\mu_x - \mu_y)^2}
    \end{align}
    where $\sigma_x$ and $\mu_x$ are the variance and the mean for the prediction variable, $\sigma_y$ and $\mu_y$ are the variance and the mean for the observation variable, and $\rho$ is the correlation coefficient between the two variables.
    \item The \textbf{Huber loss} :
    \begin{align}
        L_{\delta}(a) = \begin{cases}
        \frac{1}{2}{a^2} & \text{for} |a| \le \delta, \\
        \delta(|a| - \frac{1}{2}\delta), & \text{otherwise.}
        \end{cases}
    \end{align}
    where $a$ refers to the residuals, i.e. the difference between the observed and the predicted values.

    \item The \textbf{Cross entropy} :
    \begin{align}
        H(p,q) = - \sum_{i}^n p_i log q_i
    \end{align}
    with $n$ the number of samples, $p_i$ the predicted probability for the node i to be in the category and $q_i$ is the true probability for the node i to be in the category.

\end{itemize}

\subsection{Optimization}

After a forward pass in a neural network, the error between the prediction made by the neural network and the actual value is computed. So that the neural network can perform better and come close to the real value, it has to learn. To do so, the error is backpropagated so that the different parameters $\theta$ (weights of the filters for example) of the network adjust to create better predictions. This process is called optimization. Gradient descent is the technique used in neural network to perform this optimization. The idea is to update the parameters in the opposite direction of the ascending gradient of the cost function $C$ so that the cost function is getting minimal. There are different algorithms of gradient descents : 
\begin{itemize}
    \item \textbf{Batch Gradient Descent} is the simplest one. It computes the gradient of the cost function with respect to the parameters $\theta$ for the entire dataset. 
    \begin{align}
        \theta = \theta - \eta \nabla_{\theta} C(\theta)
    \end{align}
    where $\eta$ is the learning rate.
    \item \textbf{Stochastic Gradient Descent} performs gradient descent on each training example.
    \item \textbf{Mini-batch Gradient Descent} computes the gradient descent on each mini-batch that are subsets of the training set. 
\end{itemize}
Gradient descent algorithms have been improved to get rid of the drawbacks of the previous ones like adapting the learning rate to be more precise as the cost function gets closer to the minimum or adding coefficients to smooth the gradient descent. For example, the momentum term is added to reduce the oscillations during the path traversed in learning. It is very interesting in cases where the optimum is in a ravine situation with two big slopes. While Stochastic Gradient Descent oscillates between the slopes, adding a momentum helps to converge faster towards the minimum without big oscillations.
\begin{itemize}
    \item \textbf{Stochastic Gradient Descent with momentum} :
    \begin{align}
        v_t = \gamma v_{t-1} + \eta \nabla_{\theta} C(\theta) \\
        \theta = \theta - v_t
    \end{align}
    where $\gamma$ is the momentum. It is usually set to 0.9.
\end{itemize}
Here are some of the improved gradient descent algorithms :
\begin{itemize}
    \item \textbf{Adagrad} (for Adaptive Subgradient Descent) adapts the learning rate to the frequency of the parameters. Parameters associated with frequently occurring features are updated more than parameters associated with features less frequent.
    \begin{align}
        \theta_{t+1,i} = \theta_{t,i} - \frac{\eta}{\sqrt{G_{t,ii}+\epsilon}}.\nabla_{\theta} C(\theta_{t,i})
    \end{align}
    where $i$ is the index of one parameter and $G_{t}$ is a diagonal matrix where each diagonal element $i$ is the sum of the square of the gradients w.r.t $\theta_{i}$ up to time step $t$ while $\epsilon$ is a smoothing term to avoid division by zero (usually $10^{-18}$).\newline
    For all the parameters, we can write it :
    \begin{align}
        \theta_{t+1} = \theta_{t} - \frac{\eta}{\sqrt{G_{t}+\epsilon}}.\nabla_{\theta} C(\theta_{t})
    \end{align}
    \item \textbf{Adadelta} is an extension of Adagrad which aims at improving the monotonically decreasing learning rate by taking only an interval of the past gradients. Instead of taking the sum of all the previous squared gradients for the diagonal elements of $G_t$, the influence of the oldest gradients is reduced which leads to :
    \begin{align}
        E[g^2]_t = \gamma E[g^2]_{t-1} + (1-\gamma) g_t^2 
    \end{align}
    where $g$ is the derivative of the cost function w.r.t. to the parameters. The momentum $\gamma$ is usually fixed to 0.9. \newline
    \begin{align}
        \theta_{t+1} = \theta_t - \frac{\eta}{\sqrt{E[g^2]_t + \epsilon}} g_t
    \end{align}
    It can also be written :
    \begin{align}
        \theta_{t+1} = \theta_t - \frac{\eta}{RMS[g]_t} g_t
    \end{align}
    where $RMS[g]_t$ is the Root Mean Squared Error of the gradient.
    \item \textbf{RMSprop} is a version of Adadelta with the momentum $\gamma$ set to 0.9 and the learning rate $\eta$ set to 0.001.
    \item \textbf{Adam optimizer} (Adaptative Moment estimation) keeps an exponentially decay of the past squared gradients $v_t$ and a exponentially decay of the past gradients $m_t$. 
    \begin{align}
        m_t = \beta_1 m_{t-1} + (1-\beta_1) g_t \\
        v_t = \beta_2 v_{t-1} + (1-\beta_2) g_t^2
    \end{align}
    $m_t$ can be seen as the first moment (the mean) and $v_t$ as the second moment (the uncentered variance). As $m_t$ and $v_t$ are biased towards zero (especially during the first steps and when the decay rate is small), Adam's authors define :
    \begin{align}
        \hat{m_t} = \frac{m_t}{1-\beta_1} \\
        \hat{v_t} = \frac{v_t}{1-\beta_2}
    \end{align}
    As a result : 
    \begin{align}
        \theta_{t+1} = \theta_t - \frac{\eta}{\sqrt{\hat{v_t}} + \epsilon} \hat{m_t}
    \end{align}
    The proposed values for the parameters are $\beta_1 = 0.9$, $\beta_2 = 0.999$ and $\epsilon = 10^{-8}$. Adam method is one of the state-of-the art gradient descent algorithm that we now often see in research papers. 
\end{itemize}

\subsection{Selection of interesting architectures} \label{sec:selection_archi}

This paper only talks about the last most interesting researches focused on deep learning architectures for emotion recognition.

\subsubsection{Valence and Arousal}

The research \cite{RefWorks:doc:5af1a263e4b02dbdaeebbecb} shows how deep neural networks \cite{kollia2009interweaving} can improve emotion recognition. Only the video part of AVEC 2015 dataset is used. Two architectures are presented :
\begin{itemize}
    \item A CNN architecture, through which a single-frame video is passed, consisting of 3 layers with 64, 128, 256 filters of size $5\times5$ each with a $2\times2$ max-pooling at the end of the first two layers and a quadrant pooling at the end of the third. At the top of these 3 layers a fully-connected layer containing 300 hidden units is added with a regression layer guessing the valence score. \newline
    Parameters : The cost function used is the Mean Squared Error. The training was done using stochastic gradient descent with batch size of 128, momentum equal to 0.9 and weight decay of 1e-5. The learning rate was constant : 0.01.
    \item A CNN + RNN architecture. At time \textit{t}, the CNN is fed with \textit{W} frames from time \textit{t-W} to time \textit{t}. For each frame, a 300-dimensional vector is extracted. The \textit{W} frames are then passed to a node of the RNN.  Each node regresses the output valence score and the valence score generated by the node at time \textit{t} is used for the cost function. \newline
    Parameters : The cost function used is the Mean Squared Error. The training is made with Stochastic Gradient Descent with a learning rate of 0.01, a batch size of 128 and a momentum equal to 0.9. 
\end{itemize}

For the data preprocessing, the face was first detected using the face and landmark detector Dlibml \cite{RefWorks:doc:5b0d2daae4b0c6f743c5ef95} (Frames where the face detector failed were dropped and the valence scores were interpolated). The detected landmarks points are mapped in order to ensure correspondence between frames. The eye and nose coordinates are normalized, mean substraction and contrast normalization are applied. \newline
\newline
The best performing model is the CNN + RNN with 3 hidden layers, a window length W = 100 frames, 100 hidden units in the first two recurrent layers and 50 in the last, and ReLU as the activation function. 
\newline
\newline
\newline
\newline
The second paper proposes an end-to-end multimodal emotion recognition using deep neural networks. The neural networks work on the audio and visual data of the RECOLA database. \newline
The architecture tested is divided into the video part and the audio part on top of which LSTMs are stacked. Moreover, the two parts have been trained separately. In the report, we will focus only on the video architecture which is the subject of this research :
\begin{itemize}
    \item The pre-trained Res-Net 50 \cite{RefWorks:doc:5b0e689ee4b02f452d8e6de3} is used. This network has been previously trained on the ImageNet 2012 classification dataset of 1000 classes. The first layer of the Res-Net 50 is a $7\times7$ convolutional layer with 64 filters, followed by a max-pooling layer of size $3\times3$. The Res-Net consists of 3 bottleneck architectures with convolutional layers of size $1\times1$, $3\times3$, $1\times1$ with features maps 64, 64, 256 ; 128, 128, 512 ; 256, 256, 1024 ; 512, 512, 2048. At the end an average pooling is added.
\end{itemize}
    To train the network, a 2-layer LSTM with 256 cells for each layer is stacked on top it to capture temporal information. The visual network is fine-tuned on the dataset. Then the 2-layer LSTM is discarded and only the 2048 features from the visual network and the 1280 features of the audio network are used to feed another 2-layer LSTM with 256 cells each. The LSTMs are initialized using Glorot initialization \cite{RefWorks:doc:5b0fd825e4b0a35b73cb39e4} and the weights from the unimodal training are used to train the whole network composed of the audio and video parts with the LSTM components stacked end-to-end. Glorot initialization \cite{RefWorks:doc:5b0fd825e4b0a35b73cb39e4} proposes to initialize the weights thanks to a Gaussian distribution with the mean close to 0 and the variance based on the fan-in and fan-out of the weight.
    \newline
    The cost function to minimize during the end-to-end training is :
    \begin{align}
    \frac{(1-CCC_{valence}) + (1-CCC_{arousal})}{2}.
    \end{align}
    For training, Adam optimization is used and the learning rate is fixed at 1e-4. For the video a mini-batch of 2 videos was used (because of hardware limitations). For regularization, a dropout of 0.5 was applied to all layers except the recurrent ones. For video, sequences of length 300 have shown the best results though 150 have been chosen because the results were close to 300, audio has shown best results for 150 and too long sequences can lead to overblowing gradients.
    \newline
    Data pre-processing : The image size used is $96\times96$ pixels and then increased to $110\times110$. Finally, the image is randomly cropped to equal its original size. (This produces a scale invariant model). Color augmentation is also used by introducing random brightness and saturation to the image.
    \newline
    Data post-processing : Median-filtering, centering, scaling, time-shifting were applied.
    \newline
    The addition of appearance (Local Gabor Binary Patterns from Three Orthogonal Planes) and geometric (facial landmarks) features improved the results for both valence and arousal. 
\newline
\newline
\newline
\newline
In "Recognition of Affect in the wild using deep neural networks" \cite{kollias2}, an end-to-end architecture was trained on Aff-Wild to detect Valence and Arousal. It is of particular interest because the videos from the dataset have been recorded in-the-wild.
Different architectures have been implemented :
\begin{itemize}
    \item An architecture based on the Res-Net L50 network \cite{kollias7}.
    \item An architecture based on VGG-Face network \cite{kollias14}.
    \item An architecture based on VGG-16 network \cite{RefWorks:doc:5b0e67fce4b0a35b73cb168b}.
\end{itemize}
Moreover, two different approaches have been tried :
\begin{itemize}
    \item A CNN only.
    \item A CNN + RNN to exploit the dynamic information of the data.
\end{itemize}
For each architecture and approaches, two scenarios have been made :
\begin{itemize}
    \item The network is applied directly on cropped facial video frame to produce Valence and Arousal predictions.
    \item The network uses both the facial appearance and the facial landmarks.
\end{itemize}

For the training mode, the Adam optimizer and two loss functions : the Concordance Correlation Coefficient and the Mean Squared Error were used. 
\newline
For initialization, the weights of the networks were either randomly set or pre-trained with ImageNet database by doing fine-tuning. Fine-tuning is technique of learning for neural networks which have already been trained on big datasets and needs to be trained on a more specific task. For example Res-Net or VGG-16 have been trained for object recognition and it is interesting to fine-tune these networks on face recognition so as they can be more specific. Fine-tuning often involves truncating the last prediction layer (for example Res-Net has been trained on ImageNet to predict 1,000 classes, so its final layer has 1,000 nodes but for Valence and Arousal only two nodes are needed), a small learning rate as we expect the pre-trained weights to be good already, and freezing the first layers of the networks that focus on more abstract and general concepts to change only the last layers that focus more on details that can be interesting in Valence and Arousal for example.
\newline
For the CNN-only architectures the best results were obtained with a batch size of 80, a learning rate of 0.001, and a dropout of 0.5. 
\newline
For the CNN + RNN, the best results were obtained for 1 fully-connected layer in the CNN and 2 hidden layers in RNN with a batch size of 100 ($\approx 3 sec.$). 
\newline
\newline
Results : Only-CNN architectures perform best with VGG-Face pre-trained on ImageNet with the facial landmarks. Moreover if one network for Valence and another for Arousal is used the results are even better (Furthermore, the use of the mean of annotators also increase the performance). 
\newline 
For CNN + RNN architectures, GRU performs better than LSTM. This type of architecture outperforms the simple CNN architecture by 24\% in valence estimation and 23\% in arousal estimation. On small datasets, GRUs usually perform better than LSTMs and the neural network is faster to train as there is less computation.

\subsubsection{Facial Action Units}

In this paper \cite{RefWorks:doc:5af1cc28e4b0011b0775e7ee}, a CNN + BLSTM architecture is used with binary mask resulting in the winning results at the FERA 2015 challenge by a margin of 10\%. The FERA 2015 challenge is based on two datasets : the SEMAINE and the BP4D datasets. The latter is made of videos of people reacting to emotion elicitation tasks and records 11 action units. 
\newline
Data preprocessing :
    Facial landmarks are tracked and alignement is performed using Procrustes transform. Binary masks are then applied to the face in order to select regions. Finally a dynamic encoding is performed : the image sequences are modified by taking the difference with the current frame, making it easier to learn the dynamics of the sequences. 
\newline
The architecture of the network is :
\begin{itemize}
    \item Two input streams, one for the sequence of image regions and another for the sequence of binary mask. Both streams apply a convolutional neural network of 32 filters of size $5\times5\times(2n+1)$ (where $n$ is the number of images) followed by a max-pooling layer of size $3\times3\times1$. 
    \item The outputs of both streams are fed to two convolutional layers : the first is 64 filters of size $5\times5\times64$ and the second consists of 128 filters of size $4\times4\times64$.
    \item Finally, a fully-connected layer with 3072 units is stacked with a dropout of 0.2. (The output layer consists of 2 units : one for the positive class, the other for the negative).
    \item The 3072 units are then passed to a BLSTM network with a single hidden layer of 300 units. The ouput of the BLTSM serves as the final decision value for the occurrence of an AU.  
\end{itemize}
The activation function is ReLU. 
\newline
\newline
\newline
\newline
"Transfer learning for Action Unit Recognition" is the title of this paper \cite{RefWorks:doc:5af48017e4b0f7bd1fabb8de}. The database used is the DISFA dataset.
Different networks are tested in this research :
\begin{itemize}
    \item AlexNet \cite{RefWorks:doc:5b0e6651e4b01f2c3e37bcbb}.
    \item ZFNet \cite{RefWorks:doc:5b0e6a23e4b0dc9b736d309b}.
    \item VGGNet \cite{RefWorks:doc:5b0e67fce4b0a35b73cb168b}.
    \item GoogLeNet \cite{RefWorks:doc:5af47bc8e4b0ab1e65d9dd33}.
    \item ResNet \cite{RefWorks:doc:5b0e689ee4b02f452d8e6de3}.
\end{itemize}
The generic architecture combining these networks are :
\begin{itemize}
    \item Raw image that can be cropped, resized, normalized (For example the mean of the training set images can be substracted to the other images like for VGG-Face \cite{RefWorks:doc:5b0ff4e6e4b01f2c3e37e5ea}).
    \item One of the network mentioned above. The output dimensions vary from one network to another. 
    \item A classifier that can be a LDA, a SVM or LSTM. 
    \newline
    For the linear classifiers (LDA, SVM), the features are normalized so that the mean is zero and the standard deviation is 1. One classifier is learned for each AU. 
    \newline
    LSTM is stacked only with Res-Net 152 due to the lower dimensionality of the output vector (2048). The LSTM has a single hidden layer of 200 units. The learning rate is 0.0001 and the momentum is fixed to 0.9. To prevent overfitting, a Gaussian noise is applied with a zero mean and a standard deviation of 0.1. During training, the LSTM is fed with 3 frames before and after the frame at time \textit{t} and during testing all the frames are fed all at once. 
    \item An output classification layer which determines if an AU is present or absent. 
\end{itemize}
F1 score is used to rate the classification task. \newline
VGG-Face+SVM performs the best on F1 score. When used, LSTM is not very powerful because it does not scale to large dimensions and the lack of data.
\newline
By using fine-tuning, VGG-Fast performs better.
\\
\\
Through these papers, we saw that the best performing techniques are end-to-end deep neural networks with little computer vision techniques involved. As a result, during this project we will focus on an end-to-end deep neural network which has not been tested a lot in emotion recognition : a Generative Adversarial Network.

\chapter{Project}

This project consists of two parts :
\begin{itemize}
    \item The extension of the existing Aff-Wild (Affect-in-the-Wild) database. This dataset (described \ref{sec:va_database}) has already been annotated with Valence and Arousal. The first part of the project was to annotate it with Action Units so that the database contains both annotations. Moreover, data pre-processing has been made to prepare the dataset for the second part.
    \item The creation of an end-to-end deep neural network able to predict both types of annotations so that it achieves high performance in this newly created database and generate realistic images of faces. 
\end{itemize}

\section{Database creation}

\subsection{Context}

The Aff-Wild contains around 300 videos only annotated for Valence Arousal. For this project, a subset of this database has been selected and has been annotated for the 8 Action Units described earlier. The creation of this new database has been organized in different steps : selecting the videos, formating the videos, annotating the videos, analysing the statistics of the videos. These steps have been repeated twice so as to have a suitable database. Finally the database has been divided into a training and testing set.

\subsection{Annotation}

\begin{figure}[!ht]
    \centering
    \includegraphics[scale=0.3]{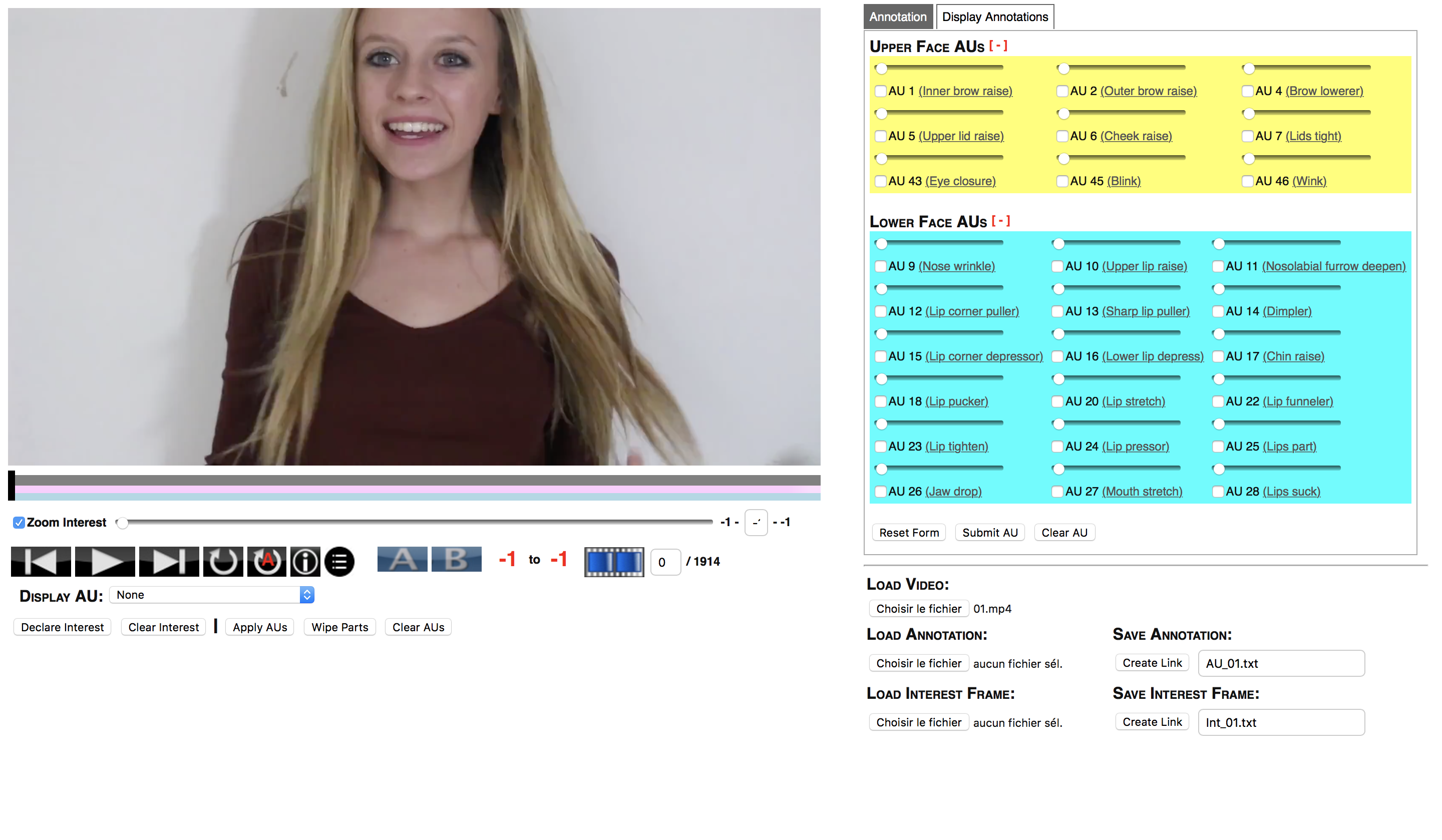}
    \caption{The GUI for the Action Unit annotation}
    \label{fig:annotation_interface}
\end{figure}

After my supervisor selected a set of videos from the Aff-Wild database containing the 8 Action Units we have decided to study, the videos were all converted to the same format, i.e. .mp4 and 30 fps. This standard format would make the further treatments and analysis easier. 
\\
\\
Then, the videos annotation phase for the 8 Action Units started. The annotation interface (on Figure \ref{fig:annotation_interface}) enables to annotate each Action Unit independently and frame-by-frame for each video. The method was simple : the video was watched once entirely so as to spot the most important parts, i.e. the parts in which the person reacts the most. Then, the annotation for each Action Unit was performed by watching the video entirely and selecting the frames where the AU was present. This process was repeated for each Action Unit. At the end of the annotation of a video, a file containing all the annotations was created. Each line of the file contained the frame number and for each AU a boolean indicating the presence or the absence followed by the intensity of this AU (for this project the intensity was always put at 1 because the project was to focus on the presence or the absence of the Action Unit not its intensity). 
\\
This part of the project was the longest because it required a lot of time and concentration to annotate precisely the videos frame-by-frame for 8 Action Units. Moreover, some Action Units were more complicated to annotate than others : for example, the Action Unit 20 (Lip stretch) is difficult to annotate because it is not very frequent or the Action Unit 25 (Lips part) because it should not be mistaken with the person speaking. 
\\
\\
After this first round of annotations (finished on May 23rd), my supervisor and I analyzed the first 35 videos annotated so as to better understand the dataset and see if more videos needed to be annotated. The Tables \ref{tab:gen_video_stats} and \ref{tab:frame_number_action_unit} describe this first batch of videos.

\begin{table}[!ht]
    \centering
\begin{tabular}{|c|c|c|c|}\hline
    {} & \textbf{Batch $n^{o} 1$} & \textbf{Batch $n^{o} 2$} & \textbf{Total} \\
    \rowcolor{gray!40}
    Number of videos & 35 & 29 & 64 \\
    \rowcolor{gray!10}
    Total number of frames & 138,754 & 94,705 & 233,459 \\
    \rowcolor{gray!40}
    \pbox{5cm}{Number of frames \\ containing at least one AU} & \pbox{5cm}{81,352 \\ (58.6\%)} & \pbox{5cm}{57,946 \\ (61.2\%)} & \pbox{5cm}{139,298 \\ (59.7\%)} \\
    \rowcolor{gray!10}
    Total number of AUs annotated & 139,952 & 95,697 & 235,649 \\
   \hline
\end{tabular}
    \caption{General statistics on the videos annotated}
    \label{tab:gen_video_stats}
\end{table}

\begin{table}[!ht]
    \centering
\begin{tabular}{|c|c c c|c c c|}\hline
  \multirow{2}{*}{\textbf{Action Units}} & \multicolumn{3}{c}{\textbf{Number of Action Units}} & \multicolumn{3}{|c|}{\textbf{Percentage}} \\
   & Batch $n^{o}$ 1 & Batch $n^{o}$ 2 & Total & Batch $n^{o}$ 1 & Batch $n^{o}$ 2 & Total \\
   \rowcolor{gray!40}
   AU 1 &  18,785 & 25,163 & 43,948 & 13.4 & 26.3 & 18.6 \\
   \rowcolor{gray!10}
   AU 2 & 13,483 & 11,829 & 25,312 & 9.6 & 12.4 & 10.7 \\
   \rowcolor{gray!40}
   AU 4 & 23,071 & 15,808 & 38,879 & 16.5 & 16.5 & 16.5 \\
   \rowcolor{gray!10}
   AU 6 & 31,263 & 16,922 & 48,185 & 22.3 & 17.7 & 20.4 \\
      \rowcolor{gray!40}
   AU 12 & 28,915 & 16,376 & 45,291 & 20.7 & 17.1 & 19.2 \\
      \rowcolor{gray!10}
   AU 15 & 4,534 & 2,489 & 7,023 & 3.2 & 2.6 & 3.1 \\
    \rowcolor{gray!40}
   AU 20 & 5,245 & 4,025 & 9,270 & 3.7 & 4.2 & 3.9 \\
         \rowcolor{gray!10}
   AU 25 & 14,656 & 3,085 & 17,741 & 10.6 & 3.2 & 7.6 \\
   \hline
\end{tabular}
    \caption{Number and percentage of frames for videos for each Action Unit}
    \label{tab:frame_number_action_unit}
\end{table}

From these figures we can see that the distribution of Action Units is not balanced. Some Action Units are over-represented like AU 6 with 22.3\% and others are under-represented like AU 15 with 3.2\%. This result could be expected as this database mainly consists of videos from YouTube and people share videos on the platform where they are happy. This is the reason why the Action Unit 6 which is the Cheek raiser is much more frequent than the Action Unit 15 that is the Lip corner depressor. To have enough frames for each Action Unit, even the Action Units under-represented, a second batch of 29 videos has been annotated (finished on June 18th). This second batch has been chosen so that videos with under-represented Action Units were chosen preferentially. As a result, even though we knew that the distribution of Action Units would still be unbalanced due to the fact that some emotions are sometimes more frequent than others and so people tend to use more some face muscles than others, we have more frames for the under-represented Action Units. The new statistics for the all 64 videos are gathered in Tables \ref{tab:gen_video_stats} and \ref{tab:frame_number_action_unit}.

Finally, we see that the dataset is still unbalanced but for Action Units it is nearly impossible to have a dataset where each Action Unit is represented equally. The most important thing is to have enough data, even for the under-represented Action Units, so that the neural networks will not overfit. The augmentation of the dataset following this second batch of videos brings about an increase of 71\% in the number of frames with at least one AU.

\subsection{Selection}

The next part of the creation of the dataset was the selection of the suitable images and the corresponding annotations. \\
Thanks to the library of the Menpo project \cite{RefWorks:doc:5b7abccce4b00447110ca60d}, the images with faces were extracted and cropped to the faces. A function in the menpodetect library \cite{RefWorks:doc:5b8aa565e4b036495fd78c2c} is an implementation of a paper (\cite{RefWorks:doc:5b8aa62ce4b01735413d84df}) that has reached top performance in face detection \cite{avrithis2000broadcast} thanks to a Deformable Part Model (DPM). Besides detecting faces, this algorithm also creates landmarks which are 4 coordinates on the original image that are the reference points for cropping the image. Most of the time, only one individual appeared in these videos but sometimes the algorithm from Menpo could detect two faces (it is sometimes a pure mistake but sometimes it is due to the fact that the videos to which people are reacting are displayed in the corner of the image and can show other people like actors when people watch trailers for example). This the reason why an algorithm was created to select only one image in the case of the Menpo algorithm detects two faces and so generates two images. When two cropped images are possible for one frame, the algorithm compare the landmarks from which the cropped images are generated from the landmarks of the previous images. The set of landmarks (which consists of 4 points) whose center is nearest to the center of the set of landmarks from the previous image is kept and the other set of landmarks is not selected. Finally, a manual check is carried out and some manual modifications are made to ensure that only the images corresponding to faces are kept in the dataset. \\
Once only the images corresponding to faces are selected in the dataset, the Action Units for the corresponding frames are selected as labels. Reciprocally, the frames without Action Units are not selected. 
\\
\\
At this point, the dataset with only the Action Units annotations  was nearly ready. To have a complete dataset, the Valence Arousal annotations needed to be added. Indeed, the goal of the first part of the project was to create a new dataset with both Action Units annotations (which had to be created) and Valence Arousal annotations (which were present in the original dataset). First, an interpolation has been made to have VA annotations corresponding to the converted videos to 30 fps. Then, as the interpolation could produce +/- 2 annotations compared to AUs, the VA annotations have been either extended or cropped if needed. Only the VA annotations matching the selected frames with faces were kept. Finally, a file containing the frame number, the AU annotations and the VA annotations combined was created corresponding to the frames previously generated during the AU annotation.

\subsection{Analysis of the newly created dataset}

To be ready for our deep learning algorithms, the dataset needed to be split between a training and a testing set with approximately equivalent distributions. To generate the training and testing set, the algorithm selected videos at random into the training set until it reaches 80\% of the frames. One subtlety was that if a video containing one person that was also present in other videos was selected into a set, then all videos containing this person should be added to the set. Indeed, this prevents data leakage which would lead to unwanted improvements of the results on the testing set because the neural network would have already seen the person in the training set. After the algorithm has run several times, the most similar distribution between the training and testing set was selected. \\
The \textbf{training set} contains the following videos : 2, 3, 4, 5, 6, 7, 8, 9, 10, 12, 13, 14, 15, 17, 18, 19, 20, 22, 23, 24, 25, 26, 27, 28, 30, 32, 33, 34, 35, 36, 37, 38, 40, 41, 42, 43, 44, 46, 47, 48, 50, 52, 53, 54, 55, 56, 57, 58, 59, 60, 61, 62, 63 and 64, making a total of \textbf{107,661 frames}. \\
The \textbf{testing set} contains the following videos : 1, 11, 16, 21, 29, 31, 39, 45, 49 and 51, totalling \textbf{23,134 frames}. \\
Therefore, the training set represents 82.3 \% of the total number of frames (\textbf{130,795}), and so the testing set 17.3 \% of the total number of frames. In total there are \textbf{59 identities}, i.e. 59 different persons : 49 in the training set and 10 in the testing set. The repartition between males and females videos is described in the Table \ref{tab:subject_database}.

\begin{table}[!ht]
    \centering
\begin{tabular}{|c|c|c|}
    \hline
    \textbf{Set} & \textbf{Number of males} & \textbf{Number of females} \\
    \rowcolor{gray!40}
    Training set & 30 & 19 \\
    \rowcolor{gray!10}
    Testing set & 5 & 5 \\
   \hline
\end{tabular}
    \caption{Number of subjects in the new database}
    \label{tab:subject_database}
\end{table}

\begin{figure}[!ht]
\centerline{
\begin{minipage}{1\linewidth}
   \centering
      \includegraphics[scale=0.5]{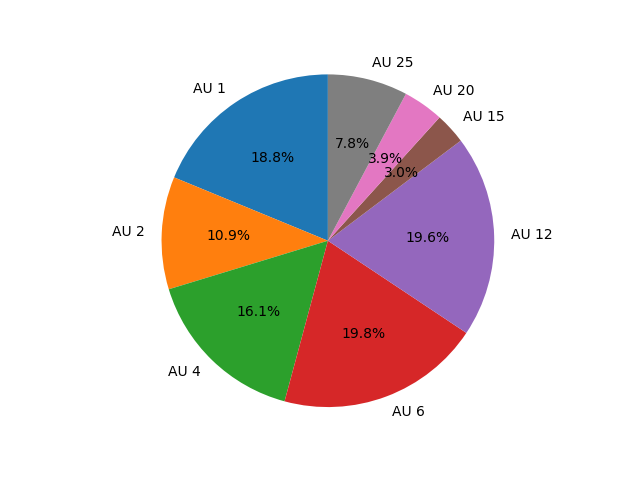}
      \caption{Pie chart of the distribution of Action Units in whole dataset}
      \label{fig:au_total_dist}
   \end{minipage} \hfill
}
\centering
\centerline{
   \begin{minipage}{.5\linewidth}
   \centering
      \includegraphics[scale=0.5]{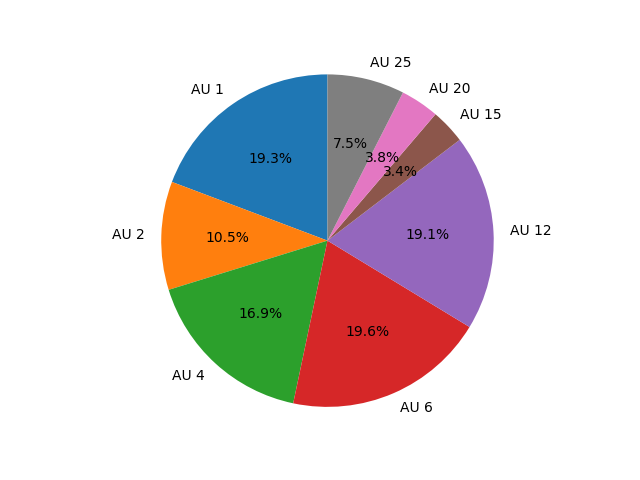}
      \caption{Pie chart of the distribution of Action Units in the training set}
      \label{fig:au_training_dist}
   \end{minipage} \hfill
   \begin{minipage}{.5\linewidth}
   \centering
      \includegraphics[scale=0.5]{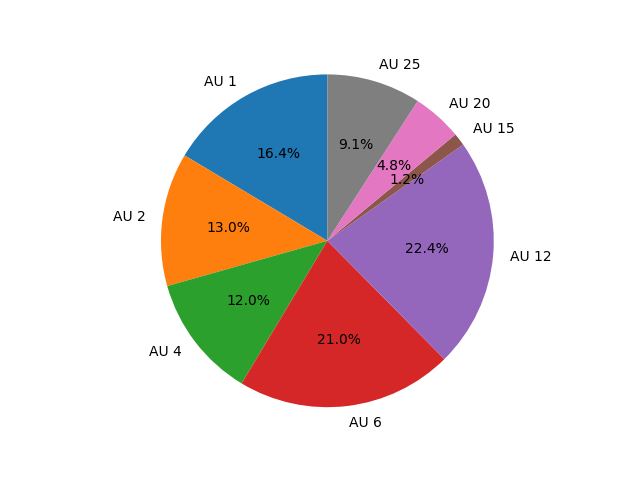}
      \caption{Pie chart of the distribution of Action Units in the testing set}
      \label{fig:au_testing_dist}
   \end{minipage} \hfill
}

\end{figure}

From the Figures \ref{fig:au_total_dist}, \ref{fig:au_training_dist} and \ref{fig:au_testing_dist}, we can see that the percentage of Action Units is quite balanced between the training and testing set considering the constraints of forming a training and testing set mentioned above. This is confirmed by the Table \ref{tab:datasetAU} : the maximum percentage gap is under 5 \% (4.91 \% for the Action Unit 4) and the minimum percentage gap is a little above 1 \% (1.09 \% for the Action Unit 20). The fact that the maximum percentage gap is for the Action Unit 4 might be explainable by the fact that the few videos displaying negative emotions (Action Unit 4 is Brow lowerer) have numerous frames with this Action Unit. The minimum percentage gap for the Action Unit 20 (Lip stretcher) can be understood because the number of frames with this Action Unit is not very high and it can be present in any type of videos (happy or sad) which explains a good distribution.

\begin{table}[!ht]
    \centering
\begin{tabular}{|c|c c c|c c c|}\hline
  \multirow{2}{*}{\textbf{Action Units}} & \multicolumn{3}{c}{\textbf{Number of Action Units}} & \multicolumn{3}{|c|}{\textbf{Percentage}} \\
   & Training & Testing & Total & Training & Testing & Total \\
   \rowcolor{gray!40}
   AU 1 & 35,628 & 6,113 & 41,741 & 19.26 & 16.42 & 18.78 \\
   \rowcolor{gray!10}
   AU 2 & 19,473 & 4,825 & 24,298 & 10.53 & 12.96 & 10.93 \\
   \rowcolor{gray!40}
   AU 4 & 31,291 & 4,467 & 35,758 & 16.91 & 12.00 & 16.09 \\
   \rowcolor{gray!10}
   AU 6 & 36,286 & 7,825 & 44,111 & 19.61 & 21.02 & 19.85 \\
      \rowcolor{gray!40}
   AU 12 & 35,268 & 8,340 & 43,608 & 19.06 & 22.41 & 19.62 \\
      \rowcolor{gray!10}
   AU 15 & 6,249 & 451 & 6,700 & 3.38 & 1.21 & 3.02 \\
    \rowcolor{gray!40}
   AU 20 & 6,947 & 1,803 & 8,750 & 3.75 & 4.84 & 3.94 \\
         \rowcolor{gray!10}
   AU 25 & 13,876 & 3,399 & 17,275 & 7.50 & 9.14 & 7.77 \\
   \hline
   Total & 185,018 & 37,223 & 222,241 & 100 & 100 & 100 \\
   \hline
\end{tabular}
    \caption{Distributions of Action Units between the training and testing sets}
    \label{tab:datasetAU}
\end{table}

\begin{figure}[!ht]
\centering
\centerline{
   \begin{minipage}{.5\linewidth}
   \centering
      \includegraphics[scale=0.5]{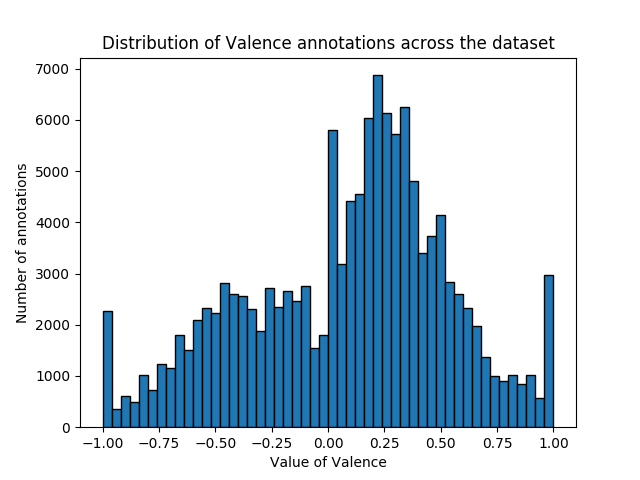}
      \caption{Histogram of the distribution of Valence annotations in the whole dataset}
      \label{fig:valence_dist}
   \end{minipage} \hfill
   \begin{minipage}{.5\linewidth}
   \centering
      \includegraphics[scale=0.5]{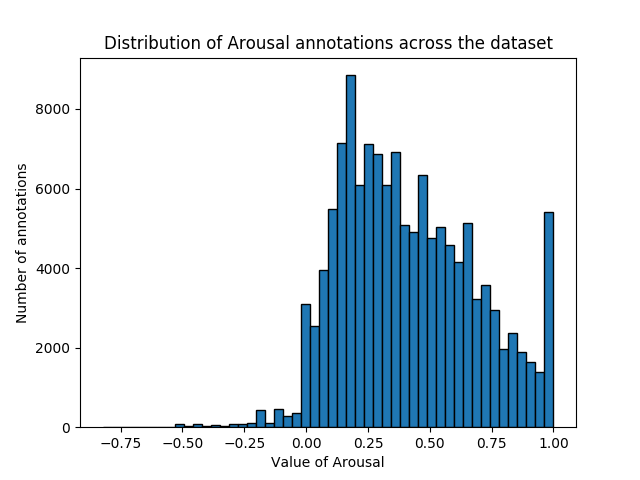}
      \caption{Histogram of the distribution of Arousal annotations in the whole dataset}
      \label{fig:arousal_dist}
   \end{minipage} \hfill
}
\end{figure}

\begin{figure}[!ht]
\centering
\centerline{
   \begin{minipage}{.5\linewidth}
   \centering
      \includegraphics[scale=0.5]{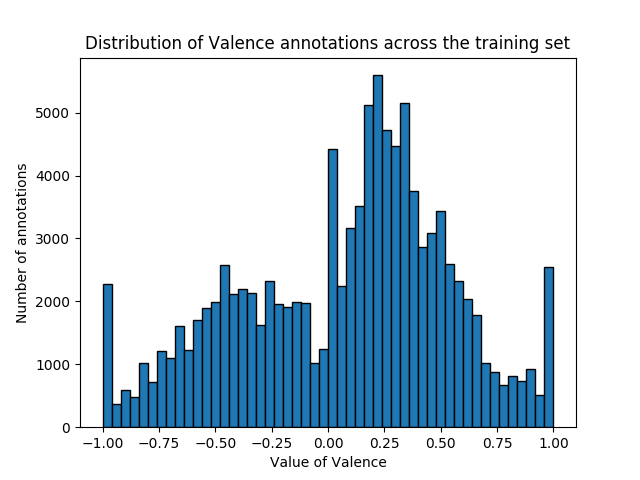}
      \caption{Histogram of the distribution of Valence annotations in the training set}
      \label{fig:valence_train_dist}
   \end{minipage} \hfill
   \begin{minipage}{.5\linewidth}
   \centering
      \includegraphics[scale=0.5]{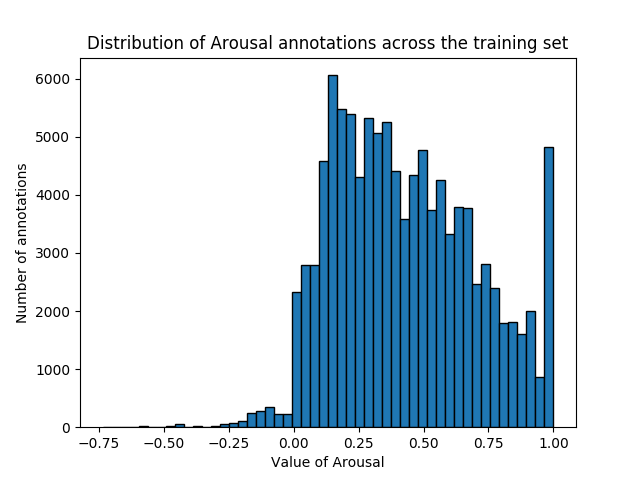}
      \caption{Histogram of the distribution of Arousal annotations in the training set}
      \label{fig:arousal_train_dist}
   \end{minipage} \hfill
}
\end{figure}

\begin{figure}[!ht]
\centering
\centerline{
   \begin{minipage}{.5\linewidth}
   \centering
      \includegraphics[scale=0.5]{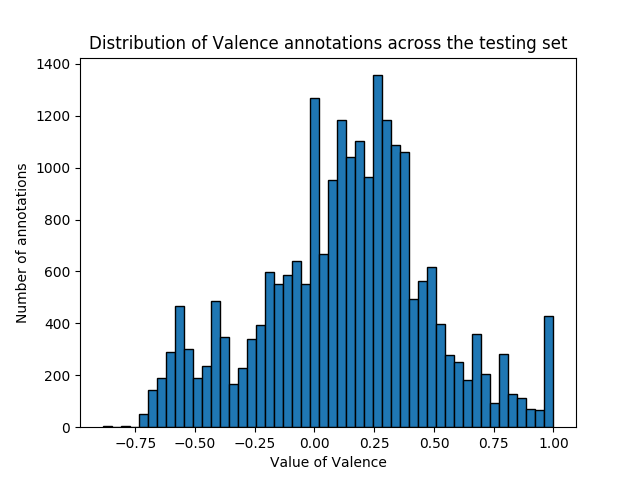}
      \caption{Histogram of the distribution of Valence annotations in the testing set}
      \label{fig:valence_test_dist}
   \end{minipage} \hfill
   \begin{minipage}{.5\linewidth}
   \centering
      \includegraphics[scale=0.5]{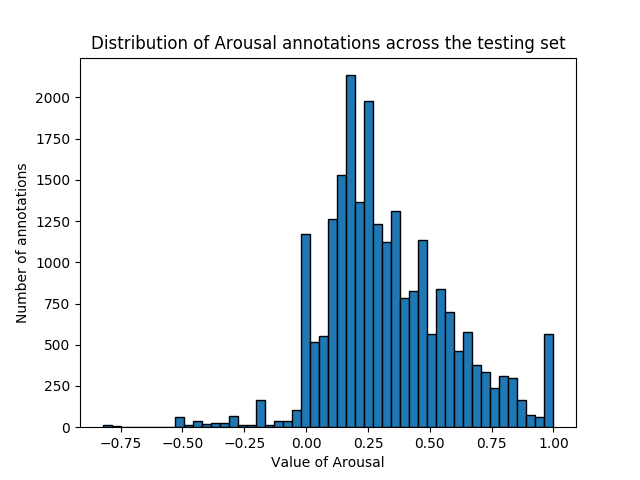}
      \caption{Histogram of the distribution of Arousal annotations in the testing set}
      \label{fig:arousal_test_dist}
   \end{minipage} \hfill
}
\end{figure}

\clearpage

\begin{figure}
\begin{tabular}{cc}
  \includegraphics[scale=0.45]{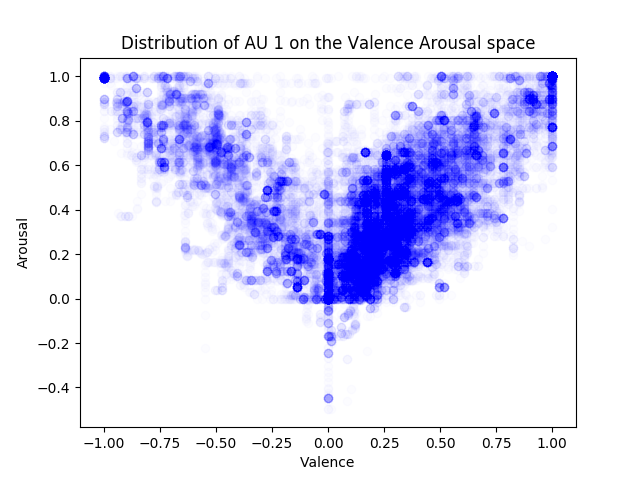} &   \includegraphics[scale=0.45]{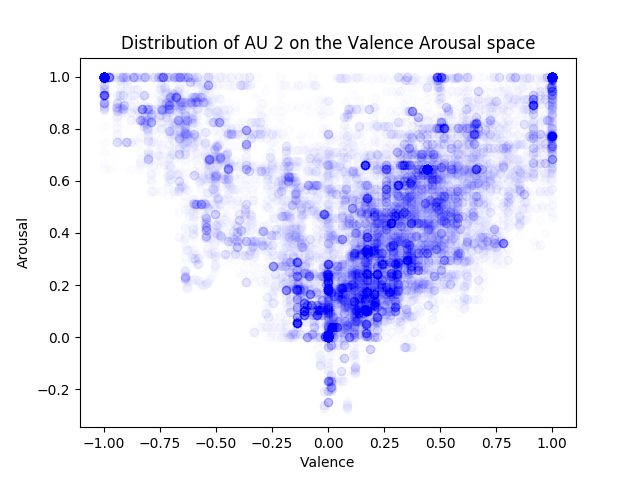} \\
 \includegraphics[scale=0.45]{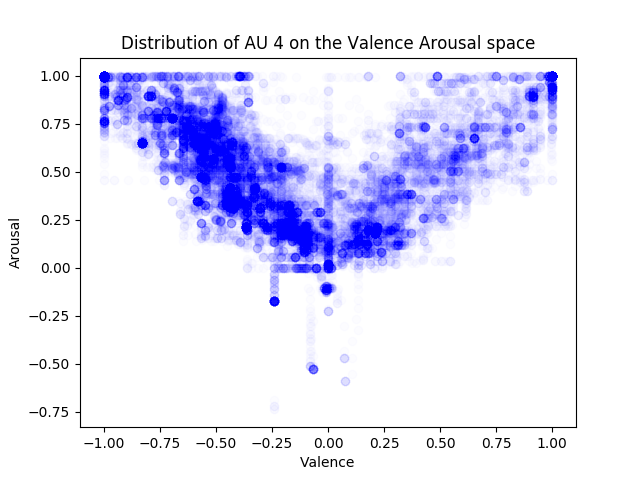} &   \includegraphics[scale=0.45]{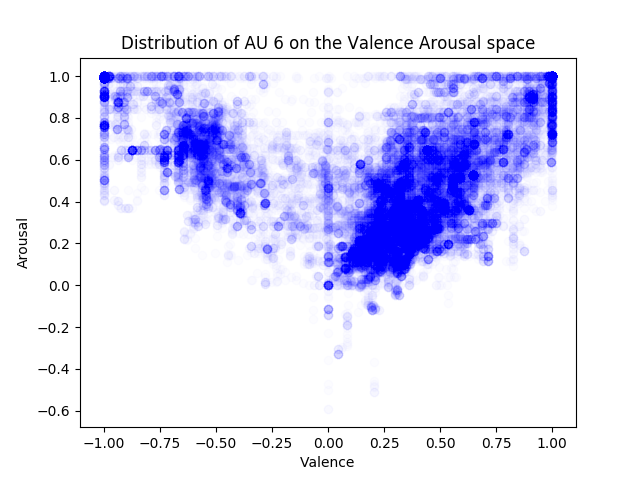} \\
   \includegraphics[scale=0.45]{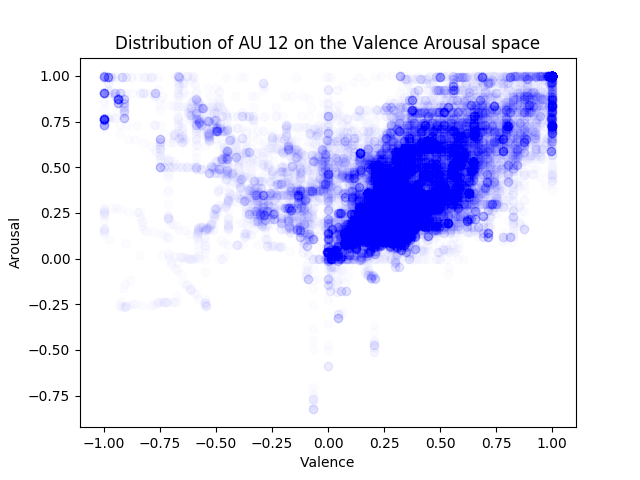} &   \includegraphics[scale=0.45]{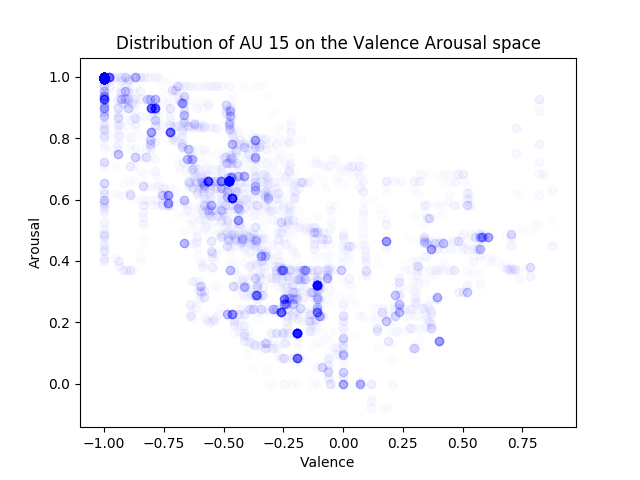} \\
 \includegraphics[scale=0.45]{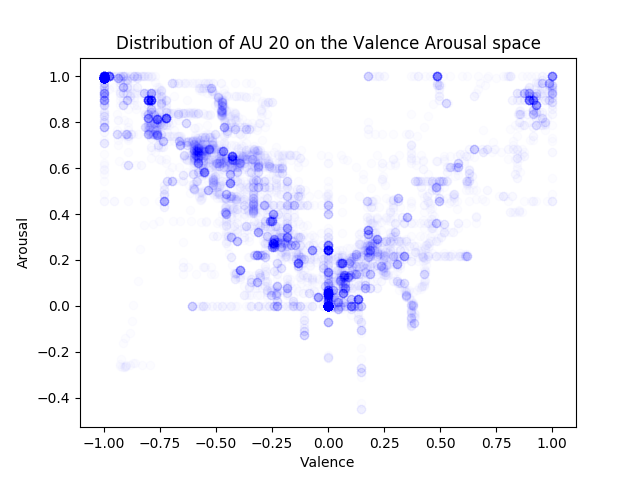} &   \includegraphics[scale=0.45]{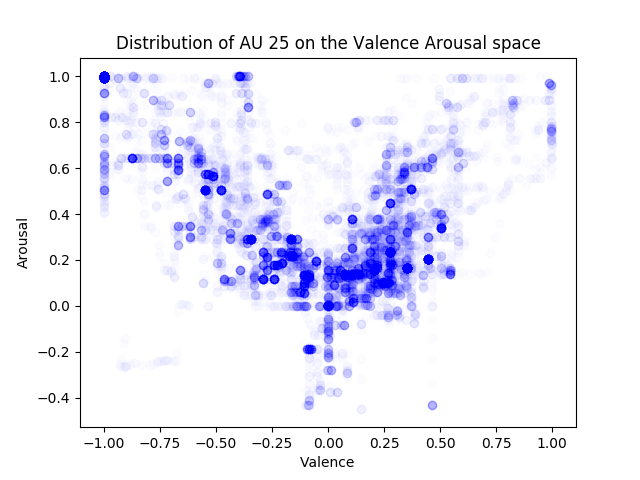} 
\end{tabular}
\caption{Graphs of the distribution of each Action Unit in the Valence Arousal space}
\label{fig:au_dist_va}
\end{figure}

Figures \ref{fig:valence_dist} and \ref{fig:arousal_dist} show the distribution of Valence and Arousal in the final dataset. Valence represents how positive ($valence>0$) or negative ($valence<0$) the person feels. On Figure \ref{fig:valence_dist}, Valence distribution is quite balanced with more data for values between 0 and 0.5 which is coherent with the videos in the database showing people filming themselves experiencing positive feelings (watching trailers, series etc.) more frequently than negative ones (crying to a movie etc.). Two spikes at -1 and 1 can be noticed : the spike at 1 corresponds to a very positive attitude (e.g. a person being in heaven) while the spike at -1 correspond to a very negative feeling (e.g. a person crying or mourning). On the other hand, the Arousal distribution is clearly unbalanced. Arousal represents how active ($arousal>0$) or ($passive<0$) a person is when feeling an emotion. This can be understood because the six main emotions are on the top part ($arousal>0$) of the Valence-Arousal graph, as shown in Figure \ref{fig:emotion_wheel}. We also can observe a spike at 1 which depicts a person being very excited. Similar distributions can be observed for the training set and the testing set : balanced distributions for Valence (Figures \ref{fig:valence_train_dist} and \ref{fig:valence_test_dist}) and unbalanced for Arousal (Figures \ref{fig:arousal_train_dist} and \ref{fig:arousal_test_dist}). Moreover most of the Action Units that we annotated correspond to a person being active like raising eyebrows or smiling. This is confirmed by Figure \ref{fig:au_dist_va} which shows the distribution of Action Units in the Valence Arousal space being more dense in the upper part of the VA space. With this Figure, Action Units can also be associated with positive or negative feelings : Action Units 1 (Inner brow raiser), 2 (Outer brow raiser), 6 (Cheek raiser) and 12 (Lip corner puller) tend to have more dots for positive valence and so are more associated with positive emotions while Action Units 4 (Brow lowerer), 15 (Lip corner depressor) and 20 (Lip stretcher) have more dots in the negative valence domain and therefore are more likely to match negative feelings. The Action Unit 25 (Lips part) is quite balanced between the negative and positive valence area. Furthermore, it is interesting to notice that for each Action Unit a line could be drawn : $y=|x|$, with $y : arousal$ and $x : valence$, which shows what is the correlation between Valence and Arousal given an Action Unit. For instance, the more positive the person is when doing the Action Unit ($x=valence>0$), the more excited ($y=arousal>0$) and the more negative when doing the Action Unit ($x=valence<0$), the more excited ($y=arousal>0$).

\clearpage

\begin{figure}[!ht]
\begin{tabular}{cccccc}
  \includegraphics[width=2.1cm]{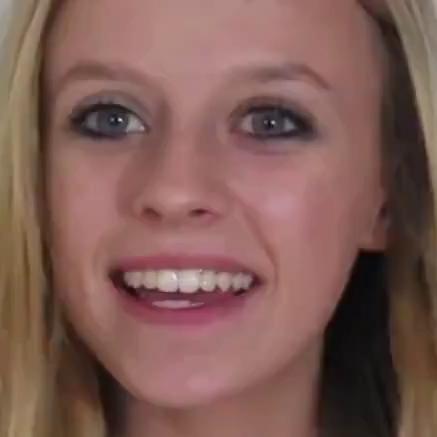} &  \includegraphics[width=2.1cm]{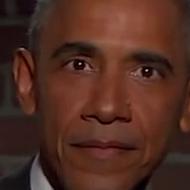} &
 \includegraphics[width=2.1cm]{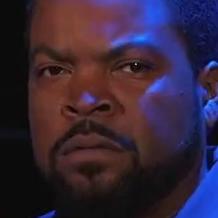} &   \includegraphics[width=2.1cm]{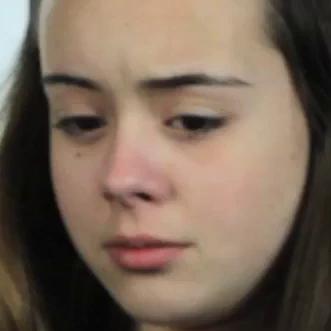} &
   \includegraphics[width=2.1cm]{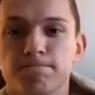} &   \includegraphics[width=2.1cm]{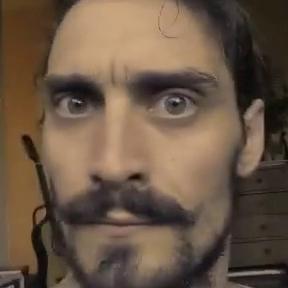} \\
\end{tabular}
\caption{Some representatives of the new database - Part 1}
\label{fig:face_database_1}
\end{figure}

\begin{figure}[!ht]
\begin{tabular}{cccccc}
  \includegraphics[width=2.1cm]{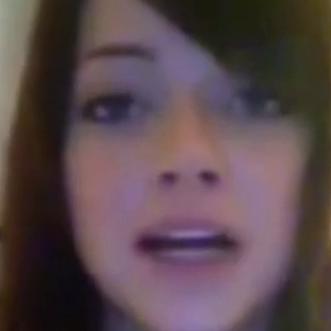} &  \includegraphics[width=2.1cm]{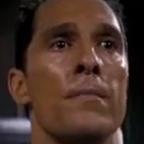} &
 \includegraphics[width=2.1cm]{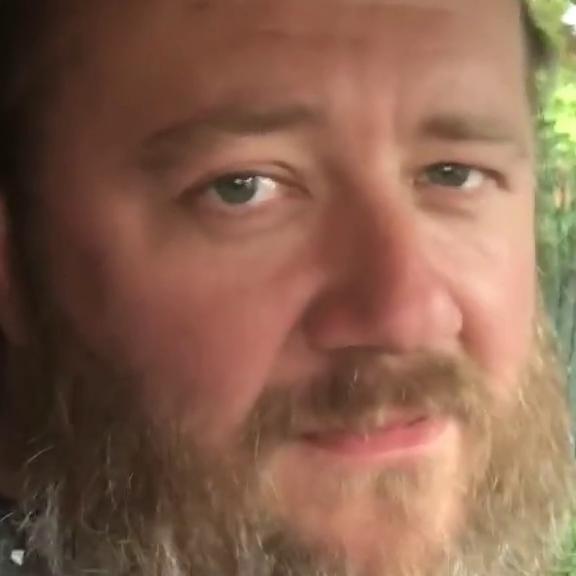} &   \includegraphics[width=2.1cm]{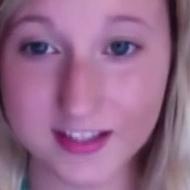} &
   \includegraphics[width=2.1cm]{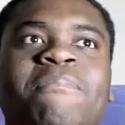} &   \includegraphics[width=2.1cm]{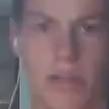} \\
\end{tabular}
\caption{Some representatives of the new database - Part 2}
\label{fig:face_database_2}
\end{figure}

\begin{figure}[!ht]
\begin{tabular}{cccccc}
  \includegraphics[width=2.1cm]{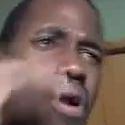} &  \includegraphics[width=2.1cm]{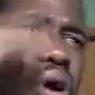} &
 \includegraphics[width=2.1cm]{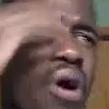} &   \includegraphics[width=2.1cm]{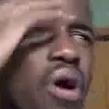} &
   \includegraphics[width=2.1cm]{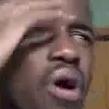} &   \includegraphics[width=2.1cm]{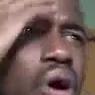} \\
\end{tabular}
\caption{Some challenging frames from the new database - Video 23}
\label{fig:face_challenge_database_1}
\end{figure}

\begin{figure}[!ht]
\begin{tabular}{cccccc}
  \includegraphics[width=2.1cm]{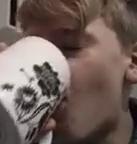} &  \includegraphics[width=2.1cm]{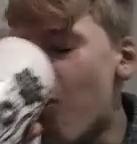} &
 \includegraphics[width=2.1cm]{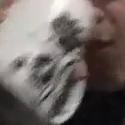} &   \includegraphics[width=2.1cm]{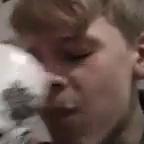} &
   \includegraphics[width=2.1cm]{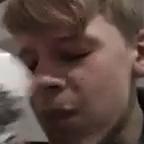} &   \includegraphics[width=2.1cm]{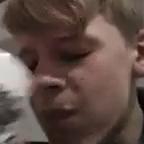} \\
\end{tabular}
\caption{Some challenging frames from the new database - Video 32}
\label{fig:face_challenge_database_2}
\end{figure}

\begin{figure}[!ht]
\begin{tabular}{cccccc}
  \includegraphics[width=2.1cm]{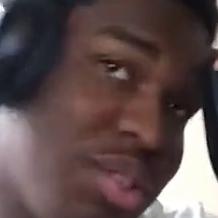} &  \includegraphics[width=2.1cm]{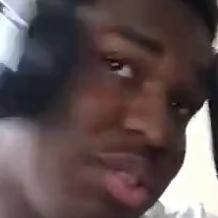} &
 \includegraphics[width=2.1cm]{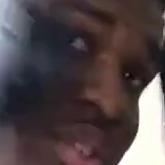} &   \includegraphics[width=2.1cm]{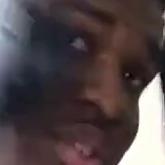} &
   \includegraphics[width=2.1cm]{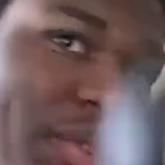} &   \includegraphics[width=2.1cm]{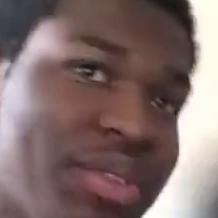} \\
\end{tabular}
\caption{Some challenging frames from the new database - Video 64}
\label{fig:face_challenge_database_3}
\end{figure}

\begin{table}[!ht]

\begin{minipage}{1\linewidth}
   
\begin{tabular}{cccccc}
    \includegraphics[width=2.1cm]{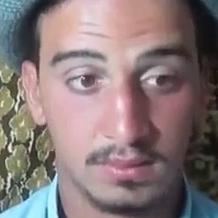} &  \includegraphics[width=2.1cm]{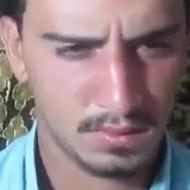} &
    \includegraphics[width=2.1cm]{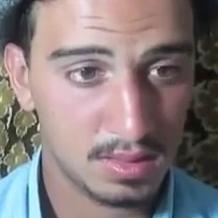} &
    \includegraphics[width=2.1cm]{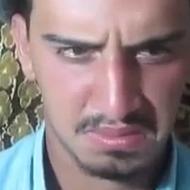} &
    \includegraphics[width=2.1cm]{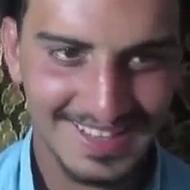} &   \includegraphics[width=2.1cm]{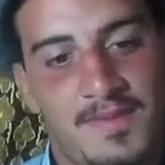} \\
    \footnotesize (a) & \footnotesize (b) & \footnotesize (c) & \footnotesize (d) & \footnotesize (e) & \footnotesize (f) \\
\end{tabular}
\caption{Frames from the video 55}
\label{fig:ex_ann_1}
\end{minipage}

	\begin{minipage}{0.5\linewidth}
		\centering

    \resizebox{220pt}{!}{
		\begin{tabular}[scale=0.5]{|l c c c c c c|}
			\hline
			Annotation & a & b & c & d & e & f \\
            \hline
            \rowcolor{gray!40}
            Valence & 0.14 & 0.17 & 0 & -0.50 & 0.38 & 0.04 \\
            \rowcolor{gray!10}
            Arousal & 0.20 & 0.24 & 0.44 & 0.61 & 0.35 & 0.01 \\
            \rowcolor{gray!40}
            AU 1 & x & x & & & & \\
            \rowcolor{gray!10}
            AU 2 & x & x & & & & \\
            \rowcolor{gray!40}
            AU 4 & & & & & & \\
            \rowcolor{gray!10}
            AU 6 & & & & & x & \\
            \rowcolor{gray!40}
            AU 12 & & & & & x & \\
            \rowcolor{gray!10}
            AU 15 & & & & x & & \\
            \rowcolor{gray!40}
            AU 20 & & & & & & \\
            \rowcolor{gray!10}
            AU 25 & & & & & x & x \\
            \hline
		\end{tabular}}

		\caption{Annotations for the video 55}
		\label{table:ann_1}
	\end{minipage}\hfill
	\begin{minipage}{0.45\linewidth}
		\centering
		\includegraphics[scale=0.5]{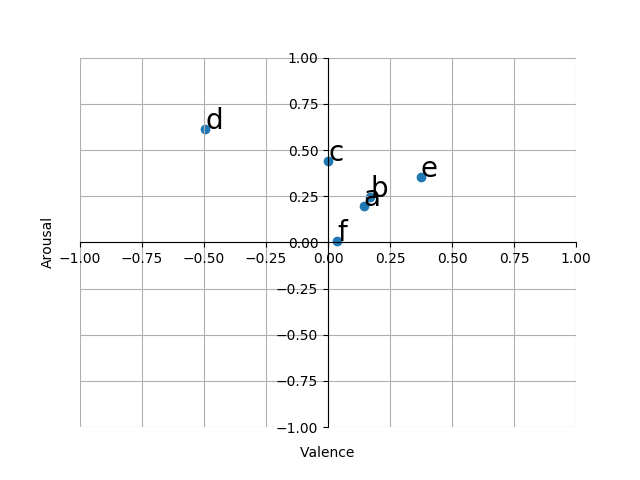}
		\captionof{figure}{Valence Arousal graph for the video 55}
		\label{fig:va_ann_1}
	\end{minipage} \hfill

\end{table}

\begin{table}[!ht]

\begin{minipage}{1\linewidth}
   
\begin{tabular}{cccccc}
  \includegraphics[width=2.1cm]{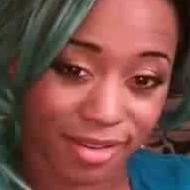} &  \includegraphics[width=2.1cm]{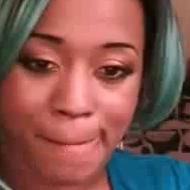} &
 \includegraphics[width=2.1cm]{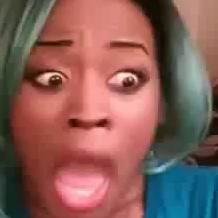} &   \includegraphics[width=2.1cm]{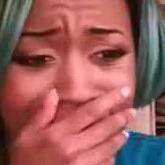} &
   \includegraphics[width=2.1cm]{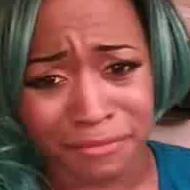} &   \includegraphics[width=2.1cm]{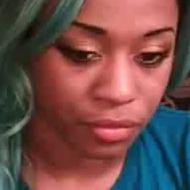} \\
    \footnotesize (a) & \footnotesize (b) & \footnotesize (c) & \footnotesize (d) & \footnotesize (e) & \footnotesize (f) \\
\end{tabular}
\caption{Frames from the video 20}
\label{fig:ex_ann_2}
\end{minipage}

	\begin{minipage}{0.5\linewidth}
		\centering

    \resizebox{220pt}{!}{
		\begin{tabular}[scale=0.5]{|l c c c c c c|}
			\hline
			Annotation & a & b & c & d & e & f \\
            \hline
            \rowcolor{gray!40}
            Valence & 0.02 & 0.07 & -0.5 & -0.54 & -0.52 & -0.40 \\
            \rowcolor{gray!10}
            Arousal & 0.14 & 0.13 & 0.68 & 0.52 & 0.71 & 0.24 \\
            \rowcolor{gray!40}
            AU 1 & & & x & & x & x \\
            \rowcolor{gray!10}
            AU 2 & & & & & & \\
            \rowcolor{gray!40}
            AU 4 & & & & x & x & \\
            \rowcolor{gray!10}
            AU 6 & & & & x & x & \\
            \rowcolor{gray!40}
            AU 12 & x & x & & & & \\
            \rowcolor{gray!10}
            AU 15 & & & & & & \\
            \rowcolor{gray!40}
            AU 20 & x & x & & & x & \\
            \rowcolor{gray!10}
            AU 25 & & & x & & & \\
            \hline
		\end{tabular}}

		\caption{Annotations for the video 20}
		\label{table:ann_2}
	\end{minipage}\hfill
	\begin{minipage}{0.45\linewidth}
		\centering
		\includegraphics[scale=0.5]{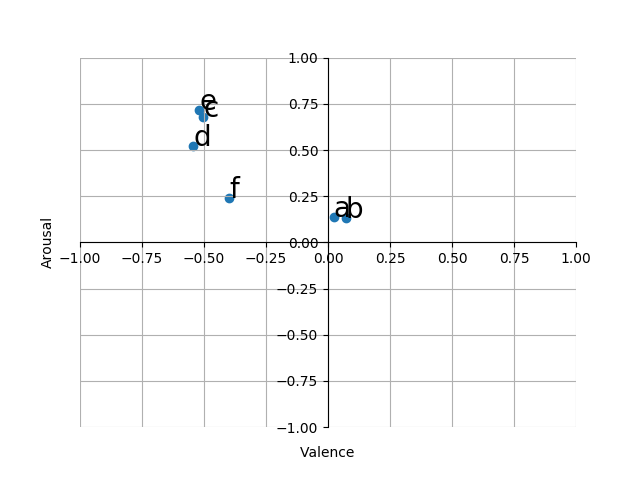}
		\captionof{figure}{Valence Arousal graph for the video 20}
		\label{fig:va_ann_2}
	\end{minipage} \hfill

\end{table}

To conclude the Figures \ref{fig:face_database_1} and \ref{fig:face_database_2} show some representatives of this newly created dataset inherited from the Aff-Wild dataset. The Figures \ref{fig:face_challenge_database_1}, \ref{fig:face_challenge_database_2} and \ref{fig:face_challenge_database_3} are examples of challenging images that the neural networks will have to analyze. Individuals can put their hands on their faces, or drink and so have a cup in front of their faces or wear a headset. Finally, the set of Table \ref{fig:ex_ann_1}, Table \ref{table:ann_1} and Figure \ref{fig:va_ann_1} and the set of \ref{fig:ex_ann_2}, Table \ref{table:ann_2} and Figure \ref{fig:va_ann_2} show examples of frames taken from videos with the corresponding annotations for Valence, Arousal and the Action Units.

\clearpage

\section{Building the network}

In the Section \ref{sec:selection_archi}, we saw different CNNs and LSTMs architectures based on well-known neural networks like VGG or ResNet stacked together and reaching top performance. The originality of this project is to try to reach the same performance by associating two existing models (Valence Arousal and Action Units) with a totally different architecture, i.e. a Generative Adversarial Network or GAN. A GAN can be seen as a two-player game because it is made up of two different neural networks. On the one hand, the first player, called the Generator, tries to create images similar to the images from the dataset. On the other hand, the second player, named the Discriminator, tries to make the difference between the real images from the dataset and the fake images created by the Generator. Currently GANs are also being used for helping network training by augmenting the training set; the results are very good and comparable with other computer vision methods \cite{kollias8,kollias9}. 

\subsection{Vanilla GAN} \label{sec:vanilla_gan}

To get more familiar with how GANs work, the project started with a vanilla GAN. This GAN has a basic architecture (Figure \ref{tab:vanilla_GAN_arch}). The Generator takes a vector of 100 random numbers between -1 and 1 and outputs a fake image of size 784*1 that is fed to the Discriminator. The Discriminator also takes real images as inputs. The images from the database are 28*28 and resized to 784*1 before entering the Discriminator. The Discriminator outputs the probability for the image to be real or fake. The weights of the neural networks for the Discriminator and the Generator are initialized with Xavier Initialization : the starting weights are random numbers generated from a normal distribution with standard deviation at $\frac{1}{\sqrt{\frac{input\text{ }dimension}{2}}}$ and mean at 0. The biases of the neural networks are initialized at 0.

\begin{table}[!ht]
    \centering
\begin{tabular}{|c|c|c|}\hline
    \textbf{Layers} & \textbf{Generator} & \textbf{Discriminator} \\ 
    \hline
    \rowcolor{gray!40}
    Hidden Layer & \pbox{2cm}{128 nodes \\ ReLU} & \pbox{2cm}{128 nodes \\ ReLU} \\ [10pt]
    \rowcolor{gray!10}
    Output Layer & \pbox{2cm}{784 nodes \\ Sigmoid} & \pbox{2cm}{1 node \\ Sigmoid} \\ [10pt]
   \hline
\end{tabular}
    \caption{Architecture of the vanilla GAN}
    \label{tab:vanilla_GAN_arch}
\end{table}

At each iteration, the loss of the Discriminator and the Generator are computed. The loss of the Discriminator is the addition of the loss for real images and the loss for fake images. The Discriminator loss for real or fake images is the mean of the cross entropies between the sigmoid of the groundtruth labels and the outputs of the Discriminator (which is the sigmoid of the node of the last layer) computed on the batch (by default, the batch size is 128). The groundtruth label is 1 if the image is real and 0 if it is fake. The Generator loss is the mean of the cross entropies between the sigmoid of the groundtruth labels and the outputs of the Discriminator for the fake images computed on the batch. Then, backpropagation is performed via the default Adam optimizer ($learning\text{ }rate = 0.001$, $\beta_1 = 0.9$, $beta_2 = 0.999$ and $epsilon = 10^{-8}$). 

\begin{figure}
\begin{tabular}{cccc}
  \includegraphics[scale=0.08]{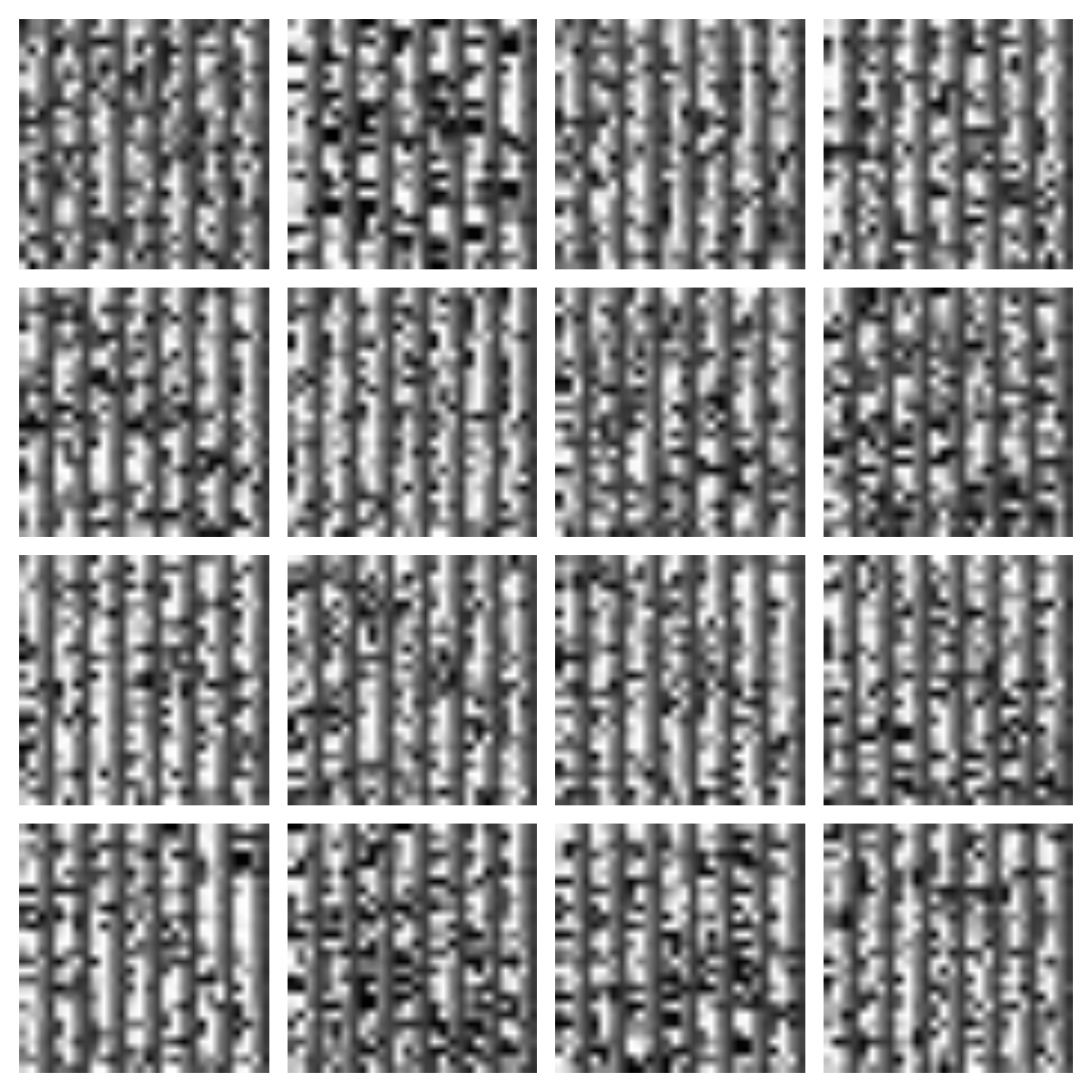} &   \includegraphics[scale=0.08]{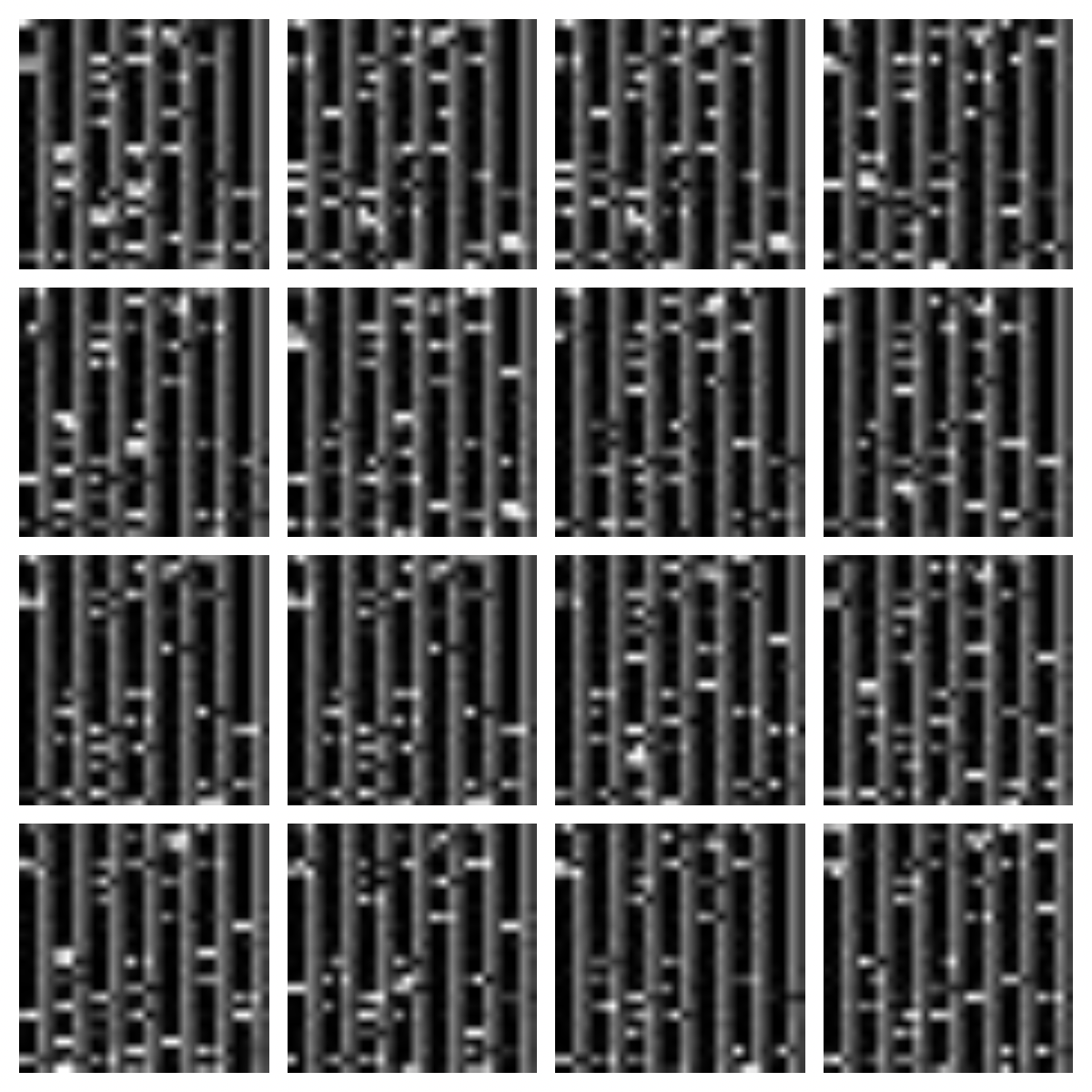} &
 \includegraphics[scale=0.08]{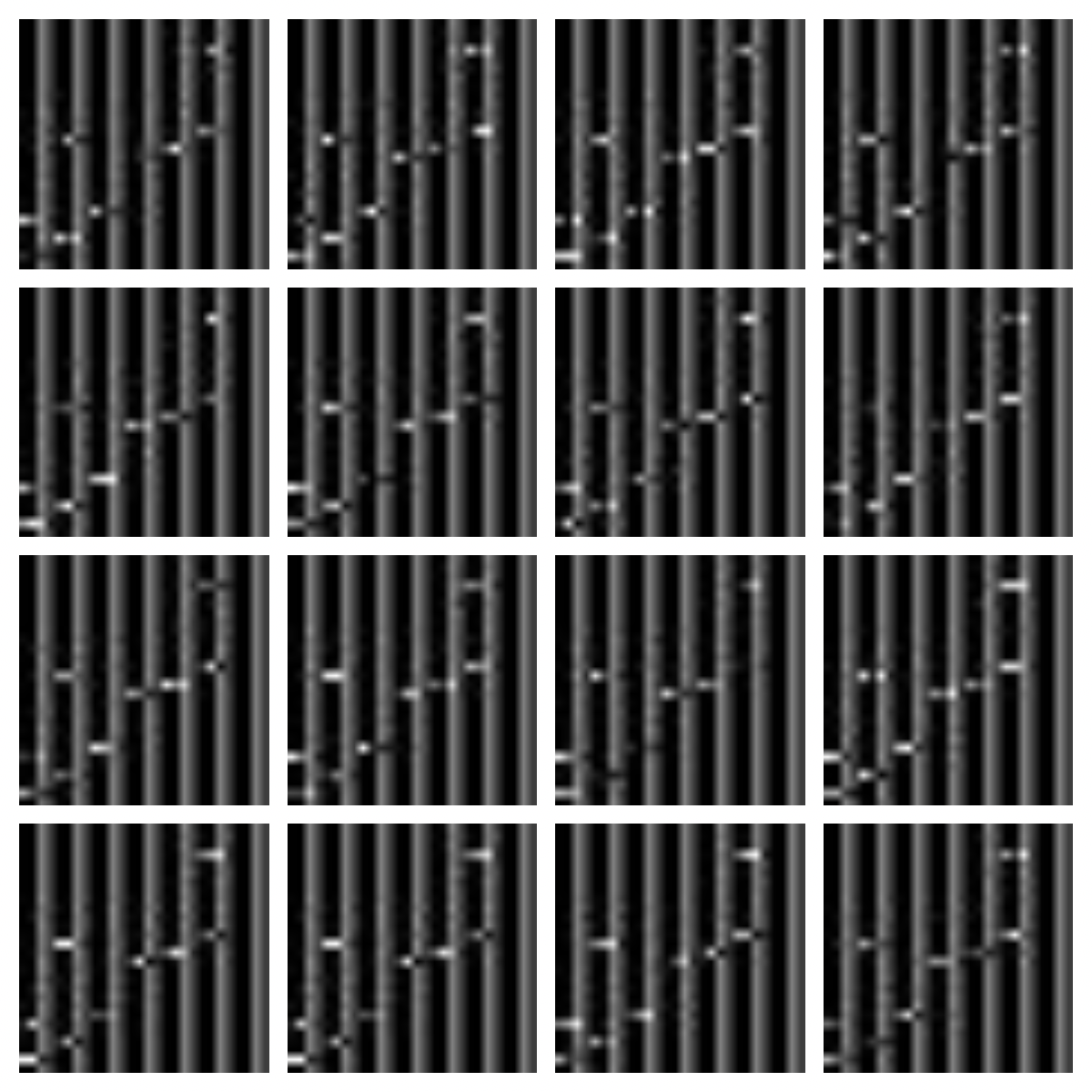} &   \includegraphics[scale=0.08]{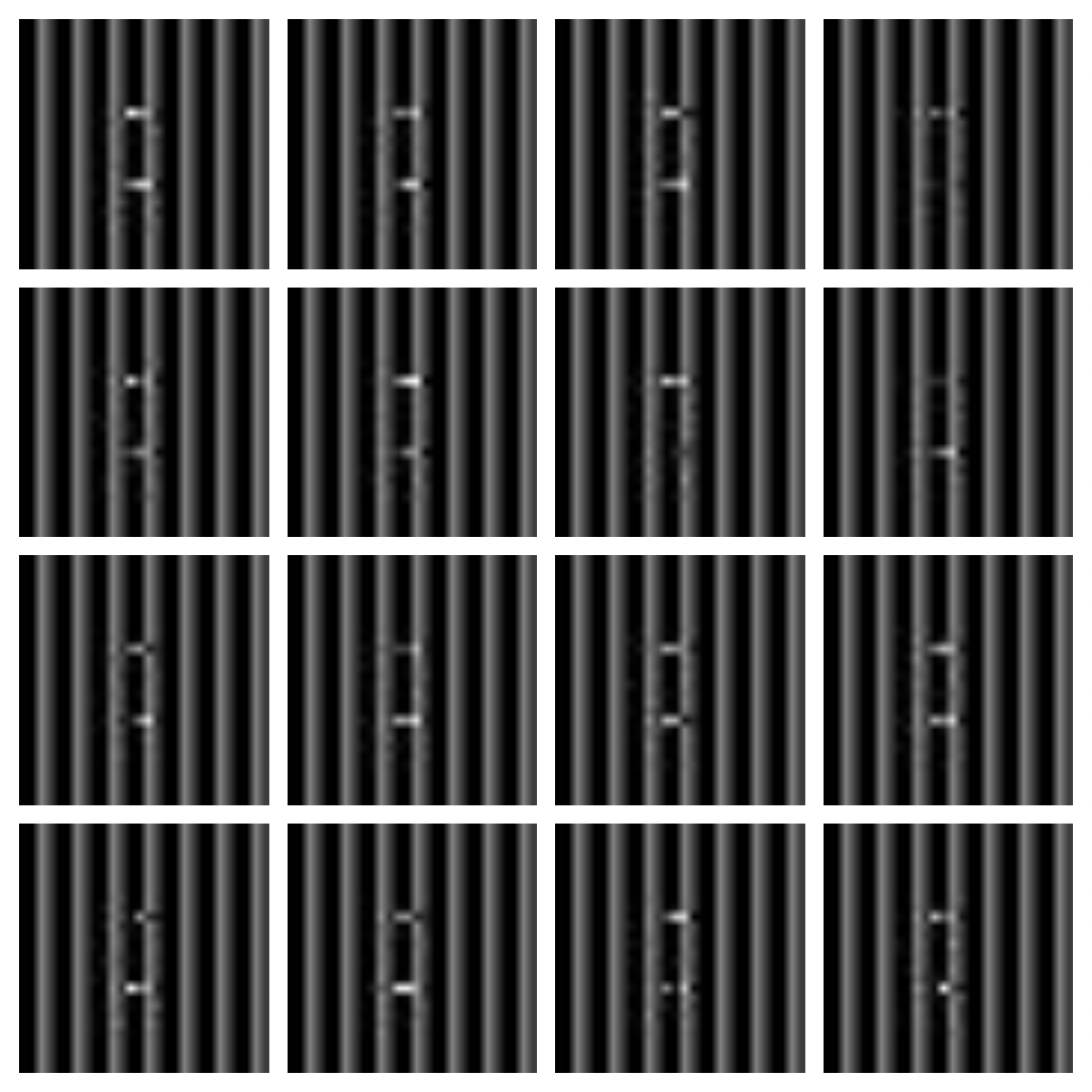} 
\end{tabular}
\caption{Images created by the Generator at iteration 0, 20, 50 and 99 (from left to right) for the VA + AU dataset}
\label{fig:gen_vanilla_dataset}
\end{figure}

\begin{figure}[!ht]
\centering
\centerline{
   \begin{minipage}{.5\linewidth}
   \centering
      \includegraphics[scale=0.5]{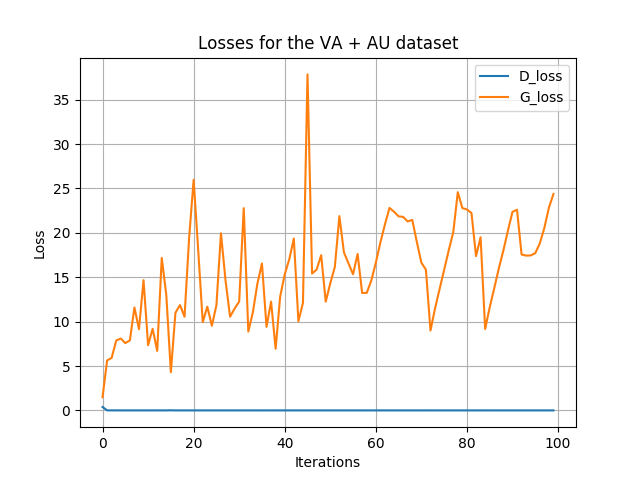}
      \caption{Losses of the Generator and the Discriminator for the VA + AU dataset}
      \label{fig:vanilla_losses_data}
   \end{minipage} \hfill
   \begin{minipage}{.5\linewidth}
   \centering
      \includegraphics[scale=0.5]{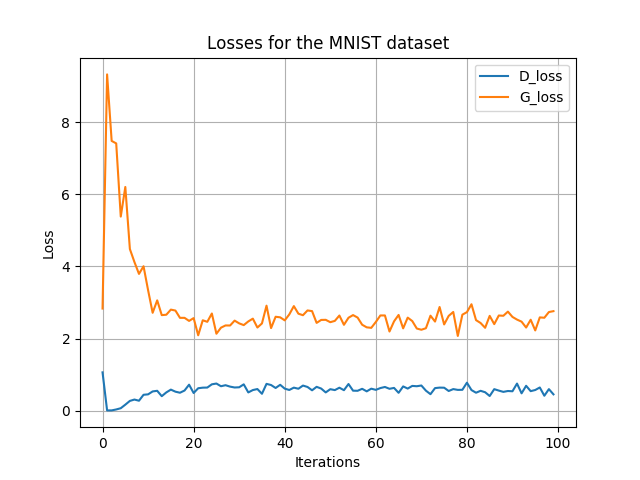}
      \caption{Losses of the Generator and the Discriminator for the MNIST dataset}
      \label{fig:vanilla_losses_mnist}
   \end{minipage} \hfill
}
\end{figure}

The images obtained with this very simple vanilla GAN are shown in Figure \ref{fig:gen_vanilla_dataset}. We can see that progressively the generator erases the noise and produces blacker images keeping gray straight lines. This phenomenon of generating always the same images is collapse mode. This can happen when the neural network is too simple to get the complexity of the data. In this case, it is likely as the generator and the discriminator only have two layers and need to remember faces from a large dataset. This confirmed by the Figure \ref{fig:vanilla_losses_data} : the Discriminator loss quickly tends to zero while the Generator loss steadily increases. The Discriminator gets too good too quickly to enable the Generator to create faces and it cannot learn as the Discriminator always classify the images it produces as fake.

\begin{figure}
\begin{tabular}{cccc}
  \includegraphics[scale=0.4]{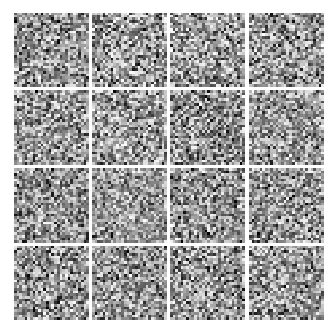} &   \includegraphics[scale=0.4]{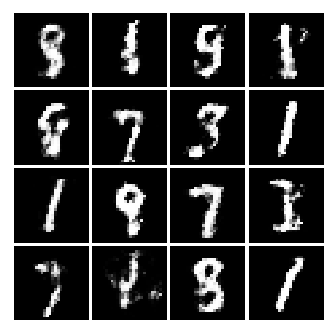} &
 \includegraphics[scale=0.4]{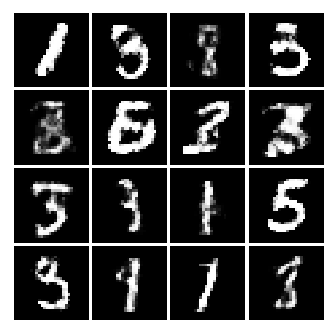} &   \includegraphics[scale=0.4]{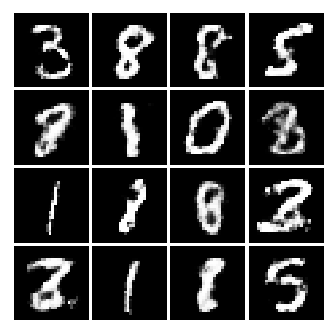}
\end{tabular}
\caption{Images created by the Generator at iteration 0, 20, 50 and 99 (from left to right) for the MNIST dataset}
\label{fig:gen_vanilla_mnist}
\end{figure}

As we can see on Figure \ref{fig:gen_vanilla_mnist}, the Generator is able to generate digits when it is trained on the MNIST dataset. It means that for a much simpler dataset like the MNIST consisting of digits in grayscale images, the GAN is suitable and performs well at generation from noise. Unlike the losses for the VA + AU dataset, the losses for the MNIST dataset converge to steady values : a little above 2 for the Generator and a little under 1 for the Discriminator. However, we can see that on a set of 16 images, a lot of 1 are produced which shows how susceptible the generator can enter the collapse mode and only produce 1.

\newpage

\subsection{Categorical GANs}

The project is about combining the Valence Arousal model and the Action Units model in order to perform better results than one model alone. The architecture tested is a Generative Adversarial Network or GAN. A Generative Adversarial Network or GAN consists of two networks competing against each other. On the one hand, the Discriminator tries to determine if the image taken as an input is produced by the Generator or is an image from the database. The image produced by the Generator is considered as fake while the image from the database is considered as real. On the other hand, the Generator tries to deceive the Discriminator by trying to counterfeit real images. \\
For this project, we chose to study a particular type of GAN : a categorical GAN. The paper "Semi-supervised learning with generative adversarial networks" \cite{RefWorks:doc:5af1a0ffe4b0ac09013ce59e} introduces the concept of categorical GAN. In a categorical GAN, the Discriminator is also a Classifier : the Discriminator has to determine if the input image is real or fake and also has to determine the category the image belongs to. For example a categorical GAN can be used on the MNIST dataset : the Discriminator tries to determine if the digit image is from the dataset or created by the Generator. Moreover the Discriminator has to classify the image into one of the 10 categories corresponding to the 10 digits. \\
The idea beyond categorical GAN is that the different categories the Discriminator has to infer will help the Generator produce better images and the better images produced will help the Discriminator classify more accurately.

\subsubsection{Presentation of the existing architecture}

The code I implemented for the project is inherited from an existing code \cite{RefWorks:doc:5b8924f2e4b036495fd77f4d} which implements the paper I talked about above \cite{RefWorks:doc:5af1a0ffe4b0ac09013ce59e}. The original code uses categorical GAN with three datasets : 
\begin{itemize}
    \item MNIST : a database containing 60,000 training images and 10,000 testing images of handwritten digits;
    \item SVHN : a dataset with 600,000 images of Street View House Numbers;
    \item CIFAR10 : a collection of 60,000 images with 10 different classes : airplanes, cars, birds, cats, deer, dogs, frogs, horses, ships and trucks. 
\end{itemize}

The architecture of the GAN is made up of a Generator and a Discriminator. \\
The Discriminator takes as an input the image and then applies the following architecture presented in Table \ref{tab:archi_orig_disc}. The first three layers consists of a convolutional layer (explained in \ref{sec:nn_types}). The first layer can be read as follow : 
\begin{itemize}
    \item The first convolutional layer ("conv1") whose filter is [5,5,3,64] with a stride [1,2,2,1] and a padding 'SAME' consists of 64 filters of size [5,5,3] where 3 is the number of channels (3 because the images are RGB). Each filter can then be broken down in 3 where each subfilter is a 2D matrix of size [5*5] that applies on a channel. These filters then slides through the image of size 28*28 in our case with a stride of [1,2,2,1]. The stride represents how many pixels the filter (= the 5*5 matrix) will jump before applying its convolution. In this case it will jump 2 pixels horizontally and 2 pixels vertically. The padding 'SAME' means that extra zeros will be added to the edges of the original image so that the filter can be applied an even number of times. This convolutional layer outputs 64 new images (with 3 channels for each image).
    \item The "lReLU" applies the leaky Rectified Linear Unit on each pixel of the images. This function is $0.54x +0.4|x|$, where $x$ is the value of the pixel.
    \item The "batch norm" computes the mean and the variance of the images over the batch (by default the batch size is 64) and then normalizes the outputs. 
    \item The "dropout" takes each output of the previous layer and initializes them back to zero with a probability of 0.5.
    \item The next layer takes the output of the previous one and applies the same transformations.
\end{itemize}
The last layer consists of :
\begin{itemize}
    \item A Fully-Connected layer (described in \ref{sec:nn_types}). This layer takes as an input a flattened image of 256*1 and outputs n+1 nodes, where n is the number of Action Units. The last node is used for determining if the image is fake or real.
    \item Then a softmax function is applied :
    \begin{align}
        \sigma(z)_j = \frac{e^{z_j}}{\sum_{k=1}^K{e^{z^k}}}
    \end{align}
    This function takes a vector z of K components (here the n+1 nodes) and outputs a vector of number between 0 and 1 whose sum of components is equal to 1. Thus, the output of the Discriminator is a vector of probability of the image to belong to one of the category or to be fake.
\end{itemize}

\begin{table}[!ht]

    \centering
    \begin{tabular}{|c|c|c|c|c|c|c|c|}
        \hline
        \textbf{Layer} & \textbf{Filter} & \textbf{Stride} & \textbf{Padding} & \pbox{5cm}{\textbf{Keeping} \\ \textbf{probability}} & \textbf{Number of units} \\
        \hline
        \rowcolor{gray!40}
        conv1 & [5,5,3,64] & [1,2,2,1] & SAME & & \\
        \rowcolor{gray!40}
        lReLU & & & & & \\
        \rowcolor{gray!40}
        batch norm & & & & & \\
        \rowcolor{gray!40}
        dropout & & & & 0.5 & \\
        \hline
        \rowcolor{gray!10}
        conv2 & [5,5,64,128] & [1,2,2,1] & SAME & & \\
        \rowcolor{gray!10}
        lReLU & & & & & \\
        \rowcolor{gray!10}
        batch norm & & & & & \\
        \rowcolor{gray!10}
        dropout & & & & 0.5 & \\
        
        \hline
        \rowcolor{gray!40}
        conv3 & [5,5,128,256] & [1,2,2,1] & SAME & & \\
        \rowcolor{gray!40}
        lReLU & & & & & \\
        \rowcolor{gray!40}
        batch norm & & & & & \\
        \rowcolor{gray!40}
        dropout & & & & 0.5 & \\
        
        \hline
        \rowcolor{gray!10}
        FC & & & & & n+1 \\
        \rowcolor{gray!10}
        Softmax & & & & & \\
        \hline
        
    \end{tabular}
    \caption{Original Architecture for the Discriminator of the categorical GAN}
    \label{tab:archi_orig_disc}
\end{table}

\begin{figure}[!ht]
    \centering
    \includegraphics[scale=0.5]{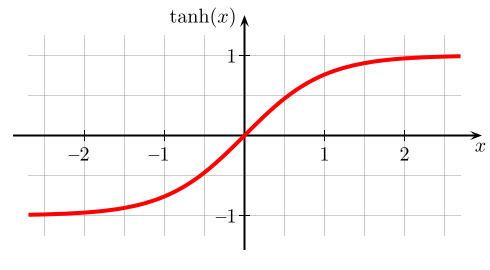}
    \caption{Hyperbolic tangent graph}
    \label{fig:tanh}
\end{figure}

The Generator takes as an input a random vector of size 100 and applies the architecture described in the Table \ref{tab:archi_orig_gen}. The three lines are made up of :
\begin{itemize}
    \item A deconvolutional layer or more accurately named a transposed convolutional layer (explained in \ref{sec:nn_types}). This first layer takes the random vector of 100 numbers between -1 and 1 and outputs a vector of size 384*1 by a applying a filter of size [2,2] (only one channel) and a stride of [1,1]. The 'VALID' padding means that some pixels of the images will be dropped so that the filter can be applied an even number of times on the image.
    \item Then, the leaky Rectified Linear Unit is applied.
    \item Finally, a batch normalization is applied.
\end{itemize}
The last layer also applies the hyperbolic tangent activation function whose plot can be seen in Figure \ref{fig:tanh}.

\begin{table}[!ht]

    \centering
    {\begin{tabular}{|c|c|c|c|c|c|c|c|}
        \hline
        \textbf{Layer} & \textbf{Filter} & \textbf{Stride} & \textbf{Padding} \\
        \hline
        \rowcolor{gray!40}
        deconv1 &  [2,2,100,384] & [1,1,1,1] & VALID \\
        \rowcolor{gray!40}
        lrelu & & & \\
        \rowcolor{gray!40}
        batch norm & & & \\
        
        \hline
        \rowcolor{gray!10}
        deconv2 & [4,4,384,128] & [1,2,2,1] & VALID \\
        \rowcolor{gray!10}
        lrelu & & & \\
        \rowcolor{gray!10}
        batch norm & & & \\
        
        \hline
        \rowcolor{gray!40}
        deconv3 & [4,4,128,64] & [1,2,2,1] & VALID \\
        \rowcolor{gray!40}
        lrelu & & & \\
        \rowcolor{gray!40}
        batch norm & & & \\
        
        \hline
        \rowcolor{gray!10}
        deconv4 & [6,6,64,3] & [1,2,2,1] & VALID \\
        \rowcolor{gray!10}
        tanh & & & \\
        \hline
        
    \end{tabular}}
    \caption{Original Architecture for the Generator of the categorical GAN}
    \label{tab:archi_orig_gen}
\end{table}

\subsubsection{Training process for the original categorical GAN}

During one iteration of the training, the following steps happen :
\begin{itemize}
    \item The Generator takes a vector of random numbers-noise \cite{raftopoulos2018beneficial} between -1 and 1 and creates a fake image. It repeats this operation as many times as the batch size;
    \item The Discriminator takes the fake images as an input and determines if they are fake or real and classify them into one of the category;
    \item The Discriminator does the same for a batch of real images from the training set;
    \item Then the loss functions of the Generator and the Discriminator are computed;
    \item The losses computed are used to back-propagate the error through the parameters of the Generator and the Discriminator. This optimization is different for the Generator and the Discriminator. The Discriminator parameters are updated at a frequency determined by the update rate. The Discriminator is updated for every multiple of $update\_rate + 1$. On the contrary the Generator is updated every iteration except at for the multiples of $update\_rate + 1$. By default, this update rate is 5. It means that when the update rate increases, the Discriminator parameters are updated less frequently while the Generator parameters are updated more frequently. \\
    The optimizer used for both the Generator and the Discriminator is the Adam optimizer (with  $\beta = 0.5$). The learning rates for the Discriminator is half the learning rate for the Generator. By default, the learning rate is $10^{-4}$. The gradients are clipped under 20 to avoid the problem of exploding gradients that can happen during training.
\end{itemize}

The loss of the GAN (\textit{GAN\_loss} in Table \ref{code:orig_loss}) consists of the loss of its two neural networks :
\begin{itemize}
    \item The Discriminator loss is divided between the loss for real and fake images (\textit{d\_loss\_real} in Table \ref{code:orig_loss}). These losses are the softmax cross entropy between the labels generated by the Discriminator and the groundtruth labels. The labels generated by the Discriminator consists of the n nodes corresponding to the n categories (in the original datasets n=10) and the last node indicating if the image is fake (1) or real (0). The groundtruth labels for real images are well-known by definition : in the original categorical GAN, there are a one-hot encoded vector with the $i^{th}$ component being 1 if the image belongs to the $i^{th}$ category (one-hot encoding), concatenated with a vector with the value 0 indicating the image is real (\textit{real\_label} in the Table \ref{code:orig_loss}). For fake images, the groundtruth labels is also a n+1-vector with the n first values being $\frac{1-\alpha}{n}$ and the last value being $\alpha$ (\textit{fake\_label} in the Table \ref{code:orig_loss}). By default, $\alpha$ is 0.9. Indeed, these labels can only be guessed and this is the reason why this deep learning approach is named semi-supervised : the labels produced by the Generator can only be guessed. (However there are similarity techniques to put labels on the generated images but there were not used in this project). \\
    The loss of the Discriminator is the mean of the loss for real images and the loss for fake images (\textit{d\_loss} in Table \ref{code:orig_loss}).
    \item The Generator loss (\textit{g\_loss} \ref{code:orig_loss}) is the addition of the mean of the logarithm of the fake labels generated by the Discriminator and the Huber loss (\ref{sec:cost_fcts}) between the real and fake images, multiplied by a coefficient. If the step is under 1500, this coefficient is $\frac{1500-step}{1500}*10$, else it is 0. 
\end{itemize}

\begin{table}[!ht]
        \includegraphics[width=1.\linewidth]{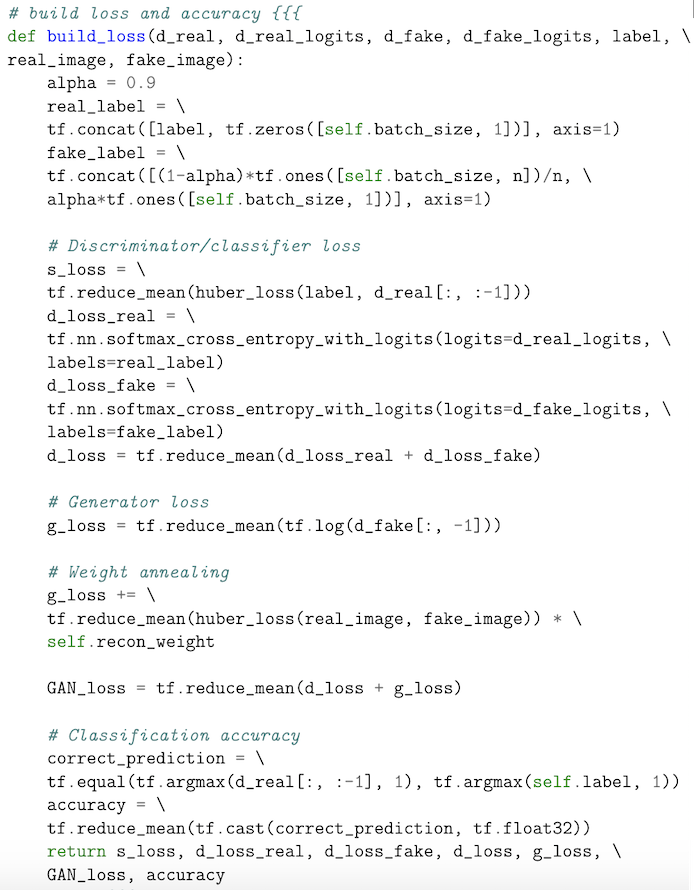}
\caption{Original loss function}
\label{code:orig_loss}
\end{table}

This existing architecture was then adapted to the project : the model was transformed to suit a non-exclusive classification and regression problem and no more an exclusive classification problem. Indeed, the original architecture was made to predict the digit on an image from the MNIST, SVHN or CIFAR10 dataset : either 0, 1, 2, 3, 4, 5, 6, 7, 8 or 9. This is an exclusive classification problem. In our project, we want to predict the presence ($label=1$) or the absence ($label=0$) of an Action Unit on a image and the values of Valence and Arousal. The prediction of the presence or the absence of an Action Unit is a non-exclusive classification problem because several Action Units can be present on the same image. The prediction of the values of Valence and Arousal is a regression problem because these values range from -1 to 1. \\
In order to address this problem progressively, the original architecture was tuned for Action Units only, then for Valence and Arousal only and finally the two prediction models were merged with more fine tuning performed. Only the last layer of the Generator and Discriminator architectures were adapted to suit the AU and VA model. In order to have the best results, losses were changed and different values of parameters like the learning rate and the update rate were tried. New metrics were also implemented so as to be relevant to the new models.
\\
\\
After understanding and playing a bit with this categorical GAN, the adaptation of the architecture for the different models begun. We first adapted the architecture for Action Units model only, then for the Valence Arousal model only, and then for both models. This way of doing made it easier to understand the influence of the choices made.

\subsection{GANs for Action Units}

\subsubsection{Model customization}

The first part of the project was to adapt the code for Action Units. \\
\\
First, the loading of the images and the labels had to be rewritten. The original code uses a special format to store the images and the labels (format HDF5). This format enables faster reading of huge amount of data. To create this file containing all the images and the corresponding labels, the images and labels were loaded from the files created during the creation of the databse, i.e. .jpg files for the images and .txt files for the labels. Then the images were cropped and resized to the suited size (28*28). Finally, the images and labels were converted to the HDF5 thanks to the h5py package in Python. 
\\
\\
The Generator architecture has not been changed : it still takes a vector of noise (size 100 with values between -1 and 1) as an input and outputs an image of size 28*28. \\
Only the final layer of the Discriminator architecture has been changed to adapt to the non-exclusive classification problem. The last layer of the Discriminator is still a fully-connected layer of n+1 nodes, where n is the number of Action Units, i.e. 8. Finally, instead of applying the softmax function to the last layer, the sigmoid function is applied (see below). Indeed, while the softmax function takes the value of the nodes and computes the probability of the nodes based on the values of the others and so is suitable for an exclusive classification problem by giving the most probable node and so the most probable category, the sigmoid function computes a probability for each node independently and so gives the probability for each Action Unit (each corresponding to a node) to be absent or present and the probability of the image of being real or fake for the last node.\\
Here is the sigmoid function :
\begin{align}
    S(x) = \frac{1}{1 + e^{-x}} = \frac{e^x}{e^x + 1}
\end{align}
\\
\\
The loss functions that are used for computing backpropagation has also been adapted to the Action Unit task. \\
The Discriminator loss is still the mean of the loss of the Discriminator for real images and the loss of the Discriminator for fake images (\textit{d\_loss} in Table \ref{code:au_loss_1}). Instead of taking the softmax cross entropy to compute these losses, we take the the sigmoid cross entropy for the same reason as explained previously. This sigmoid cross entropy loss is computed between the labels generated by the Discriminator and the groundtruth labels. The labels generated by the Discriminator is composed of the n nodes corresponding to the n Action Units (in our dataset n=8) and the last node indicating if the image is fake (1) or real (0). The groundtruth labels for real images are well-known by definition : in this modified GAN, there are a 8-long vector with the $i^{th}$ component at 1 if the Action Unit is present and at 0 if it is absent, concatenated with a vector with the value 0 indicating the image is real (\textit{real\_label} in Table \ref{code:au_loss_1}). For fake images, the groundtruth labels is a n+1-vector with the n first values being $\frac{1-\alpha}{n}$ and the last value being $\alpha$ (\textit{fake\_label} in Table \ref{code:au_loss_1}). By default, $\alpha$ is 0.9. Indeed, these labels can only be guessed and this is the reason why this GAN is semi-supervised. For the loss of the Discriminator for real images and fake images, the mean of the sigmoid cross entropies is computed (\textit{d\_loss\_real} and \textit{d\_loss\_fake} in Table \ref{code:au_loss_1}). \\
The Generator loss is the same as the original architecture (\textit{g\_loss} in Table \ref{code:au_loss_2}).

\begin{table}[]
        \includegraphics[width=1.\linewidth]{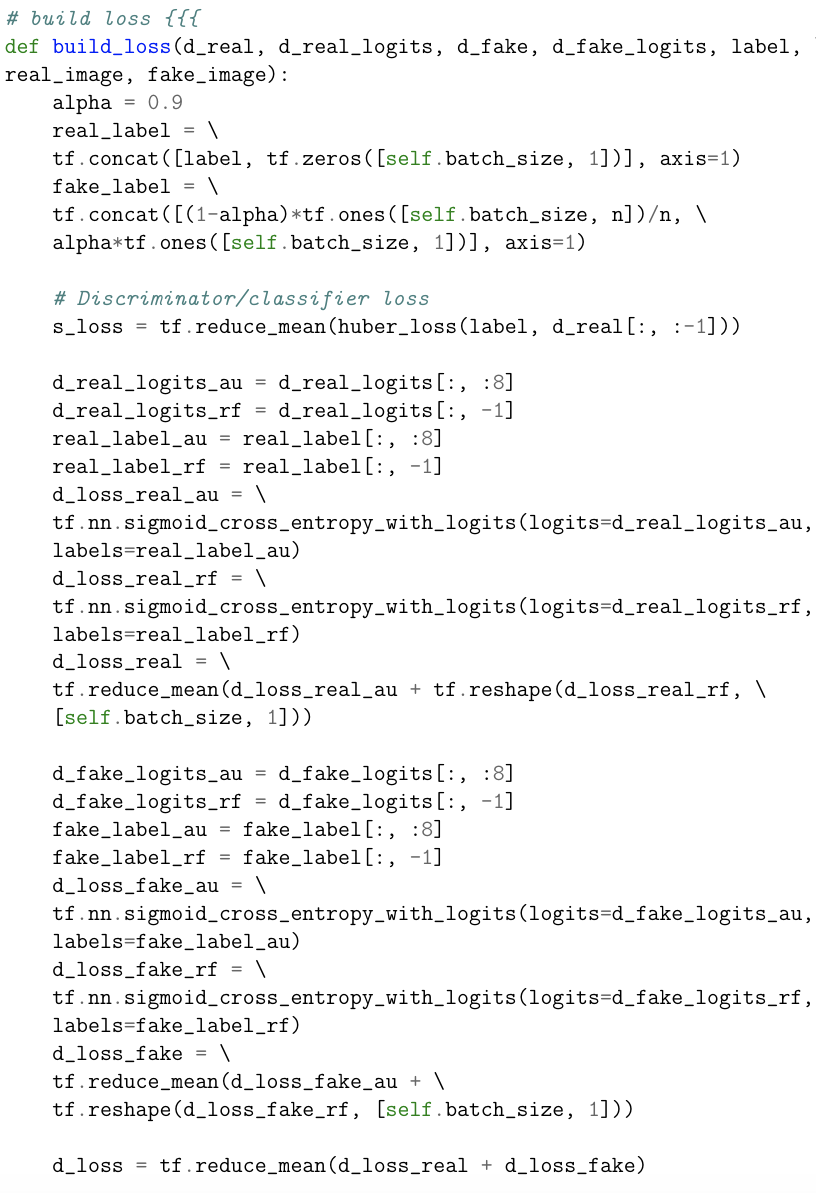}
\caption{Loss function for the AU GAN - Part 1}
\label{code:au_loss_1}
\end{table}

\begin{table}
        \includegraphics[width=1.\linewidth]{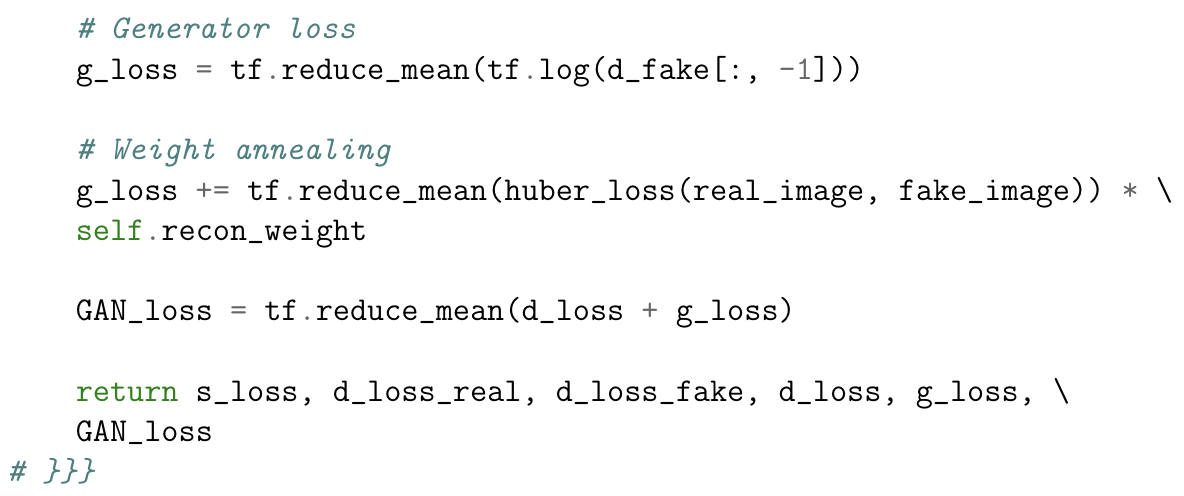}
\caption{Loss function for the AU GAN - Part 2}
\label{code:au_loss_2}
\end{table}

\subsubsection{Training \& Evaluation process}

For the training, 1,000,000 iterations were performed. Every 1,000 iterations the weights of the GAN were saved. Different metrics have been chosen to follow the training :
\begin{itemize}
    \item The precision for each Action Unit :
    \begin{align}
     precision = \frac{true\text{ }positives}{selected\text{ }elements}
    \end{align}
    where the $selected\text{ }elements$ is made up of the sum of $true\text{ }positives$ and $false\text{ }positives$ (see Figure \ref{fig:precision_recall})
    \item The recall for each Action Unit :
    \begin{align}
    recall = \frac{true\text{ }positives}{relevant\text{ }elements}
    \end{align}
     where the $relevant\text{ }elements$ is made up of the sume of $true\text{ }positives$ and $false\text{ }negatives$ (see Figure \ref{fig:precision_recall})
    \item The F1 score for each Action Unit where 
    \begin{align}
        F_1 = \frac{2}{\frac{1}{recall}+\frac{1}{precision}} = 2 \frac{precision.recall}{precision+recall}
    \end{align}
    \item The accuracy for each Action Unit, where 
    \begin{align}
        accuracy = \frac{true\text{ }positives + true\text{ }negatives}{true\text{ }positives + false\text{ }positives + true\text{ }negatives + false\text{ }negatives}
    \end{align}
    
    \begin{figure}
        \centering
        \includegraphics[scale=0.4]{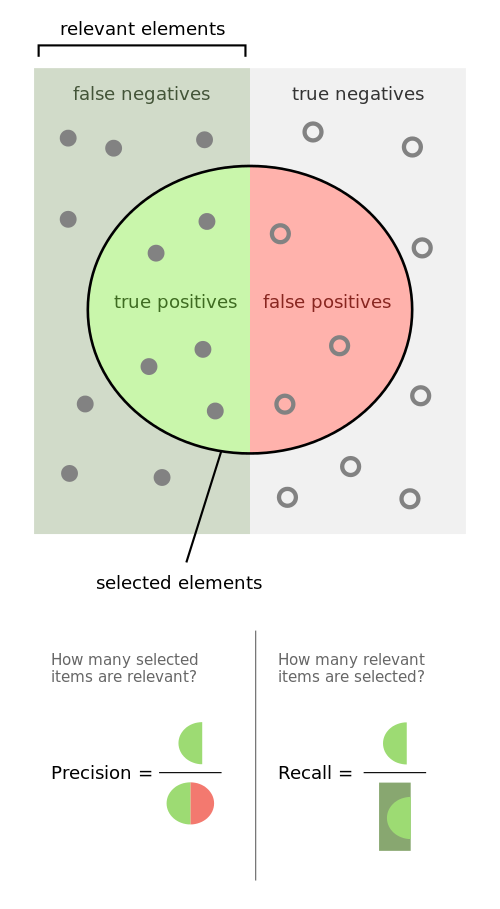}
        \caption{Precision and Recall}
        \label{fig:precision_recall}
    \end{figure}
\end{itemize}
We chose to have metrics for each specific Action Units so as to better understand how the GAN trained. Moreover, only an average on all Action Units of this metrics would not have given insights on the learning process because the dataset is unbalanced between Action Units and therefore it is easier for the neural networks to learn for most represented Action Units. Having the precision and recall helped better understand the F1 score which is computed thanks to these two metrics. The accuracy is a more general metric including all the basic elements of the classification (true positives, true negatives, false positives, false negatives).  
Moreover other indicators have been chosen so as to understand how the GAN trains and learns over time :
\begin{itemize}
    \item The loss of the Discriminator;
    \item The loss of the Generator;
    \item The gradients of the parameters of the neural networks;
    \item The images created by the Generator.
\end{itemize}
All these variables were visible in real-time thanks to TensorBoard, an auxiliary tool of TensorFlow \cite{RefWorks:doc:5b8e9cc0e4b0ad32cc6389f3}. It was possible to see these variables at every iteration for both the training set and the testing set because at every iteration one batch of the training set and one batch of the testing set were evaluated.
\\
\\
For the GAN customized for Action Units, the set of hyperparameters chosen is :
\begin{itemize}
    \item The learning rate : $10^{-4}$, $10^{-5}$;
    \item The update rate : $2$, $5$, $7$.
\end{itemize}

\subsubsection{Results}

After the 1,000,000 iterations of training, the GAN with its architecture and its chosen hyperparameters is evaluated. During the training, different iterations of the model have been saved. Indeed, during training, the parameters or weights of the Discriminator and the Generator evolved to perform the task of classifying and generating images in a way that minimizes their own loss function. The goal of evaluation is to determine the best scores for the chosen metrics (described after) on all the saved iterations of the model. The chosen metrics for evaluating the model on the testing set are :
\begin{itemize}
    \item The best F1 score for each Action Unit;
    \item The best mean F1 score computed on all Action Units;
    \item The best mean accuracy computed on all Action Units;
    \item The best mean between the first two metrics;
    \item The best score for the number of real images classified as real.  
\end{itemize}
The F1 score metric has been chosen because it gives the best insight on the data as it allies precision and recall.

Here are the results for the different models and metrics (the first line is the score, the second line in italics with parenthesis is the iteration in thousands at which this score has been reached, figures in bold are the best scores for all models): \\

\begin{table}[!ht]

    \centering
    {\begin{tabular}{|c|c c c|c|}
        \hline
        \multirow{2}{*}{\textbf{Model}} & \multicolumn{3}{c|}{\textbf{Average on all Action Units}} & \pbox{5cm}{\textbf{\% of real images} \\ \textbf{classified as real}}\\
        & F1 score & Accuracy & Mean & \\
        \hline
        \rowcolor{gray!40}
        \pbox{5cm}{learning rate : $10^{-4}$ \\ update rate : $2$} & \pbox{5cm}{0.707 \\ \small{\textit{(931)}}} &  \pbox{5cm}{0.893 \\ \small{\textit{(970)}}} & 
        \pbox{5cm}{0.797 \\ \small{\textit{(964)}}} &  \pbox{5cm}{0.979 \\ \small{\textit{(3)}}} \\
        
        \rowcolor{gray!10}
        \pbox{5cm}{learning rate : $10^{-4}$ \\ update rate : $5$} & \pbox{5cm}{0.716 \\ \small{\textit{(998)}}} & 
        \pbox{5cm}{0.899 \\ \small{\textit{(999)}}} & 
        \pbox{5cm}{0.807 \\ \small{\textit{(998)}}} & 
        \pbox{5cm}{0.993 \\ \small{\textit{(0)}}} \\
        \rowcolor{gray!40}
        \pbox{5cm}{learning rate : $10^{-4}$ \\ update rate : $7$} & \pbox{5cm}{0.783 \\ \small{\textit{(996)}}} & 
        \pbox{5cm}{0.920 \\ \small{\textit{(995)}}} & 
        \pbox{5cm}{0.851 \\ \small{\textit{(996)}}} & \pbox{5cm}{\textbf{0.996} \\ \small{\textit{(7)}}} \\
        
        \hline
        \rowcolor{gray!10}
        \pbox{5cm}{learning rate : $10^{-5}$ \\ update rate : $2$} & \pbox{5cm}{0.698 \\ \small{\textit{(790)}}} & 
        \pbox{5cm}{0.892 \\ \small{\textit{(887)}}} & 
        \pbox{5cm}{0.794 \\ \small{\textit{(910)}}} & 
        \pbox{5cm}{0.990 \\ \small{\textit{(4)}}} \\
        \rowcolor{gray!40}
        \pbox{5cm}{learning rate : $10^{-5}$ \\ update rate : $5$} & \pbox{5cm}{0.767 \\ \small{\textit{(995)}}} & 
        \pbox{5cm}{0.915 \\ \small{\textit{(991)}}} & 
        \pbox{5cm}{0.841 \\ \small{\textit{(995)}}} & \pbox{5cm}{\textbf{0.996} \\ \small{\textit{(8)}}} \\
        \rowcolor{gray!10}
        \pbox{5cm}{learning rate : $10^{-5}$ \\ update rate : $7$} & \pbox{5cm}{\textbf{0.799} \\ \small{\textit{(994)}}} & \pbox{5cm}{\textbf{0.924} \\ \small{\textit{(994)}}} & \pbox{5cm}{\textbf{0.861} \\ \small{\textit{(994)}}} & \pbox{5cm}{0.967 \\ \small{\textit{(54)}}} \\
        \hline
    \end{tabular}}
    \caption{Best mean scores for different parameters of the GAN customized for Action Units \textit{(first line is the best score, second line in italics with parenthesis is the iteration in thousands at which the best score has been reached, figures in bold are the best scores for all models)}}
    \label{tab:gan_au_mean_results}
\end{table}

\begin{table}[!ht]
    \centering
    \begin{adjustwidth}{-0.03\textwidth}{-0.0\textwidth}
    {\begin{tabular}{|c|c c c c c c c c|}
        \hline
        \multirow{2}{*}{\textbf{Model}} & \multicolumn{8}{c|}{\textbf{F1 score}}\\

         & AU 1 & AU 2 & AU 4 & AU 6 & AU 12 & AU 15 & AU 20 & AU 25 \\
         \hline
         
        \rowcolor{gray!40}
        \pbox{5cm}{learning rate : $10^{-4}$ \\ update rate : $2$} & \pbox{5cm}{0.768 \\ \small{\textit{(775)}} } & \pbox{5cm}{0.648 \\ \small{\textit{(963)}} } & \pbox{5cm}{0.776 \\ \small{\textit{(963)}} } & \pbox{5cm}{0.772 \\ \small{\textit{(781)}} } & \pbox{5cm}{0.789 \\ \small{\textit{(964)}} } & \pbox{5cm}{0.670 \\ \small{\textit{(910)}} } & \pbox{5cm}{0.625 \\ \small{\textit{(929)}} } & \pbox{5cm}{0.678 \\ \small{\textit{(771)}} } \\
        \rowcolor{gray!10}
        \pbox{5cm}{learning rate : $10^{-4}$ \\ update rate : $5$} & \pbox{5cm}{0.782 \\ \small{\textit{(993)}}} & 
        \pbox{5cm}{0.650 \\ \small{\textit{(995)}}} & 
        \pbox{5cm}{0.791 \\ \small{\textit{(998)}}} & 
        \pbox{5cm}{0.783 \\ \small{\textit{(996)}}} & 
        \pbox{5cm}{0.804 \\ \small{\textit{(990)}}} & 
        \pbox{5cm}{0.684 \\ \small{\textit{(994)}}} & 
        \pbox{5cm}{0.627 \\ \small{\textit{(965)}}} & 
        \pbox{5cm}{0.682 \\ \small{\textit{(996)}}} \\
        \rowcolor{gray!40}
        \pbox{5cm}{learning rate : $10^{-4}$ \\ update rate : $7$} & \pbox{5cm}{0.828 \\ \small{\textit{(962)}}} & 
        \pbox{5cm}{0.720 \\ \small{\textit{(955)}}} & 
        \pbox{5cm}{0.841 \\ \small{\textit{(919)}}} & 
        \pbox{5cm}{0.827 \\ \small{\textit{(966)}}} & 
        \pbox{5cm}{0.841 \\ \small{\textit{(908)}}} & 
        \pbox{5cm}{0.766 \\ \small{\textit{(996)}}} & 
        \pbox{5cm}{0.731 \\ \small{\textit{(894)}}} & 
        \pbox{5cm}{0.754 \\ \small{\textit{(971)}}} \\
        
        \hline
        
        \rowcolor{gray!10}
        \pbox{5cm}{learning rate : $10^{-5}$ \\ update rate : $2$} & \pbox{5cm}{0.765 \\ \small{\textit{(922)}}} & 
        \pbox{5cm}{0.646 \\ \small{\textit{(981)}}} & 
        \pbox{5cm}{0.771 \\ \small{\textit{(912)}}} & 
        \pbox{5cm}{0.771 \\ \small{\textit{(989)}}} & 
        \pbox{5cm}{0.785 \\ \small{\textit{(926)}}} & 
        \pbox{5cm}{0.681 \\ \small{\textit{(837)}}} & 
        \pbox{5cm}{0.613 \\ \small{\textit{(597)}}} & 
        \pbox{5cm}{0.682 \\ \small{\textit{(871)}}} \\
        \rowcolor{gray!40}
        \pbox{5cm}{learning rate : $10^{-5}$ \\ update rate : $5$} & \pbox{5cm}{0.812 \\ \small{\textit{(990)}}} & 
        \pbox{5cm}{0.710 \\ \small{\textit{(989)}}} & 
        \pbox{5cm}{0.832 \\ \small{\textit{(971)}}} & 
        \pbox{5cm}{0.820 \\ \small{\textit{(997)}}} & 
        \pbox{5cm}{0.830 \\ \small{\textit{(994)}}} & 
        \pbox{5cm}{0.742 \\ \small{\textit{(993)}}} & 
        \pbox{5cm}{0.717 \\ \small{\textit{(998)}}} & 
        \pbox{5cm}{0.741 \\ \small{\textit{(984)}}} \\
        \rowcolor{gray!10}
        \pbox{5cm}{learning rate : $10^{-5}$ \\ update rate : $7$} & \pbox{5cm}{\textbf{0.835} \\ \small{\textit{(994)}}} & 
        \pbox{5cm}{\textbf{0.741} \\ \small{\textit{(994)}}} & 
        \pbox{5cm}{\textbf{0.853} \\ \small{\textit{(982)}}} & 
        \pbox{5cm}{\textbf{0.837} \\ \small{\textit{(913)}}} & 
        \pbox{5cm}{\textbf{0.852} \\ \small{\textit{(976)}}} & 
        \pbox{5cm}{\textbf{0.787} \\ \small{\textit{(998)}}} & 
        \pbox{5cm}{\textbf{0.755} \\ \small{\textit{(994)}}} & 
        \pbox{5cm}{\textbf{0.772} \\ \small{\textit{(995)}}} \\
        \hline
    \end{tabular}}
    \end{adjustwidth}
    \caption{Best F1 scores for the 8 Action Units for different parameters of the GAN customized for Action Units \textit{(first line is the best score, second line in italics with parenthesis is the iteration in thousands at which the best score has been reached, figures in bold are the best scores for all models)} }
    \label{tab:gan_au_results}
\end{table}

\begin{figure}[!ht]
\centering
\begin{tabular}{cccc}
  \includegraphics[scale=2.5]{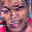} &
  \includegraphics[scale=2.5]{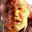} &   \includegraphics[scale=2.5]{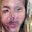} &
 \includegraphics[scale=2.5]{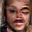}   \\
\end{tabular}
\caption{Images created by the Generator (sigmoid AU + RF / learning rate : $10^{-4}$ / update rate : 2 / $\alpha=0.9$) at iteration 261,000, 397,000, 672,000 and 693,000 (from left to right)}
\label{fig:gen_au_img_1}
\end{figure}

\begin{figure}[!ht]
\centering
\begin{tabular}{cccc}
  \includegraphics[scale=2.5]{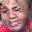} &
  \includegraphics[scale=2.5]{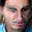} &   \includegraphics[scale=2.5]{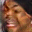} &
 \includegraphics[scale=2.5]{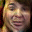}   \\
\end{tabular}
\caption{Images created by the Generator (sigmoid AU + RF / learning rate : $10^{-4}$ / update rate : 5 / $\alpha=0.9$) at iteration 147,000, 261,000, 397,000 and 688,000 (from left to right)}
\label{fig:gen_au_img_2}
\end{figure}

\begin{figure}[!ht]
\centering
\begin{tabular}{cccc}
  \includegraphics[scale=2.5]{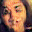} &
  \includegraphics[scale=2.5]{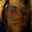} &
  \includegraphics[scale=2.5]{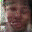} &   \includegraphics[scale=2.5]{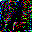}
   \\
\end{tabular}
\caption{Images created by the Generator (sigmoid AU + RF / learning rate : $10^{-4}$ / update rate : 7 / $\alpha=0.9$) at iteration 73,000, 147,000, 385,000 and 646,000 (from left to right)}
\label{fig:gen_au_img_1}
\end{figure}

\begin{figure}
\centering
\begin{tabular}{cccc}
  \includegraphics[scale=2.5]{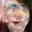} &
  \includegraphics[scale=2.5]{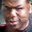} &   \includegraphics[scale=2.5]{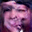} &
 \includegraphics[scale=2.5]{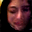}   \\
\end{tabular}
\caption{Images created by the Generator (sigmoid AU + RF / learning rate : $10^{-5}$ / update rate : 2 / $\alpha=0.9$) at iteration 148,000, 385,000, 688,000 and 999,000 (from left to right)}
\label{fig:gen_au_img_4}
\end{figure}

\begin{figure}
\centering
\begin{tabular}{cccc}
  \includegraphics[scale=2.5]{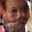} &
  \includegraphics[scale=2.5]{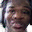} &   \includegraphics[scale=2.5]{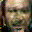} &
 \includegraphics[scale=2.5]{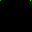}   \\
\end{tabular}
\caption{Images created by the Generator (sigmoid AU + RF / learning rate : $10^{-5}$ / update rate : 5 / $\alpha=0.9$) at iteration 147,000, 385,000, 672,000 and 999,000 (from left to right)}
\label{fig:gen_au_img_5}
\end{figure}

\begin{figure}
\centering
\begin{tabular}{cccc}
  \includegraphics[scale=2.5]{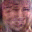} &
  \includegraphics[scale=2.5]{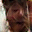} &   \includegraphics[scale=2.5]{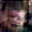} &
 \includegraphics[scale=2.5]{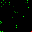}   \\
\end{tabular}
\caption{Images created by the Generator (sigmoid AU + RF / learning rate : $10^{-5}$ / update rate : 7 / $\alpha=0.9$) at iteration 73,000, 198,000, 567,000 and 817,000 (from left to right)}
\label{fig:gen_au_img_6}
\end{figure}

\clearpage

\subsubsection{Analysis}

From Tables \ref{tab:gan_au_mean_results} and \ref{tab:gan_au_results}, we can see that the best model to determine the presence or the absence of an Action Unit is for the learning rate $10^{-5}$ and the update rate 7. This model performs best for every score except the percentage of real images classified as real. This is quite surprising because it means that the less we backpropagate the error through the Discriminator, i.e. the less training for the Discriminator, the better it is. \\
For a given learning rate, we can see this increase in the scores when the update rate increases. \\
For a given update rate of 2 and 7, we see that changing the learning rate does not affect the scores much, while for an update rate of 5 the increase is more significant when the learning rate decreases from $10^{-4}$ to $10^{-5}$. \\
While the averaged scores and the F1 scores for the Action Units are reached at around 900,000 iterations (second line of the Tables), the best scores for the percentage of real images classified as real are reached at an early stage (at iteration 6,000 for the learning rate $10^{-4}$ and update rate 7 with a score of 0.996 for example). Not surprisingly, the less frequent Action Units in the dataset like the Action Unit 15 and 20 are the Action Units with the worse F1 score, while the Action Units more frequent like 1, 6, 12 have the best score for a given set of hyperparameters.
\\
\\
The best images produced by the Generator are for a learning rate of $10^{-4}$ and an update rate of 5. These images are 28*28. We can see from Figures \ref{fig:gen_au_img_4}, \ref{fig:gen_au_img_5} and \ref{fig:gen_au_img_6} that for a given learning rate of $10^{-5}$, increasing the update rate and so the frequency of the backpropagation through the Generator does not help it to generate better images, quite the contrary. After too many iterations, the Generator can produce black images because the Discriminator has won the battle against the Generator. This is the phenomenon of collapse mode already seen for the vanilla GAN (\ref{sec:vanilla_gan}).

\clearpage

\subsection{GANs for Valence Arousal}

\subsubsection{Model customization}

In this part, the architecture of the model was adapted to suit the need of predicting the Valence and Arousal values annotated on images picturing people reacting. \\ 
\\
As before, the Generator architecture has not been changed. \\
On the contrary, the last layer of the Discriminator has been modified : the fully-connected layer has 3 nodes, the first twos for predicting the value of Valence, the second for Arousal and the last one for forecasting if an image is real or fake. The sigmoid function is applied to this last node so as to turn the output of the fake or real node into a probability. \\
\\
While the Generator loss function has stayed the same, two different Discriminator loss functions has been tested for this architecture :
\begin{itemize}
    \item The Mean Squared Error function (MSE) (defined in \ref{sec:cost_fcts}) is computed between the predicted Valence and Arousal and the groundtruth labels for both the fake and real images (\textit{d\_loss\_real\_v, d\_loss\_real\_a, d\_loss\_fake\_v, d\_loss\_fake\_a} in Table \ref{code:va_loss_1} and Table \ref{code:va_loss_2}). 
    \item For Valence and Arousal, 1-CCC (described in \ref{sec:cost_fcts}) is calculated between the predictions for Valence and Arousal and the groundtruth labels (\textit{d\_loss\_real\_v, d\_loss\_real\_a, d\_loss\_fake\_v, d\_loss\_fake\_a} commented in Table \ref{code:va_loss_1} and Table \ref{code:va_loss_2}). \\
    \\
    For the first two configurations, the sigmoid cross entropy is applied between the fake or real output node and the groundtruth label indicating if the image is fake or real (\textit{d\_loss\_real\_rf} in Table \ref{code:va_loss_1} and \textit{d\_loss\_fake\_rf} in Table \ref{code:va_loss_2}) . \\
    Then, the mean of the loss function for Valence and Arousal is applied and taken to be averaged with the loss function computed for the real or fake image. This gives a loss function for both the real and fake images (\textit{d\_loss\_real\_rf} in Table \ref{code:va_loss_1} and \textit{d\_loss\_fake\_rf} in Table \ref{code:va_loss_2}). The mean of these two loss functions makes the total loss function of the
    Generator (\textit{d\_loss} in Table \ref{code:va_loss_2}).

\end{itemize}

\begin{table}[!ht]
        \includegraphics[width=1.\linewidth]{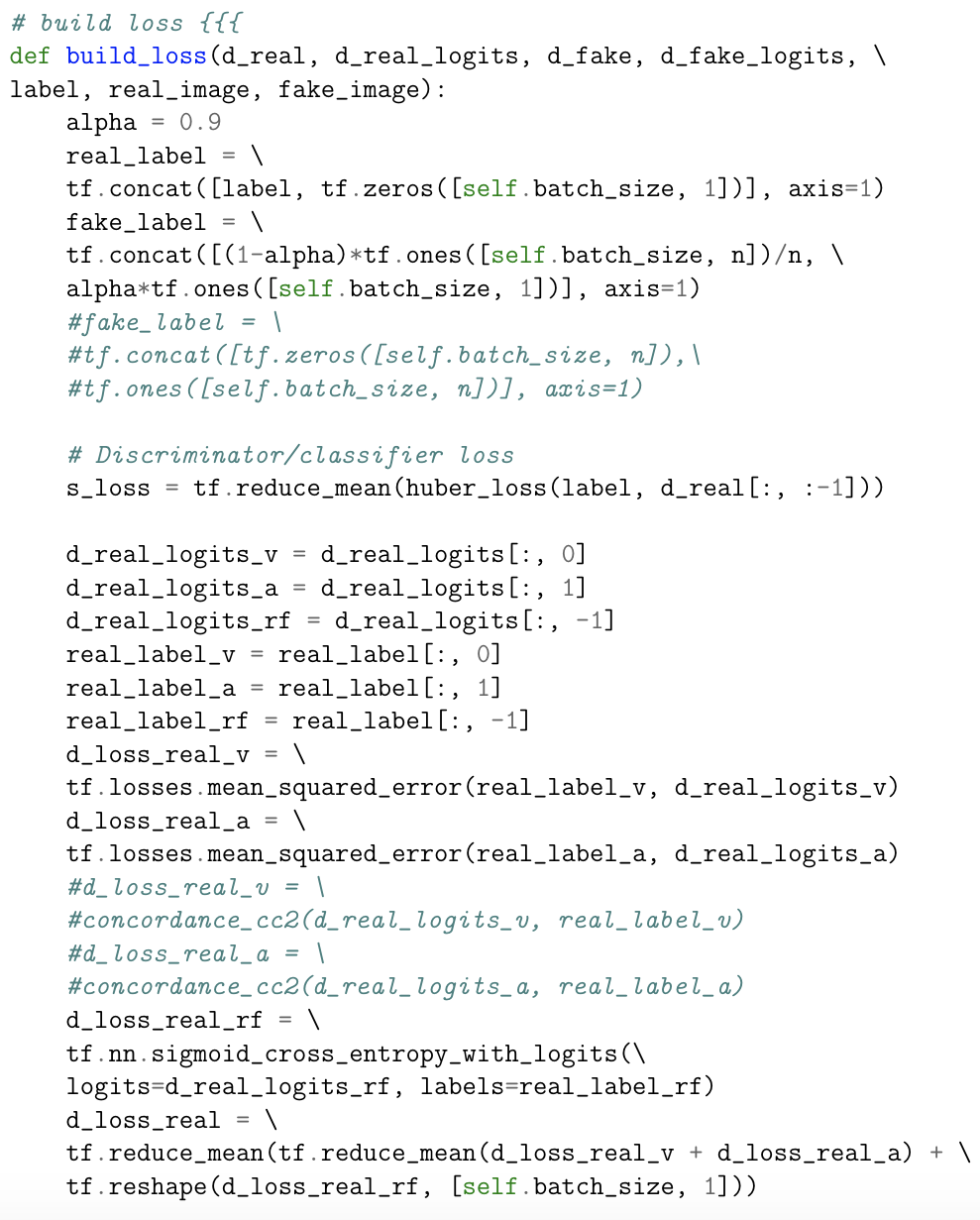}
\caption{Discriminator loss function for the VA GAN - Part 1}
\label{code:va_loss_1}
\end{table}

\begin{table}[!ht]
        \includegraphics[width=1.\linewidth]{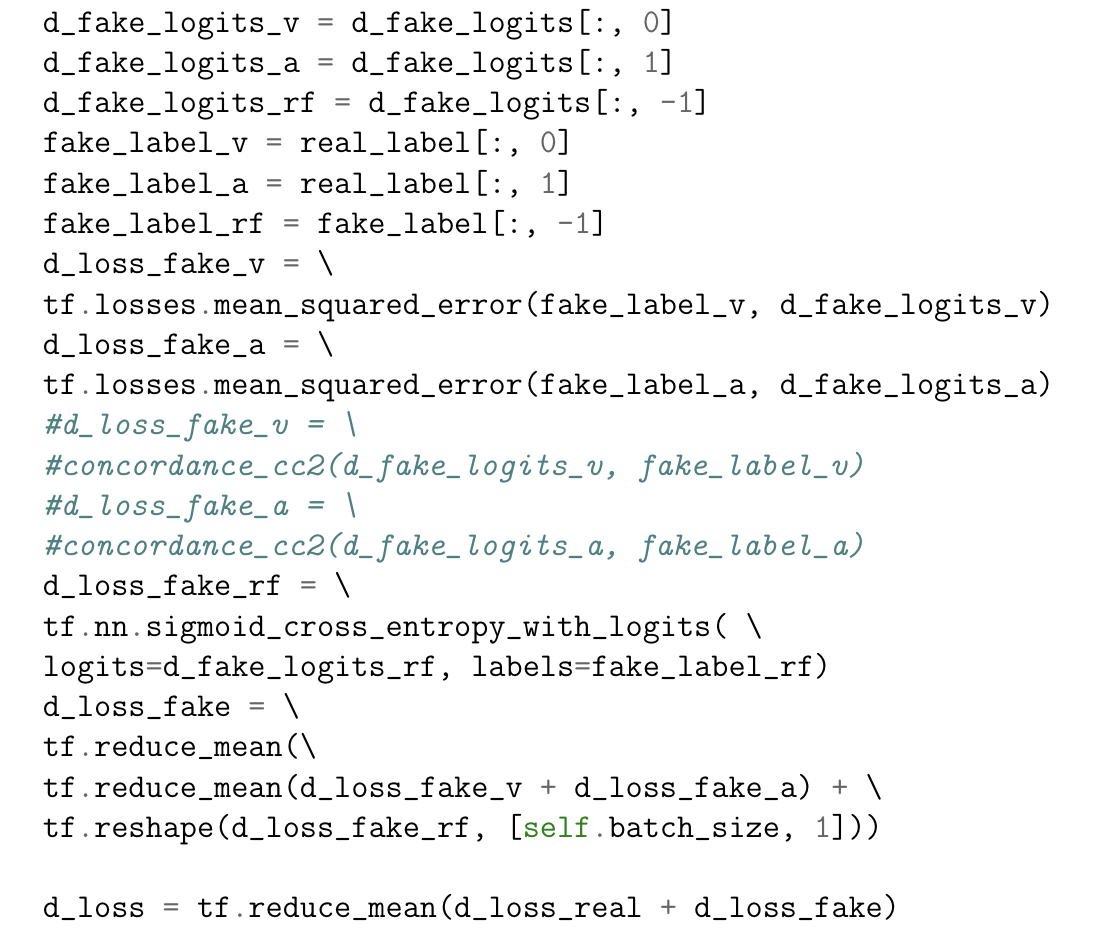}
\caption{Discriminator loss function for the VA GAN - Part 2}
\label{code:va_loss_2}
\end{table}

As mentioned previously, the groundtruth labels that are used for fake images can only be guessed as there are no real labels for fake images. This is the reason why this algorithm is semi-supervised. By default, the fake labels are also a n+1-vector with the n first values being $\frac{1-\alpha}{n}$ and the last value being $\alpha$ (\textit{fake\_label} in Table \ref{code:va_loss_1}). By default, $\alpha$ is 0.9. When the MSE was applied to the Valence and Arousal outputs, two values of $\alpha$ have been tested : $\alpha = 0.9$ (the default value) and $\alpha=1$. This change was only made for the MSE loss function (because the 1-CCC loss function would always have been 1 if all the labels were the same because because the numerator of the CCC will be 0 as y - being the label - will be equal to the mean of y). This operation of changing the value of the fake labels is called label smoothing and can help the neural networks to learn easily.

\subsubsection{Training \& Evaluation process}

The number of iterations is still 1,000,000. 
For the MSE loss, the different hyperparameters tried are : 
\begin{itemize}
    \item The learning rate : $10^{-4}$;
    \item The update rate : $2$, $5$;
    \item Alpha : $0.9$, $1$.
\end{itemize}
For the 1-CCC loss, the different hyperparameters tried are : 
\begin{itemize}
    \item The learning rate : $10^{-4}$, $10^{-5}$;
    \item The update rate : $2$, $5$.
\end{itemize}

For the evaluation of the different models once the training was finished, the following metrics have been chosen :
\begin{itemize}
    \item The Concordance Correlation Coefficient for Valence and Arousal between predictions and labels (defined in \ref{sec:cost_fcts});
    \item The MSE for Valence and Arousal for Valence and Arousal between predictions and labels (defined in \ref{sec:cost_fcts});
    \item The percentage of real images classified as real.
\end{itemize}

These metrics have been chosen because they give a really good insight of the performance of the neural network. The more the Concordance Correlation Coefficient is near 1, the more the variables (predictions and labels) are correlated. The more the CCC is near 0, the less variables are correlated. The more the CCC is near -1, the more the variables are correlated in opposite ways. The smaller the MSE (i.e. the closer to 0), the better. 

\subsubsection{Results}

The results for the MSE loss function are (first line is the best score, second line in italics with parenthesis is the iteration in thousands at which the best score has been reached, figures in bold are the best scores for all models) : \\
\begin{table}[!ht]
    \centering
    {\begin{tabular}{|c|c c|c c|c|}
        \hline
        \multirow{2}{*}{\textbf{Model}} & \multicolumn{2}{c}{\textbf{CCC}} & \multicolumn{2}{|c|}{\textbf{MSE}} & \multirow{2}{*}{\pbox{5cm}{\textbf{\% of real images} \\ \textbf{classified as real}}} \\ 
         & Valence & Arousal & Valence & Arousal & \\
         \hline
         
        \rowcolor{gray!40}
        \pbox{5cm}{MSE VA + sigmoid RF \\ learning rate : $10^{-4}$ \\ update rate : 2 \\ $alpha=0.9$} & 
        \pbox{5cm}{0.725 \\ \small{\textit{(879)}}} &  
        \pbox{5cm}{0.537 \\ \small{\textit{(878)}}} & 
        \pbox{5cm}{0.081 \\ \small{\textit{(879)}}} & 
        \pbox{5cm}{0.045 \\ \small{\textit{(878)}}} & 
        \pbox{5cm}{0.956 \\ \small{\textit{(694)}}} \\
        \rowcolor{gray!10}
        \pbox{5cm}{MSE VA + sigmoid RF \\ learning rate : $10^{-4}$ \\ update rate : 2 \\ $alpha=1$} & 
        \pbox{5cm}{\textbf{0.884} \\ \small{\textit{(845)}}} & 
        \pbox{5cm}{\textbf{0.784} \\ \small{\textit{(974)}}} & 
        \pbox{5cm}{\textbf{0.043} \\ \small{\textit{(942)}}} & 
        \pbox{5cm}{\textbf{0.027} \\ \small{\textit{(919)}}} & 
        \pbox{5cm}{0.896 \\ \small{\textit{(1)}}} \\
        
        \hline
        \rowcolor{gray!40}
        \pbox{5cm}{MSE VA + sigmoid RF \\ learning rate : $10^{-4}$ \\ update rate : 5 \\ $alpha=0.9$} & 
        \pbox{5cm}{0.761 \\ \small{\textit{(986)}}} & 
        \pbox{5cm}{0.567 \\ \small{\textit{(966)}}} & 
        \pbox{5cm}{0.075 \\ \small{\textit{(986)}}} & 
        \pbox{5cm}{0.044 \\ \small{\textit{(950)}}} & 
        \pbox{5cm}{\textbf{0.978} \\ \small{\textit{(8)}}} \\
        \rowcolor{gray!10}
        \pbox{5cm}{MSE VA + sigmoid RF \\ learning rate : $10^{-4}$ \\ update rate : 5 \\ $alpha=1$} & 
        \pbox{5cm}{0.850 \\ \small{\textit{(982}}} & 
        \pbox{5cm}{0.713 \\ \small{\textit{(958)}}} & 
        \pbox{5cm}{0.054 \\ \small{\textit{(987)}}} & 
        \pbox{5cm}{0.034 \\ \small{\textit{(994)}}} & 
        \pbox{5cm}{0.993 \\ \small{\textit{(10)}}} \\
        
        \hline
        
    \end{tabular}}

    \caption{Best scores of the metrics chosen to evaluate the GAN customized for Valence Arousal with the MSE loss function \textit{(first line is the best score, second line in italics with parenthesis is the iteration in thousands at which the best score has been reached, figures in bold are the best scores for all models)}}
    \label{tab:gan_va_results_mse}
\end{table}

\begin{figure}[!ht]
\centering
\begin{tabular}{cccc}
  \includegraphics[scale=2.5]{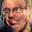} & 
  \includegraphics[scale=2.5]{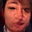} &
 \includegraphics[scale=2.5]{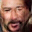} &   
 \includegraphics[scale=2.5]{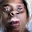} \\
\end{tabular}
\caption{Images created by the Generator (MSE VA + sigmoid RF / learning rate : $10^{-4}$ / update rate : 2 / $\alpha=0.9$) at iteration 15,000, 319,000, 319,000 and 529,000 (from left to right)}
\label{fig:gen_mse_img_1}
\end{figure}

\begin{figure}[!ht]
\centering
\begin{tabular}{cccc}
  \includegraphics[scale=2.5]{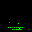} &   \includegraphics[scale=2.5]{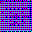} &
 \includegraphics[scale=2.5]{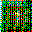} &   \includegraphics[scale=2.5]{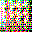} \\
\end{tabular}
\caption{Images created by the Generator (MSE VA + sigmoid RF / learning rate : $10^{-4}$ / update rate : 2 / $\alpha=1$) at iteration 147,000, 380,000, 646,000 and 999,000 (from left to right)}
\label{fig:gen_mse_img_2}
\end{figure}

\begin{figure}[!ht]
\centering
\begin{tabular}{cccc}
  \includegraphics[scale=2.5]{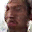} &   \includegraphics[scale=2.5]{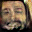} &
 \includegraphics[scale=2.5]{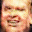} &   \includegraphics[scale=2.5]{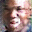} \\
\end{tabular}
\caption{Images created by the Generator (MSE VA + sigmoid RF / learning rate : $10^{-4}$ / update rate : 5 / $\alpha=0.9$) at iteration 147,000, 397,000, 646,000 and 688,000 (from left to right)}
\label{fig:gen_mse_img_3}
\end{figure}

\begin{figure}[!ht]
\centering
\begin{tabular}{cccc}
  \includegraphics[scale=2.5]{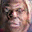} &   \includegraphics[scale=2.5]{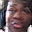} &
 \includegraphics[scale=2.5]{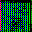} &   \includegraphics[scale=2.5]{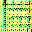} \\
\end{tabular}
\caption{Images created by the Generator (MSE VA + sigmoid RF / learning rate : $10^{-4}$ / update rate : 5 / $\alpha=1$) at iteration 147,000, 385,000, 688,000 and 919,000 (from left to right)}
\label{fig:gen_mse_img_4}
\end{figure}

\clearpage

The results for the 1-CCC loss function are (first line is the best score, second line in italics with parenthesis is the iteration in thousands at which the best score has been reached, figures in bold are the best scores for all models):
\begin{table}[!ht]
    \centering

    {\begin{tabular}{|c|c c|c c|c|}
        \hline
        \multirow{2}{*}{\textbf{Model}} & \multicolumn{2}{c}{\textbf{CCC}} & \multicolumn{2}{|c|}{\textbf{MSE}} & \multirow{2}{*}{\pbox{5cm}{\textbf{\% of real images} \\ \textbf{classified as real}}} \\ 
         & Valence & Arousal & Valence & Arousal & \\
        \hline
        \rowcolor{gray!40}
        \pbox{5cm}{CCC VA + sigmoid RF \\ learning rate : $10^{-4}$ \\ update rate : 2 \\ $alpha=0.9$} & 
        \pbox{5cm}{0.808 \\ \small{\textit{(950)}}} & 
        \pbox{5cm}{0.727 \\ \small{\textit{(977)}}} & 
        \pbox{5cm}{0.078 \\ \small{\textit{(950)}}} & 
        \pbox{5cm}{0.044 \\ \small{\textit{(959)}}} & 
        \pbox{5cm}{0.721 \\ \small{\textit{(243)}}} \\
        \rowcolor{gray!10}
        \pbox{5cm}{CCC VA + sigmoid RF \\ learning rate : $10^{-4}$ \\ update rate : 5 \\ $alpha=0.9$} & 
        \pbox{5cm}{\textbf{0.840} \\ \small{\textit{(996)}}} & 
        \pbox{5cm}{\textbf{0.770} \\ \small{\textit{(937)}}} & 
        \pbox{5cm}{\textbf{0.065} \\ \small{\textit{(883)}}} & 
        \pbox{5cm}{\textbf{0.035} \\ \small{\textit{(988)}}} & 
        \pbox{5cm}{0.789 \\ \small{\textit{(11)}}} \\
        
        \hline
        
        \rowcolor{gray!40}
        \pbox{5cm}{CCC VA + sigmoid RF \\ learning rate : $10^{-5}$ \\ update rate : 2 \\ $alpha=0.9$} & 
        \pbox{5cm}{0.808 \\ \small{\textit{(979)}}} & 
        \pbox{5cm}{0.728 \\ \small{\textit{(941)}}} & 
        \pbox{5cm}{0.081 \\ \small{\textit{(985)}}} & 
        \pbox{5cm}{0.044 \\ \small{\textit{(996)}}} & 
        \pbox{5cm}{0.749 \\ \small{\textit{(399)}}} \\
        \rowcolor{gray!10}
        \pbox{5cm}{CCC VA + sigmoid RF \\ learning rate : $10^{-5}$ \\ update rate : 5 \\ $alpha=0.9$} & 
        \pbox{5cm}{0.839 \\ \small{\textit{(866)}}} & 
        \pbox{5cm}{0.768 \\ \small{\textit{(968)}}} & 
        \pbox{5cm}{0.066 \\ \small{\textit{(917)}}} & 
        \pbox{5cm}{0.036 \\ \small{\textit{(973)}}} & 
        \pbox{5cm}{\textbf{0.843} \\ \small{\textit{(13)}}} \\
        \hline
        
    \end{tabular}}
    \caption{Best scores of the metrics chosen to evaluate the GAN customized for Valence Arousal with the 1-CCC loss function \textit{(first line is the best score, second line in italics with parenthesis is the iteration in thousands at which the best score has been reached, figures in bold are the best scores for all models)}}
    \label{tab:gan_va_results_ccc}
\end{table}

\begin{figure}[!ht]
\centering
\begin{tabular}{cccc}
  \includegraphics[scale=2.5]{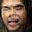} & 
  \includegraphics[scale=2.5]{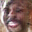} &
 \includegraphics[scale=2.5]{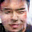} & 
 \includegraphics[scale=2.5]{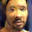} \\
\end{tabular}
\caption{Images created by the Generator (CCC VA + sigmoid RF / learning rate : $10^{-4}$ / update rate : 2 / $\alpha=0.9$) at iteration 147,000, 261,000, 646,000 and 999,000 (from left to right)}
\label{fig:gen_ccc_img_1}
\end{figure}

\begin{figure}[!ht]
\centering
\begin{tabular}{cccc}
  \includegraphics[scale=2.5]{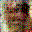} & 
  \includegraphics[scale=2.5]{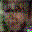} &
 \includegraphics[scale=2.5]{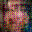} &   
 \includegraphics[scale=2.5]{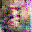} \\
\end{tabular}
\caption{Images created by the Generator (CCC VA + sigmoid RF / learning rate : $10^{-4}$ / update rate : 5 / $\alpha=0.9$) at iteration 147,000, 385,000, 646,000 and 688,000 (from left to right)}
\label{fig:gen_ccc_img_2}
\end{figure}

\begin{figure}[!ht]
\centering
\begin{tabular}{cccc}
  \includegraphics[scale=2.5]{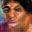} & 
  \includegraphics[scale=2.5]{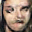} &
 \includegraphics[scale=2.5]{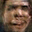} &   
 \includegraphics[scale=2.5]{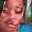} \\
\end{tabular}
\caption{Images created by the Generator (CCC VA + sigmoid RF / learning rate : $10^{-5}$ / update rate : 2 / $\alpha=0.9$) at iteration 261,000, 646,000, 688,000 and 999,000 (from left to right)}
\label{fig:gen_ccc_img_3}
\end{figure}

\begin{figure}[!ht]
\centering
\begin{tabular}{cccc}
  \includegraphics[scale=2.5]{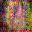} & 
  \includegraphics[scale=2.5]{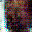} &
 \includegraphics[scale=2.5]{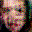} &   
 \includegraphics[scale=2.5]{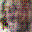} \\
\end{tabular}
\caption{Images created by the Generator (CCC VA + sigmoid RF / learning rate : $10^{-5}$ / update rate : 5 / $\alpha=0.9$) at iteration 147,000, 385,000, 646,000 and 999,000 (from left to right)}
\label{fig:gen_ccc_img_4}
\end{figure}

2
\clearpage

\subsubsection{Analysis}

For the MSE loss function, the best scores for CCC and MSE are reached with the learning rate $10^{-4}$, an update rate of 2 and $alpha=1$. We can notice that for a given learning rate and a given an update rate the model performs better with no label smoothing, i.e. when $alpha=1$. Furthermore, changing the update rate from 2 to 5 seem to have little effect on the scores. Finally the best percentage for real images classified as real is performed by the model with a learning rate of $10^{-4}$, an update rate of 5 and $alpha=0.9$.\\
The best looking images are generated for the model with a learning rate of $10^{-4}$, an update rate of 2 and $alpha=0.9$. From Figures \ref{fig:gen_mse_img_1} and \ref{fig:gen_mse_img_2} and the Figures \ref{fig:gen_mse_img_3} and \ref{fig:gen_mse_img_4}, we can see that increasing the update rate decreases the ability of the Generator to produce good images. Indeed, on the Figure \ref{fig:gen_mse_img_1} with an update rate of 2 the Generator produces good-looking images while on the Figure \ref{fig:gen_mse_img_2} with an update rate of 5 the Generator cannot produce any face-looking image and produce only noise.\\
\\
For the CCC loss function, the best scores are reached for a learning rate of $10^{-4}$ and an update rate of 5 (with $alpha=0.9$). The best percentage of images classified as real is made by the model with a learning rate of $10^{-5}$ and an update rate of 5. We can notice that changing the learning rate between $10^{-4}$ and $10^{-5}$ has very little effect on the performance. On the contrary increasing the update rate from 2 to 5 increases the performance. \\
Once again the best looking images are produced for the lowest update rates whatever the learning rates as shown when comparing the Figures \ref{fig:gen_ccc_img_1} and \ref{fig:gen_ccc_img_2} and the Figures \ref{fig:gen_ccc_img_3} and \ref{fig:gen_ccc_img_4}. The images with a higher update rate are less sharp and noisier. \\
\\
For both loss functions we can see that the Valence scores are better than the Arousal scores. This might be explained by the distributions of the two annotations across the dataset : the Valence score is quite balanced between positive and negative values (\ref{fig:valence_dist}) while the Arousal scores are mainly situated in the positive part (\ref{fig:arousal_dist}). 
Finally the MSE loss function perform better than the 1-CCC loss function for both CCC and MSE scores and the percentage of images classified as real. The images generated by the MSE loss function seem to be more prone to mode collapse while the CCC loss function performs a mix between noise and face in the worst cases. These two loss functions produce good looking images in the best scenarios.

\newpage

\subsection{GANs for Valence Arousal and Action Units}

\subsubsection{Model customization}

This part of the project consists of taking what has been done for the GAN customized for Action Units and the GAN customized for Valence and Arousal.\\ \\
Like the two previous GANs, the Generator architecture has not been changed. \\
The last layer of the Discriminator has been modified : it takes into account both change done in the previous GANs. As a result, the last layer is made up of 11 nodes : the two first nodes predict Valence and Arousal values, the following 8 nodes predict the presence or the absence of Action Units and the last one forecasts if the image fed into the Generator is real or fake. The sigmoid function is applied to the nodes dedicated to Action Units and the node prediciting if the image is real or fake.
\\
\\
The loss function for the Generator does not change as in the two previous GANs.
The 1-CCC loss function as well as the MSE loss function has been tested on the Valence Arousal nodes for this GAN. For the Action Units and the fake or real node the loss function is the sigmoid cross entropy. Then, the mean of the loss functions for Valence Arousal and the mean of the loss functions for the Action Units are computed in parallel. The average of these two previous means with the loss function for the real or fake node is computed. It gives the losses for Valence Arousal, Action Units and Real or Fake the same weights. This operation is made for both fake and real images. The mean of this fake and real loss gives the mean of the Generator.
\\
As the results obtained were good for Valence Arousal and bad for Action Units some loss functions have been weighted such that the loss for Valence Arousal weighted 0.27, the loss for Action Units 0.40 and the loss for Real or Fake 0.33. In according more importance to the loss of Action Units, better scores for this model were expected.

\subsubsection{Training \& Evaluation process}

Different hyperparameters have been tried for the 1-CCC loss function :
\begin{itemize}
    \item The learning rate : $10^{-4}$, $10^{-5}$;
    \item The update rate : $2$, $5$, $7$.
\end{itemize}
The hyperparameters tried for the MSE loss function are : 
\begin{itemize}
    \item The learning rate : $10^{-4}$, $10^{-5}$;
    \item The update rate : $2$, $5$;
    \item Alpha : $0.9$, $1$.
\end{itemize}

The different models tested after 1,000,000 iterations of training are evaluated using the metrics of the previous models : 
\begin{itemize}
    \item The best F1 score for each Action Unit;
    \item The best mean F1 score computed on all Action Units;
    \item The best mean accuracy computed on all Action Units;
    \item The Concordance Correlation Coefficient for Valence and Arousal between predictions and labels (defined in \ref{sec:cost_fcts});
    \item The MSE for Valence and Arousal for Valence and Arousal between predictions and labels (defined in \ref{sec:cost_fcts});
    \item The best score for the number of real images classified as real.  
\end{itemize}

\subsubsection{Results}

The results for the 1-CCC loss functions are (first line is the best score, second line in italics with parenthesis is the iteration in thousands at which the best score has been reached, figures in bold are the best scores for all models) :

\begin{table}[!ht]
    \centering
    {\begin{tabular}{|c|c c|c c|c|}
        \hline
        \multirow{2}{*}{\textbf{Model}} & \multicolumn{2}{c}{\textbf{CCC}} & \multicolumn{2}{|c|}{\textbf{MSE}} & \multirow{2}{*}{\pbox{5cm}{\textbf{\% of real images} \\ \textbf{classified as real}}} \\ 
         & Valence & Arousal & Valence & Arousal & \\
        \hline
        \rowcolor{gray!40}
        \pbox{5cm}{learning rate : $10^{-4}$ \\ update rate : 2} & \pbox{5cm}{0.812 \\ \small{\textit{(943)}}} & 
        \pbox{5cm}{0.730 \\ \small{\textit{(888)}}} & 
        \pbox{5cm}{0.080 \\ \small{\textit{(960)}}} & 
        \pbox{5cm}{0.040 \\ \small{\textit{(964)}}} & 
        \pbox{5cm}{0.983 \\ \small{\textit{(20)}}} \\
        \rowcolor{gray!10}
        \pbox{5cm}{learning rate : $10^{-4}$ \\ update rate : 5} & \pbox{5cm}{0.836 \\ \small{\textit{(914)}}} & 
        \pbox{5cm}{0.763 \\ \small{\textit{(996)}}} & 
        \pbox{5cm}{0.067 \\ \small{\textit{(924)}}} & 
        \pbox{5cm}{0.036 \\ \small{\textit{(958)}}} & 
        \pbox{5cm}{0.998 \\ \small{\textit{(10)}}} \\
        \rowcolor{gray!40}
        \pbox{5cm}{learning rate : $10^{-4}$ \\ update rate : 7} & \pbox{5cm}{0.833 \\ \small{\textit{(945)}}} & 
        \pbox{5cm}{0.762 \\ \small{\textit{(991)}}} & 
        \pbox{5cm}{0.068 \\ \small{\textit{(872)}}} & 
        \pbox{5cm}{0.037 \\ \small{\textit{(997)}}} & 
        \pbox{5cm}{\textbf{0.999} \\ \small{\textit{(26)}}} \\
        
        \hline
        \rowcolor{gray!10}
        \pbox{5cm}{learning rate : $10^{-5}$ \\ update rate : 2} & \pbox{5cm}{0.812 \\ \small{\textit{(990)}}} & 
        \pbox{5cm}{0.728 \\ \small{\textit{(967)}}} & 
        \pbox{5cm}{0.080 \\ \small{\textit{(986)}}} & 
        \pbox{5cm}{0.044 \\ \small{\textit{(698)}}} & 
        \pbox{5cm}{0.993 \\ \small{\textit{(2)}}}\\
        \rowcolor{gray!40}
        \pbox{5cm}{learning rate : $10^{-5}$ \\ update rate : 5} & \pbox{5cm}{\textbf{0.837} \\ \small{\textit{(970)}}} & 
        \pbox{5cm}{\textbf{0.764} \\ \small{\textit{(938)}}} & 
        \pbox{5cm}{0.067 \\ \small{\textit{(982)}}} & 
        \pbox{5cm}{\textbf{0.035} \\ \small{\textit{(920)}}} & 
        \pbox{5cm}{0.997 \\ \small{\textit{(16)}}} \\
        \rowcolor{gray!10}
        \pbox{5cm}{learning rate : $10^{-5}$ \\ update rate : 7} & \pbox{5cm}{0.832 \\ \small{\textit{(979)}}} & 
        \pbox{5cm}{0.759 \\ \small{\textit{(973)}}} & 
        \pbox{5cm}{0.068 \\ \small{\textit{(915)}}} & 
        \pbox{5cm}{0.037 \\ \small{\textit{(973)}}} & 
        \pbox{5cm}{0.998 \\ \small{\textit{(31)}}} \\
        
        \hline
        \hline
        
        \rowcolor{gray!40}
        \pbox{5cm}{learning rate : $10^{-5}$ \\ update rate : 2 \\ ponderated loss} & 
        \pbox{5cm}{0.804 \\ \small{\textit{(910)}}} & 
        \pbox{5cm}{0.718 \\ \small{\textit{(983)}}} & 
        \pbox{5cm}{0.083 \\ \small{\textit{(973)}}} & 
        \pbox{5cm}{0.046 \\ \small{\textit{(933)}}} & 
        \pbox{5cm}{0.981 \\ \small{\textit{(4)}}}\\
        \rowcolor{gray!10}
        \pbox{5cm}{learning rate : $10^{-5}$ \\ update rate : 5 \\ ponderated loss} & 
        \pbox{5cm}{\textbf{0.837} \\ \small{\textit{(994)}}} & 
        \pbox{5cm}{0.761 \\ \small{\textit{(947)}}} & 
        \pbox{5cm}{\textbf{0.066} \\ \small{\textit{(966)}}} & 
        \pbox{5cm}{0.037 \\ \small{\textit{(999)}}} & 
        \pbox{5cm}{0.996 \\ \small{\textit{(22)}}} \\
        
        \hline

    \end{tabular}}
    \caption{Best scores for the VA metrics for the GAN customized for Valence Arousal and Action Units with the 1-CCC loss function \textit{(first line is the best score, second line in italics with parenthesis is the iteration in thousands at which the best score has been reached, figures in bold are the best scores for all models)}}
    \label{tab:gan_all_results_1}
\end{table}

\begin{table}[!ht]

    \centering
    {\begin{tabular}{|c|c c c|c|}
        \hline
        \multirow{2}{*}{\textbf{Model}} & \multicolumn{3}{c|}{\textbf{Average on all Action Units}} \\
        & F1 score & Accuracy & Mean \\
        \hline
        \rowcolor{gray!40}
        \pbox{5cm}{learning rate : $10^{-4}$ \\ update rate : $2$} & \pbox{5cm}{0.218 \\ \small{\textit{(976)}}} & 
        \pbox{5cm}{0.550 \\ \small{\textit{(24)}}} & 
        \pbox{5cm}{0.323 \\ \small{\textit{(688)}}} \\
        \rowcolor{gray!10}
        \pbox{5cm}{learning rate : $10^{-4}$ \\ update rate : $5$} & \pbox{5cm}{0.130 \\ \small{\textit{(953)}}} & 
        \pbox{5cm}{0.559 \\ \small{\textit{(14)}}} & 
        \pbox{5cm}{0.316 \\ \small{\textit{(953)}}} \\
        \rowcolor{gray!40}
        \pbox{5cm}{learning rate : $10^{-4}$ \\ update rate : $7$} & \pbox{5cm}{0.125 \\ \small{\textit{(758)}}} & 
        \pbox{5cm}{0.561 \\ \small{\textit{(25)}}} & 
        \pbox{5cm}{0.314 \\ \small{\textit{(744)}}} \\
        
        \hline
        
        \rowcolor{gray!10}
        \pbox{5cm}{learning rate : $10^{-5}$ \\ update rate : $2$} & \pbox{5cm}{\textbf{0.230} \\ \small{\textit{(461)}}} & 
        \pbox{5cm}{0.559 \\ \small{\textit{(2)}}} & 
        \pbox{5cm}{\textbf{0.326} \\ \small{\textit{(293)}}} \\
        \rowcolor{gray!40}
        \pbox{5cm}{learning rate : $10^{-5}$ \\ update rate : $5$} & \pbox{5cm}{0.128 \\ \small{\textit{(977)}}} & 
        \pbox{5cm}{0.560 \\ \small{\textit{(15)}}} & 
        \pbox{5cm}{0.314 \\ \small{\textit{(675)}}} \\
        \rowcolor{gray!10}
        \pbox{5cm}{learning rate : $10^{-5}$ \\ update rate : $7$} & \pbox{5cm}{0.129 \\ \small{\textit{(960)}}} & 
        \pbox{5cm}{\textbf{0.561} \\ \small{\textit{(9)}}} & 
        \pbox{5cm}{0.314 \\ \small{\textit{(449)}}} \\

        \hline
        \hline
        
        \rowcolor{gray!40}
        \pbox{5cm}{learning rate : $10^{-5}$ \\ update rate : $2$ \\ ponderated loss} & 
        \pbox{5cm}{0.218 \\ \small{\textit{(446)}}} & 
        \pbox{5cm}{0.536 \\ \small{\textit{(5)}}} & 
        \pbox{5cm}{0.321 \\ \small{\textit{(519)}}} \\
        \rowcolor{gray!10}
        \pbox{5cm}{learning rate : $10^{-5}$ \\ update rate : $5$ \\ ponderated loss} & 
        \pbox{5cm}{0.128 \\ \small{\textit{(911)}}} & 
        \pbox{5cm}{0.558 \\ \small{\textit{(1)}}} & 
        \pbox{5cm}{0.314 \\ \small{\textit{(971)}}} \\
        
        \hline
        
    \end{tabular}}
    \caption{Best mean AU scores for different parameters of the GAN customized for Valence Arousal and Action Units with the 1-CCC loss function \textit{(first line is the best score, second line in italics with parenthesis is the iteration in thousands at which the best score has been reached, figures in bold are the best scores for all models)}}
    \label{tab:gan_all_mean_results}
\end{table}

\begin{table}[!ht]
    \centering
    \begin{adjustwidth}{-0.03\textwidth}{-0.0\textwidth}
    \begin{tabular}{|c|c c c c c c c c|}
        \hline
        \multirow{2}{*}{\textbf{Model}} & \multicolumn{8}{c|}{\textbf{F1 score}} \\
         & AU 1 & AU 2 & AU 4 & AU 6 & AU 12 & AU 15 & AU 20 & AU 25 \\
         \hline 
         
        \rowcolor{gray!40}
        \pbox{5cm}{learning rate : $10^{-4}$ \\ update rate : $2$} & \pbox{5cm}{0.435 \\ \small{\textit{(430)}}} & 
        \pbox{5cm}{0.508 \\ \small{\textit{(976)}}} & 
        \pbox{5cm}{0.225 \\ \small{\textit{(107)}}} & 
        \pbox{5cm}{0.243 \\ \small{\textit{(10)}}} & 
        \pbox{5cm}{0.210 \\ \small{\textit{(3)}}} & 
        \pbox{5cm}{0.068 \\ \small{\textit{(430)}}} & 
        \pbox{5cm}{0.022 \\ \small{\textit{(443)}}} & 
        \pbox{5cm}{0.102 \\ \small{\textit{(881)}}} \\
        \rowcolor{gray!10}
        \pbox{5cm}{learning rate : $10^{-4}$ \\ update rate : $5$} & \pbox{5cm}{0.328 \\ \small{\textit{(864)}}} & 
        \pbox{5cm}{0.368 \\ \small{\textit{(954)}}} & 
        \pbox{5cm}{0.080 \\ \small{\textit{(820)}}} & 
        \pbox{5cm}{0.162 \\ \small{\textit{(2)}}} & 
        \pbox{5cm}{0.163 \\ \small{\textit{(2)}}} & 
        \pbox{5cm}{0.062 \\ \small{\textit{(805)}}} & 
        \pbox{5cm}{0.019 \\ \small{\textit{(2)}}} & 
        \pbox{5cm}{0.091 \\ \small{\textit{(2)}}} \\
        \rowcolor{gray!40}
        \pbox{5cm}{learning rate : $10^{-4}$ \\ update rate : $7$} & \pbox{5cm}{0.311 \\ \small{\textit{(967)}}} & 
        \pbox{5cm}{0.341 \\ \small{\textit{(967)}}} & 
        \pbox{5cm}{0.084 \\ \small{\textit{(888)}}} & 
        \pbox{5cm}{\textbf{0.281} \\ \small{\textit{(1)}}} & 
        \pbox{5cm}{0.153 \\ \small{\textit{(1)}}} & 
        \pbox{5cm}{0.062 \\ \small{\textit{(966)}}} & 
        \pbox{5cm}{0.009 \\ \small{\textit{(891)}}} & 
        \pbox{5cm}{0.084 \\ \small{\textit{(5)}}} \\
        
        \hline
        
        \rowcolor{gray!10}
        \pbox{5cm}{learning rate : $10^{-5}$ \\ update rate : $2$} & \pbox{5cm}{\textbf{0.449} \\ \small{\textit{(461)}}} & 
        \pbox{5cm}{\textbf{0.537} \\ \small{\textit{(461)}}} & 
        \pbox{5cm}{0.224 \\ \small{\textit{(461)}}} & 
        \pbox{5cm}{0.242 \\ \small{\textit{(202)}}} & 
        \pbox{5cm}{\textbf{0.232} \\ \small{\textit{(202)}}} & 
        \pbox{5cm}{0.065 \\ \small{\textit{(306)}}} & 
        \pbox{5cm}{\textbf{0.028} \\ \small{\textit{(1)}}} & 
        \pbox{5cm}{\textbf{0.108} \\ \small{\textit{(430)}}} \\
        \rowcolor{gray!40}
        \pbox{5cm}{learning rate : $10^{-5}$ \\ update rate : $5$} & \pbox{5cm}{0.318 \\ \small{\textit{(789)}}} & 
        \pbox{5cm}{0.352 \\ \small{\textit{(953)}}} & 
        \pbox{5cm}{0.078 \\ \small{\textit{(937)}}} & 
        \pbox{5cm}{0.217 \\ \small{\textit{(1)}}} & 
        \pbox{5cm}{0.166 \\ \small{\textit{(1)}}} & 
        \pbox{5cm}{0.061 \\ \small{\textit{(997)}}} & 
        \pbox{5cm}{0.012 \\ \small{\textit{(621)}}} & 
        \pbox{5cm}{0.086 \\ \small{\textit{(567)}}} \\
        \rowcolor{gray!10}
        \pbox{5cm}{learning rate : $10^{-5}$ \\ update rate : $7$} & \pbox{5cm}{0.317 \\ \small{\textit{(962)}}} & 
        \pbox{5cm}{0.361 \\ \small{\textit{(960)}}} & 
        \pbox{5cm}{0.118 \\ \small{\textit{(4)}}} & 
        \pbox{5cm}{0.145 \\ \small{\textit{(536)}}} & 
        \pbox{5cm}{0.186 \\ \small{\textit{(3)}}} & 
        \pbox{5cm}{0.059 \\ \small{\textit{(833)}}} & 
        \pbox{5cm}{0.009 \\ \small{\textit{(922)}}} & 
        \pbox{5cm}{0.077 \\ \small{\textit{(747)}}} \\

        \hline
        \hline
        \rowcolor{gray!40}
        \pbox{5cm}{learning rate : $10^{-5}$ \\ update rate : $2$ \\ ponderated loss} & 
        \pbox{5cm}{0.431 \\ \small{\textit{(839)}}} & 
        \pbox{5cm}{0.484 \\ \small{\textit{(446)}}} & 
        \pbox{5cm}{0.220 \\ \small{\textit{(446)}}} & 
        \pbox{5cm}{0.243 \\ \small{\textit{(446)}}} & 
        \pbox{5cm}{0.195 \\ \small{\textit{(446)}}} & 
        \pbox{5cm}{\textbf{0.072} \\ \small{\textit{(653)}}} & 
        \pbox{5cm}{0.025 \\ \small{\textit{(5)}}} & 
        \pbox{5cm}{0.097 \\ \small{\textit{(888)}}} \\
        \rowcolor{gray!10}
        \pbox{5cm}{learning rate : $10^{-5}$ \\ update rate : $5$ \\ ponderated loss} & 
        \pbox{5cm}{0.317 \\ \small{\textit{(823)}}} & 
        \pbox{5cm}{0.362 \\ \small{\textit{(913)}}} & 
        \pbox{5cm}{\textbf{0.318} \\ \small{\textit{(1)}}} & 
        \pbox{5cm}{0.143 \\ \small{\textit{(806)}}} & 
        \pbox{5cm}{0.128 \\ \small{\textit{(3)}}} & 
        \pbox{5cm}{0.063 \\ \small{\textit{(986)}}} & 
        \pbox{5cm}{0.013 \\ \small{\textit{(547)}}} & 
        \pbox{5cm}{0.080 \\ \small{\textit{(647)}}} \\
        
        \hline

    \end{tabular}
    \end{adjustwidth}
    \caption{Best F1 scores for the Action Units for different parameters of the GAN customized for Valence Arousal and Action Units with the 1-CCC loss function \textit{(first line is the best score, second line in italics with parenthesis is the iteration in thousands at which the best score has been reached, figures in bold are the best scores for all models)}}
    \label{tab:gan_all_results}
\end{table}

\clearpage

\begin{figure}[!ht]
\centering
\begin{tabular}{cccc}
  \includegraphics[scale=2.5]{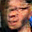} & 
  \includegraphics[scale=2.5]{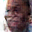} &
 \includegraphics[scale=2.5]{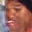} &   
 \includegraphics[scale=2.5]{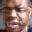} \\
\end{tabular}
\caption{Images created by the Generator (CCC VA + sigmoid AU + RF / learning rate : $10^{-4}$ / update rate : 2 / $\alpha=0.9$) at iteration 261,000, 385,000, 646,000 and 693,000 (from left to right)}
\label{fig:gen_all_ccc_img_1}
\end{figure}

\begin{figure}[!ht]
\centering
\begin{tabular}{cccc}
  \includegraphics[scale=2.5]{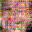} & 
  \includegraphics[scale=2.5]{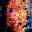} &
 \includegraphics[scale=2.5]{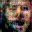} &   
 \includegraphics[scale=2.5]{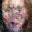} \\
\end{tabular}
\caption{Images created by the Generator (CCC VA + sigmoid AU + RF / learning rate : $10^{-4}$ / update rate : 5 / $\alpha=0.9$) at iteration 147,000, 385,000, 672,000 and 693,000 (from left to right)}
\label{fig:gen_all_ccc_img_2}
\end{figure}

\begin{figure}[!ht]
\centering
\begin{tabular}{cccc}
  \includegraphics[scale=2.5]{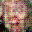} & 
  \includegraphics[scale=2.5]{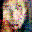} &
 \includegraphics[scale=2.5]{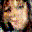} &   
 \includegraphics[scale=2.5]{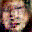} \\
\end{tabular}
\caption{Images created by the Generator (CCC VA + sigmoid AU + RF / learning rate : $10^{-4}$ / update rate : 7 / $\alpha=0.9$) at iteration 147,000, 261,000, 672,000 and 693,000 (from left to right)}
\label{fig:gen_all_ccc_img_3}
\end{figure}

\begin{figure}[!ht]
\centering
\begin{tabular}{cccc}
 \includegraphics[scale=2.5]{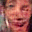} &
  \includegraphics[scale=2.5]{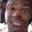} & 
  \includegraphics[scale=2.5]{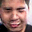} &
 \includegraphics[scale=2.5]{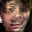}
 \\
\end{tabular}
\caption{Images created by the Generator (CCC VA + sigmoid AU + RF / learning rate : $10^{-5}$ / update rate : 2 / $\alpha=0.9$) at iteration 73,000, 261,000, 397,000 and 688,000 (from left to right)}
\label{fig:gen_all_ccc_img_4}
\end{figure}

\begin{figure}[!ht]
\centering
\begin{tabular}{cccc}
  \includegraphics[scale=2.5]{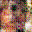} & 
  \includegraphics[scale=2.5]{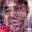} &
 \includegraphics[scale=2.5]{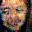} &   
 \includegraphics[scale=2.5]{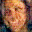} \\
\end{tabular}
\caption{Images created by the Generator (CCC VA + sigmoid AU + RF / learning rate : $10^{-5}$ / update rate : 5 / $\alpha=0.9$) at iteration 147,000, 347,000, 693,000 and 999,000 (from left to right)}
\label{fig:gen_all_ccc_img_5}
\end{figure}

\begin{figure}[!ht]
\centering
\begin{tabular}{cccc}
  \includegraphics[scale=2.5]{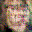} & 
  \includegraphics[scale=2.5]{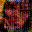} &
 \includegraphics[scale=2.5]{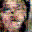} &   
 \includegraphics[scale=2.5]{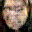} \\
\end{tabular}
\caption{Images created by the Generator (CCC VA + sigmoid AU + RF / learning rate : $10^{-5}$ / update rate : 7 / $\alpha=0.9$) at iteration 261,000, 646,000, 688,000 and 999,000 (from left to right)}
\label{fig:gen_all_ccc_img_6}
\end{figure}

\begin{figure}[!ht]
\centering
\begin{tabular}{cccc}
  \includegraphics[scale=2.5]{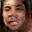} & 
  \includegraphics[scale=2.5]{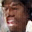} &
 \includegraphics[scale=2.5]{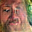} &   
 \includegraphics[scale=2.5]{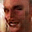} \\
\end{tabular}
\caption{Images created by the Generator (CCC VA + sigmoid AU + RF / learning rate : $10^{-5}$ / update rate : 2 / $\alpha=0.9$ / ponderated loss) at iteration 147,000, 646,000, 672,000 and 693,000 (from left to right)}
\label{fig:gen_all_ccc_img_7}
\end{figure}

\begin{figure}[!ht]
\centering
\begin{tabular}{cccc}
  \includegraphics[scale=2.5]{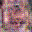} & 
  \includegraphics[scale=2.5]{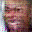} &
 \includegraphics[scale=2.5]{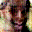} &   
 \includegraphics[scale=2.5]{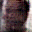} \\
\end{tabular}
\caption{Images created by the Generator (CCC VA + sigmoid AU + RF / learning rate : $10^{-5}$ / update rate : 5 / $\alpha=0.9$ / ponderated loss) at iteration 147,000, 261,000, 688,000 and 999,000 (from left to right)}
\label{fig:gen_all_ccc_img_8}
\end{figure}

\clearpage

The results for the MSE loss function are (first line is the best score, second line in italics with parenthesis is the iteration in thousands at which the best score has been reached, figures in bold are the best scores for all models) : \\
\begin{table}[!ht]
    \centering
    {\begin{tabular}{|c|c c|c c|c|}
        \hline
        \multirow{2}{*}{\textbf{Model}} & \multicolumn{2}{c}{\textbf{CCC}} & \multicolumn{2}{|c|}{\textbf{MSE}} & \multirow{2}{*}{\pbox{5cm}{\textbf{\% of real images} \\ \textbf{classified as real}}} \\ 
         & Valence & Arousal & Valence & Arousal & \\
         \hline
        \rowcolor{gray!40}
        \pbox{5cm}{MSE VA + sigmoid RF \\ learning rate : $10^{-4}$ \\ update rate : 2 \\ $alpha=0.9$} & 
        \pbox{5cm}{0.709 \\ \small{\textit{(997)}}} &  
        \pbox{5cm}{0.519 \\ \small{\textit{(837)}}} & 
        \pbox{5cm}{0.084 \\ \small{\textit{(997)}}} & 
        \pbox{5cm}{0.046 \\ \small{\textit{(816)}}} & 
        \pbox{5cm}{0.987 \\ \small{\textit{(3)}}} \\
        \rowcolor{gray!10}
        \pbox{5cm}{MSE VA + sigmoid RF \\ learning rate : $10^{-4}$ \\ update rate : 2 \\ $alpha=1$} & 
        \pbox{5cm}{\textbf{0.851} \\ \small{\textit{(945)}}} & 
        \pbox{5cm}{\textbf{0.733} \\ \small{\textit{(962)}}} & 
        \pbox{5cm}{\textbf{0.053} \\ \small{\textit{(995)}}} & 
        \pbox{5cm}{0.033 \\ \small{\textit{(959)}}} & 
        \pbox{5cm}{0.983 \\ \small{\textit{(6)}}} \\
        \hline
        \rowcolor{gray!40}
        \pbox{5cm}{MSE VA + sigmoid RF \\ learning rate : $10^{-4}$ \\ update rate : 5 \\ $alpha=0.9$} & 
        \pbox{5cm}{0.838 \\ \small{\textit{(934)}}} & 
        \pbox{5cm}{0.710 \\ \small{\textit{(929)}}} & 
        \pbox{5cm}{0.057 \\ \small{\textit{(998)}}} & 
        \pbox{5cm}{0.035 \\ \small{\textit{(928)}}} & 
        \pbox{5cm}{0.995 \\ \small{\textit{(6)}}} \\
        \rowcolor{gray!10}
        \pbox{5cm}{MSE VA + sigmoid RF \\ learning rate : $10^{-4}$ \\ update rate : 5 \\ $alpha=1$} & 
        \pbox{5cm}{0.715 \\ \small{\textit{(966)}}} & 
        \pbox{5cm}{0.529 \\ \small{\textit{(990)}}} & 
        \pbox{5cm}{0.084 \\ \small{\textit{(966)}}} & 
        \pbox{5cm}{0.056 \\ \small{\textit{(993)}}} & 
        \pbox{5cm}{0.995 \\ \small{\textit{(3)}}} \\
        
        \hline
        
        \rowcolor{gray!40}
        \pbox{5cm}{MSE VA + sigmoid RF \\ learning rate : $10^{-5}$ \\ update rate : 2 \\ $alpha=0.9$} & 
        \pbox{5cm}{0.709 \\ \small{\textit{(915)}}} & 
        \pbox{5cm}{0.519 \\ \small{\textit{(637)}}} & 
        \pbox{5cm}{0.084 \\ \small{\textit{(915)}}} & 
        \pbox{5cm}{0.046 \\ \small{\textit{(942)}}} & 
        \pbox{5cm}{0.991 \\ \small{\textit{(3)}}} \\
        \rowcolor{gray!10}
        \pbox{5cm}{MSE VA + sigmoid RF \\ learning rate : $10^{-5}$ \\ update rate : 2 \\ $alpha=1$} & 
        \pbox{5cm}{0.850 \\ \small{\textit{(756)}}} & 
        \pbox{5cm}{0.728 \\ \small{\textit{(820)}}} & 
        \pbox{5cm}{\textbf{0.053} \\ \small{\textit{(951)}}} & 
        \pbox{5cm}{\textbf{0.032} \\ \small{\textit{(928)}}} & 
        \pbox{5cm}{0.980 \\ \small{\textit{(2)}}} \\
        \hline 
        \rowcolor{gray!40}
        \pbox{5cm}{MSE VA + sigmoid RF \\ learning rate : $10^{-5}$ \\ update rate : 5 \\ $alpha=0.9$} & 
        \pbox{5cm}{0.710 \\ \small{\textit{(983)}}} & 
        \pbox{5cm}{0.523 \\ \small{\textit{(972)}}} & 
        \pbox{5cm}{0.084 \\ \small{\textit{(983)}}} & 
        \pbox{5cm}{0.046 \\ \small{\textit{(972)}}} & 
        \pbox{5cm}{\textbf{0.996} \\ \small{\textit{(2)}}} \\
        \rowcolor{gray!10}
        \pbox{5cm}{MSE VA + sigmoid RF \\ learning rate : $10^{-5}$ \\ update rate : 5 \\ $alpha=1$} & 
        \pbox{5cm}{0.833 \\ \small{\textit{(999)}}} & 
        \pbox{5cm}{0.697 \\ \small{\textit{(984)}}} & 
        \pbox{5cm}{0.058 \\ \small{\textit{(996)}}} & 
        \pbox{5cm}{0.035 \\ \small{\textit{(962)}}} & 
        \pbox{5cm}{0.994 \\ \small{\textit{(3)}}} \\
        
        \hline
        \hline
        
        \rowcolor{gray!40}
        \pbox{5cm}{MSE VA + sigmoid RF \\ learning rate : $10^{-4}$ \\ update rate : 2 \\ $alpha=1$ \\ ponderated loss} & \pbox{5cm}{0.836 \\ \small{\textit{(942)}}} & 
        \pbox{5cm}{0.706 \\ \small{\textit{(934)}}} & 
        \pbox{5cm}{0.058 \\ \small{\textit{(926)}}} & 
        \pbox{5cm}{0.035 \\ \small{\textit{(947)}}} & 
        \pbox{5cm}{0.984 \\ \small{\textit{(4)}}} \\

        \hline
        
    \end{tabular}}

    \caption{Best scores for the CCC and MSE for the GAN customized for Valence Arousal and Action Units with the MSE loss function \textit{(first line is the best score, second line in italics with parenthesis is the iteration in thousands at which the best score has been reached, figures in bold are the best scores for all models)}}
    \label{tab:gan_all_results_mse}
\end{table}

\begin{table}[!ht]

    \centering
    {\begin{tabular}{|c|c c c|c|}
        \hline
        \multirow{2}{*}{\textbf{Model}} & \multicolumn{3}{c|}{\textbf{Average on all Action Units}} \\
        & F1 score & Accuracy & Mean \\
        \hline
        \rowcolor{gray!40}
        \pbox{5cm}{MSE VA + sigmoid RF \\ learning rate : $10^{-4}$ \\ update rate : 2 \\ $alpha=0.9$} & 
        \pbox{5cm}{0.227 \\ \small{\textit{(728)}}} & 
        \pbox{5cm}{0.558 \\ \small{\textit{(1)}}} & 
        \pbox{5cm}{0.325 \\ \small{\textit{(287)}}} \\
        \rowcolor{gray!10}
        \pbox{5cm}{MSE VA + sigmoid RF \\ learning rate : $10^{-4}$ \\ update rate : 2 \\ $alpha=1$} & 
        \pbox{5cm}{0.183 \\ \small{\textit{(873)}}} & 
        \pbox{5cm}{0.545 \\ \small{\textit{(2)}}} & 
        \pbox{5cm}{0.323 \\ \small{\textit{(6)}}} \\
        \hline
        \rowcolor{gray!40}
        \pbox{5cm}{MSE VA + sigmoid RF \\ learning rate : $10^{-4}$ \\ update rate : 5 \\ $alpha=0.9$} & 
        \pbox{5cm}{0.243 \\ \small{\textit{(328)}}} & 
        \pbox{5cm}{0.560 \\ \small{\textit{(8)}}} & 
        \pbox{5cm}{0.327 \\ \small{\textit{(332)}}} \\
        \rowcolor{gray!10}
        \pbox{5cm}{MSE VA + sigmoid RF \\ learning rate : $10^{-4}$ \\ update rate : 5 \\ $alpha=1$} & 
        \pbox{5cm}{0.228 \\ \small{\textit{(679)}}} & 
        \pbox{5cm}{0.557 \\ \small{\textit{(11)}}} & 
        \pbox{5cm}{0.323 \\ \small{\textit{(827)}}} \\

        \hline
        \hline

        \rowcolor{gray!40}
        \pbox{5cm}{MSE VA + sigmoid RF \\ learning rate : $10^{-5}$ \\ update rate : 2 \\ $alpha=0.9$} & 
        \pbox{5cm}{\textbf{0.247} \\ \small{\textit{(879)}}} &
        \pbox{5cm}{0.558 \\ \small{\textit{(3)}}} & 
        \pbox{5cm}{0.324 \\ \small{\textit{(879)}}} \\
        \rowcolor{gray!10}
        \pbox{5cm}{MSE VA + sigmoid RF \\ learning rate : $10^{-5}$ \\ update rate : 2 \\ $alpha=1$} & 
        \pbox{5cm}{0.220 \\ \small{\textit{(84)}}} & 
        \pbox{5cm}{0.554 \\ \small{\textit{(1)}}} & 
        \pbox{5cm}{0.326 \\ \small{\textit{(43)}}} \\
        \hline
        \rowcolor{gray!40}
        \pbox{5cm}{MSE VA + sigmoid RF \\ learning rate : $10^{-5}$ \\ update rate : 5 \\ $alpha=0.9$} & 
        \pbox{5cm}{0.220 \\ \small{\textit{(389)}}} & 
        \pbox{5cm}{\textbf{0.562} \\ \small{\textit{(8)}}} & 
        \pbox{5cm}{\textbf{0.339} \\ \small{\textit{(1)}}} \\
        \rowcolor{gray!10}
        \pbox{5cm}{MSE VA + sigmoid RF \\ learning rate : $10^{-5}$ \\ update rate : 5 \\ $alpha=1$} & 
        \pbox{5cm}{0.237 \\ \small{\textit{(324)}}} & 
        \pbox{5cm}{0.559 \\ \small{\textit{(2)}}} & 
        \pbox{5cm}{0.325 \\ \small{\textit{(324)}}} \\
        
        \hline
        \hline
        
        \rowcolor{gray!40}
        \pbox{5cm}{MSE VA + sigmoid RF \\ learning rate : $10^{-4}$ \\ update rate : 2 \\ $alpha=1$ \\ ponderated loss} & \pbox{5cm}{0.181 \\ \small{\textit{(611)}}} & 
        \pbox{5cm}{0.541 \\ \small{\textit{(1)}}} & 
        \pbox{5cm}{0.320 \\ \small{\textit{(611)}}} \\
        
        \hline

    \end{tabular}}
    \caption{Best mean AU scores for different parameters of the GAN customized for Valence Arousal and Action Units with MSE loss function \textit{(first line is the best score, second line in italics with parenthesis is the iteration in thousands at which the best score has been reached, figures in bold are the best scores for all models)}}
    \label{tab:gan_all_mean_results_mse}
\end{table}

\begin{table}[!ht]
    \centering
    \begin{adjustwidth}{-0.05\textwidth}{-0.0\textwidth}
    \begin{tabular}{|c|c c c c c c c c|}
        \hline
        \multirow{2}{*}{\textbf{Model}} & \multicolumn{8}{c|}{\textbf{F1 score}} \\
         & AU 1 & AU 2 & AU 4 & AU 6 & AU 12 & AU 15 & AU 20 & AU 25 \\
         \hline 
         
        \rowcolor{gray!40}
        \pbox{5cm}{MSE VA + sigmoid RF \\ learning rate : $10^{-4}$ \\ update rate : 2 \\ $alpha=0.9$} & 
        \pbox{5cm}{0.443 \\ \small{\textit{(402)}}} & 
        \pbox{5cm}{\textbf{0.551} \\ \small{\textit{(729)}}} & 
        \pbox{5cm}{0.221 \\ \small{\textit{(96)}}} & 
        \pbox{5cm}{0.245 \\ \small{\textit{(401)}}} & 
        \pbox{5cm}{0.205 \\ \small{\textit{(412)}}} & 
        \pbox{5cm}{0.080 \\ \small{\textit{(728)}}} & 
        \pbox{5cm}{0.030 \\ \small{\textit{(401)}}} & 
        \pbox{5cm}{0.114 \\ \small{\textit{(72)}}} \\
        \rowcolor{gray!10}
        \pbox{5cm}{MSE VA + sigmoid RF \\ learning rate : $10^{-4}$ \\ update rate : 2 \\ $alpha=1$} & 
        \pbox{5cm}{0.412 \\ \small{\textit{(78)}}} & 
        \pbox{5cm}{0.510 \\ \small{\textit{(858)}}} & 
        \pbox{5cm}{0.229 \\ \small{\textit{(84)}}} & 
        \pbox{5cm}{0.242 \\ \small{\textit{(88)}}} & 
        \pbox{5cm}{0.201 \\ \small{\textit{(88)}}} & 
        \pbox{5cm}{0.084 \\ \small{\textit{(650)}}} & 
        \pbox{5cm}{0.048 \\ \small{\textit{(650)}}} & 
        \pbox{5cm}{\textbf{0.123} \\ \small{\textit{(84)}}} \\
        \hline
        \rowcolor{gray!40}
        \pbox{5cm}{MSE VA + sigmoid RF \\ learning rate : $10^{-4}$ \\ update rate : 5 \\ $alpha=0.9$} & 
        \pbox{5cm}{0.462 \\ \small{\textit{(399)}}} & 
        \pbox{5cm}{0.550 \\ \small{\textit{(332)}}} & 
        \pbox{5cm}{\textbf{0.250} \\ \small{\textit{(328)}}} & 
        \pbox{5cm}{0.255 \\ \small{\textit{(328)}}} & 
        \pbox{5cm}{0.243 \\ \small{\textit{(328)}}} & 
        \pbox{5cm}{\textbf{0.100} \\ \small{\textit{(328)}}} & 
        \pbox{5cm}{0.065 \\ \small{\textit{(378)}}} & 
        \pbox{5cm}{0.113 \\ \small{\textit{(332)}}} \\
        \rowcolor{gray!10}
        \pbox{5cm}{MSE VA + sigmoid RF \\ learning rate : $10^{-4}$ \\ update rate : 5 \\ $alpha=1$} & 
        \pbox{5cm}{0.458 \\ \small{\textit{(453)}}} & 
        \pbox{5cm}{0.529 \\ \small{\textit{(679)}}} & 
        \pbox{5cm}{0.242 \\ \small{\textit{(514)}}} & 
        \pbox{5cm}{0.240 \\ \small{\textit{(514)}}} & 
        \pbox{5cm}{0.226 \\ \small{\textit{(679)}}} & 
        \pbox{5cm}{0.075 \\ \small{\textit{(679)}}} & 
        \pbox{5cm}{0.035 \\ \small{\textit{(1)}}} & 
        \pbox{5cm}{0.105 \\ \small{\textit{(514)}}} \\
        
        \hline
        \hline
        
        \rowcolor{gray!40}
        \pbox{5cm}{MSE VA + sigmoid RF \\ learning rate : $10^{-5}$ \\ update rate : 2 \\ $alpha=0.9$} & 
        \pbox{5cm}{0.445 \\ \small{\textit{(878)}}} & 
        \pbox{5cm}{\textbf{0.551} \\ \small{\textit{(878)}}} & 
        \pbox{5cm}{0.275 \\ \small{\textit{(878)}}} & 
        \pbox{5cm}{0.263 \\ \small{\textit{(878)}}} & 
        \pbox{5cm}{0.233 \\ \small{\textit{(878)}}} & 
        \pbox{5cm}{0.071 \\ \small{\textit{(250)}}} & 
        \pbox{5cm}{0.042 \\ \small{\textit{(878)}}} & 
        \pbox{5cm}{0.108 \\ \small{\textit{(878)}}} \\
        \rowcolor{gray!10}
        \pbox{5cm}{MSE VA + sigmoid RF \\ learning rate : $10^{-5}$ \\ update rate : 2 \\ $alpha=1$} & 
        \pbox{5cm}{0.416 \\ \small{\textit{(601)}}} & 
        \pbox{5cm}{0.511 \\ \small{\textit{(734)}}} & 
        \pbox{5cm}{0.167 \\ \small{\textit{(4)}}} & 
        \pbox{5cm}{0.221 \\ \small{\textit{(41)}}} & 
        \pbox{5cm}{0.227 \\ \small{\textit{(4)}}} & 
        \pbox{5cm}{0.098 \\ \small{\textit{(169)}}} & 
        \pbox{5cm}{0.035 \\ \small{\textit{(432)}}} & 
        \pbox{5cm}{0.104 \\ \small{\textit{(1)}}} \\
        \hline
        \rowcolor{gray!40}
        \pbox{5cm}{MSE VA + sigmoid RF \\ learning rate : $10^{-5}$ \\ update rate : 5 \\ $alpha=0.9$} & 
        \pbox{5cm}{\textbf{0.491} \\ \small{\textit{(1)}}} & 
        \pbox{5cm}{0.501 \\ \small{\textit{(819)}}} & 
        \pbox{5cm}{0.234 \\ \small{\textit{(1)}}} & 
        \pbox{5cm}{\textbf{0.258} \\ \small{\textit{(1)}}} & 
        \pbox{5cm}{\textbf{0.275} \\ \small{\textit{(1)}}} & 
        \pbox{5cm}{0.085 \\ \small{\textit{(1)}}} & 
        \pbox{5cm}{\textbf{0.030} \\ \small{\textit{(315)}}} & 
        \pbox{5cm}{0.110 \\ \small{\textit{(315)}}} \\
        \rowcolor{gray!10}
        \pbox{5cm}{MSE VA + sigmoid RF \\ learning rate : $10^{-5}$ \\ update rate : 5 \\ $alpha=1$} & 
        \pbox{5cm}{0.435 \\ \small{\textit{(324)}}} & 
        \pbox{5cm}{0.518 \\ \small{\textit{(466)}}} & 
        \pbox{5cm}{0.252 \\ \small{\textit{(324)}}} & 
        \pbox{5cm}{0.242 \\ \small{\textit{(324)}}} & 
        \pbox{5cm}{0.218 \\ \small{\textit{(324)}}} & 
        \pbox{5cm}{0.099 \\ \small{\textit{(383)}}} & 
        \pbox{5cm}{0.058 \\ \small{\textit{(416)}}} & 
        \pbox{5cm}{0.129 \\ \small{\textit{(352)}}} \\
        
        \hline
        \hline
        
        \rowcolor{gray!40}
        \pbox{5cm}{MSE VA + sigmoid RF \\ learning rate : $10^{-4}$ \\ update rate : 2 \\ $alpha=1$ \\ ponderated loss} & \pbox{5cm}{0.420 \\ \small{\textit{(681)}}} & 
        \pbox{5cm}{0.522 \\ \small{\textit{(611)}}} & 
        \pbox{5cm}{0.198 \\ \small{\textit{(2)}}} & 
        \pbox{5cm}{0.202 \\ \small{\textit{(5)}}} & 
        \pbox{5cm}{0.244 \\ \small{\textit{(2)}}} & 
        \pbox{5cm}{0.088 \\ \small{\textit{(278)}}} & 
        \pbox{5cm}{0.034 \\ \small{\textit{(1)}}} & 
        \pbox{5cm}{0.099 \\ \small{\textit{(5)}}} \\
        
        \hline
        
    \end{tabular}
    \end{adjustwidth}
    \caption{Best F1 scores for the Action Units for different parameters of the GAN customized for Valence Arousal and Action Units with the MSE loss function \textit{(first line is the best score, second line in italics with parenthesis is the iteration in thousands at which the best score has been reached, figures in bold are the best scores for all models)}}
    \label{tab:gan_all_results_mse}
\end{table}

\clearpage

\begin{figure}[!ht]
\centering
\begin{tabular}{cccc}
  \includegraphics[scale=2.5]{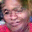} & 
  \includegraphics[scale=2.5]{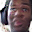} &
 \includegraphics[scale=2.5]{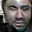} &   
 \includegraphics[scale=2.5]{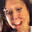} \\
\end{tabular}
\caption{Images created by the Generator (MSE VA + sigmoid AU + RF / learning rate : $10^{-4}$ / update rate : 2 / $\alpha=0.9$) at iteration 147,000, 385,000, 397,000 and 688,000 (from left to right)}
\label{fig:gen_all_mse_img_1}
\end{figure}

\begin{figure}[!ht]
\centering
\begin{tabular}{cccc}
  \includegraphics[scale=2.5]{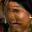} & 
  \includegraphics[scale=2.5]{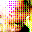} &
 \includegraphics[scale=2.5]{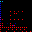} &   
 \includegraphics[scale=2.5]{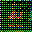} \\
\end{tabular}
\caption{Images created by the Generator (MSE VA + sigmoid AU + RF / learning rate : $10^{-4}$ / update rate : 2 / $\alpha=1$) at iteration 73,000, 147,000, 397,000 and 672,000 (from left to right)}
\label{fig:gen_all_mse_img_2}
\end{figure}

\begin{figure}[!ht]
\centering
\begin{tabular}{cccc}
  \includegraphics[scale=2.5]{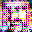} & 
  \includegraphics[scale=2.5]{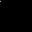} &
 \includegraphics[scale=2.5]{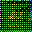} &   
 \includegraphics[scale=2.5]{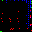} \\
\end{tabular}
\caption{Images created by the Generator (MSE VA + sigmoid AU + RF / learning rate : $10^{-4}$ / update rate : 5 / $\alpha=0.9$) at iteration 147,000, 385,000, 672,000 and 999,000 (from left to right)}
\label{fig:gen_all_mse_img_3}
\end{figure}

\begin{figure}[!ht]
\centering
\begin{tabular}{cccc}
  \includegraphics[scale=2.5]{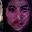} & 
  \includegraphics[scale=2.5]{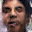} &
 \includegraphics[scale=2.5]{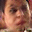} &   
 \includegraphics[scale=2.5]{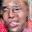} \\
\end{tabular}
\caption{Images created by the Generator (MSE VA + sigmoid AU + RF / learning rate : $10^{-4}$ / update rate : 5 / $\alpha=1$) at iteration 73,000, 261,000, 688,000 and 693,000 (from left to right)}
\label{fig:gen_all_mse_img_4}
\end{figure}

\begin{figure}[!ht]
\centering
\begin{tabular}{cccc}
  \includegraphics[scale=2.5]{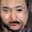} & 
  \includegraphics[scale=2.5]{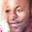} &
 \includegraphics[scale=2.5]{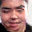} &   
 \includegraphics[scale=2.5]{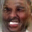} \\
\end{tabular}
\caption{Images created by the Generator (MSE VA + sigmoid AU + RF / learning rate : $10^{-5}$ / update rate : 2 / $\alpha=0.9$) at iteration 147,000, 261,000, 397,000 and 691,000 (from left to right)}
\label{fig:gen_all_mse_img_5}
\end{figure}

\begin{figure}[!ht]
\centering
\begin{tabular}{cccc}
  \includegraphics[scale=2.5]{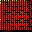} & 
  \includegraphics[scale=2.5]{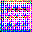} &
 \includegraphics[scale=2.5]{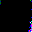} &   
 \includegraphics[scale=2.5]{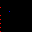} \\
\end{tabular}
\caption{Images created by the Generator (MSE VA + sigmoid AU + RF / learning rate : $10^{-5}$ / update rate : 2 / $\alpha=1$) at iteration 147,000, 261,000, 397,000 and 999,000 (from left to right)}
\label{fig:gen_all_mse_img_6}
\end{figure}

\begin{figure}[!ht]
\centering
\begin{tabular}{cccc}
  \includegraphics[scale=2.5]{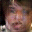} & 
  \includegraphics[scale=2.5]{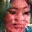} &
 \includegraphics[scale=2.5]{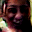} &   
 \includegraphics[scale=2.5]{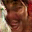} \\
\end{tabular}
\caption{Images created by the Generator (MSE VA + sigmoid AU + RF / learning rate : $10^{-5}$ / update rate : 5 / $\alpha=0.9$) at iteration 147,000, 385,000, 646,000 and 999,000 (from left to right)}
\label{fig:gen_all_mse_img_7}
\end{figure}

\begin{figure}[!ht]
\centering
\begin{tabular}{cccc}
  \includegraphics[scale=2.5]{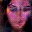} & 
  \includegraphics[scale=2.5]{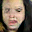} &
 \includegraphics[scale=2.5]{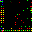} &   
 \includegraphics[scale=2.5]{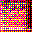} \\
\end{tabular}
\caption{Images created by the Generator (MSE VA + sigmoid AU + RF / learning rate : $10^{-5}$ / update rate : 5 / $\alpha=1$) at iteration 147,000, 385,000, 672,000 and 688,000 (from left to right)}
\label{fig:gen_all_mse_img_8}
\end{figure}

\begin{figure}[!ht]
\centering
\begin{tabular}{cccc}
  \includegraphics[scale=2.5]{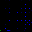} & 
  \includegraphics[scale=2.5]{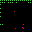} &
 \includegraphics[scale=2.5]{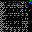} &   
 \includegraphics[scale=2.5]{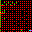} \\
\end{tabular}
\caption{Images created by the Generator (MSE VA + sigmoid AU + RF / learning rate : $10^{-4}$ / update rate : 2 / $\alpha=1$ / ponderated loss) at iteration 147,000, 385,000, 646,000 and 999,000 (from left to right)}
\label{fig:gen_all_mse_img_9}
\end{figure}

\clearpage

\subsubsection{Analysis}

First, we analyse the results for the 1-CCC loss function. 
For the CCC scores of Valence Arousal, the best model is for a learning rate of $10^{-5}$ and an update rate of 5 (0.837 and 0.764 respectively). We can notice that the Valence score is also equalized for the same learning rate and update rate but with a ponderated loss. This ponderated loss performs best for the Valence MSE score (0.066) while the non ponderated loss is better for Arousal MSE score (0.035).  These scores are quite similar to those obtained with the GAN for Valence Arousal only with the same loss function (1-CCC) but a little under the scores achieved with the MSE loss function, as we could have expected from the previous experiments with the GAN customized for Valence Arousal. The best percentage of images classified as real is for a learning rate of $10^{-4}$ and an update rate of 7, with a high score of 0.999. Weighting the loss function differently does not change the scores much for Valence Arousal. \\
Contrary to the models for Action Units only, there is not one model performing the best scores for all Action Units. The model with a learning rate of $10^{-5}$ and an update rate of 2 performs best for the F1 score of the Action Unit 1, 2, 12, 20 and 25 and for the F1 score and the mean between the F1 score and the accuracy. Two ponderated models perform best for the F1 score of the Action Unit 4 and 15. For the Action Unit 4 the improvement is quite significant. We can see that the scores for Action Units are really low compared to those obtained with the GAN for Action Units only, especially for the Action Units 15 and 20 which are the less frequent Action Units. The Action Units more frequent in the dataset like AU 1 and 2 perform best. \\
For all the given images produced by the Generator (\ref{fig:gen_all_ccc_img_1}, \ref{fig:gen_all_ccc_img_2}, \ref{fig:gen_all_ccc_img_3}, \ref{fig:gen_all_ccc_img_4}, \ref{fig:gen_all_ccc_img_5}, \ref{fig:gen_all_ccc_img_6}, \ref{fig:gen_all_ccc_img_7}, \ref{fig:gen_all_ccc_img_8}), the lower the update rate the better the images. Moreover, when the update rate increases the noise is more visible (\ref{fig:gen_all_ccc_img_3}, \ref{fig:gen_all_ccc_img_6} for example). 
\\
\\
For the MSE loss function, we can see that the best Valence Arousal CCC scores are achieved by the model with the learning rate $10^{-4}$, an update rate 2 and $alpha=1$. The best scores for the VA MSE scores are achieved by the model with the learning rate $10^{-5}$, the update rate of 2 and $alpha=1$. The best percentage of images classified as real is achieved by the model with the same learning rate but with an update rate of 2 and $alpha=0.9$. \\
For the Action Units F1 scores there is not only model performing best but two models are distinguishable : the model with a learning rate of $10^{-5}$ and an update rate of 5 performs best for the Action 1, 6 and 12 and the model with the same update rate and a learning rate of $10^{-4}$ performs best for the Action Unit 4, 15 and 20. Adjusting the weights of the Discriminator loss function in favor of the Action Units did not help get better score. \\
The images created on Figures \ref{fig:gen_all_mse_img_1}, \ref{fig:gen_all_mse_img_4} and \ref{fig:gen_all_mse_img_5} are the best generated pictures. Depending on the update rate the influence of the label smoothing (i.e. $alpha=0.9$) is inverted : when the update rate is 2, with label smoothing the images generated are looking good (\ref{fig:gen_all_mse_img_1}) while the images generated without label smoothing are noisy (\ref{fig:gen_all_mse_img_2}). This effect is also observed for a learning rate of $10^{-5}$ and an update rate of 2 and 5. On the other hand, for the update rate of 5 when there is label smoothing only noise is generated (\ref{fig:gen_all_mse_img_3}) while without label smoothing good looking pictures are created (\ref{fig:gen_all_mse_img_4}). We can see that the different weighting in the Generator loss function did not change anything between the images generated (\ref{fig:gen_all_mse_img_2} and \ref{fig:gen_all_mse_img_9}) : they are still noisy. Surprisingly, we can also notice that when the Arousal score is low (around 0.5) the images generated by the Generator are better. This can be seen for the model with the learning rate $10^{-4}$, the update rate 2 and $alpha=0.9$ ($CCC(arousal)=0.519$) and the images \ref{fig:gen_all_mse_img_1}, the model with the learning rate $10^{-4}$, the update rate 5 and $alpha=1$ ($CCC(arousal)=0.529$) and the images \ref{fig:gen_all_mse_img_4}, the model with the learning rate $10^{-4}$, the update rate 2 and $alpha=0.9$ ($CCC(arousal)=0.519$) and the images \ref{fig:gen_all_mse_img_5} and the model with the learning rate $10^{-5}$, the update rate 5 and $alpha=0.9$ ($CCC(arousal)=0.523$) and the images \ref{fig:gen_all_mse_img_5}. This might show the adversarial aspect of the GAN : when the Discriminator gets worse (the CCC for Arousal lowers) the Generator gets better.  \\
\\
When comparing the two loss functions we can see that the results for the 1-CCC loss function and the MSE loss function are in the same range and there is not a more suitable loss function for the GAN customized for Action Units and Valence Arousal. However for the images the observation is different. The 1-CCC loss function seems to help generate images of faces more easily while the MSE loss function seems to help create better images but depending on the chosen parameters the model can collapse and be unable to generate any images of faces.

\newpage

\section{Results \& Comparisons}

For both Action Units and Valence Arousal, we can see that the GAN customized specifically for one model, i.e. Action Units or Valence Arousal performs best than the GAN including both models. We might assume that this is due to the complexity of the data : the GAN has to predict for both a non exclusive classification problem for 8 Action Units, a regression problem with Valence and Arousal and if the image is fake or real. The results for the GANs dedicated for Action Units and Valence Arousal are interesting because the images are only 28*28 pixels and so it is difficult to distinguish the Action Units and Valence Arousal. Moreover, the Discriminator of the GAN is only 4-layer deep which is not very deep to get the complexity of the dataset. This even more true for the GAN assembling both models and it can explain why the results for the Action Units drop dramatically. \\
\\

\begin{figure}[!ht]
\centering
\begin{tabular}{cccc}
 \includegraphics[scale=0.45]{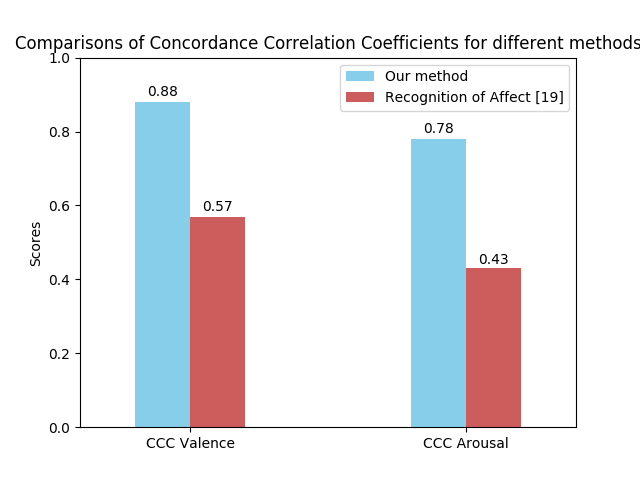} &
  \includegraphics[scale=0.45]{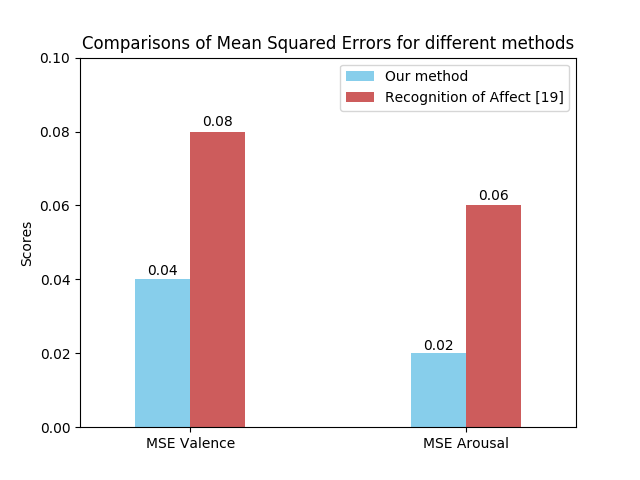} \\
\end{tabular}
\caption{Comparisons of the Concordance Correlation Coefficients and the MSE for our best method and the best method for 'Recognition of Affect in the wild using Deep Neural Networks' \cite{kollias2}}
\label{fig:va_comp}
\end{figure}

The CCC and MSE scores obtained with our best method is better than the methods proposed in 'Recognition of Affect in the wild using Deep Neural Networks' \cite{kollias2} as shown in Figure \ref{fig:va_comp}. This comparison is interesting because they have been obtained from the same database (Aff-Wild). However these results must be minimized for the following reasons :
\begin{itemize}
    \item The newly created dataset was only part of the Aff-Wild database and not the whole;
    \item The training and testing sets are not the same as the ones used in the paper \cite{kollias2} we compare;
    \item Data processing has been done on this new dataset like changing the fps of the videos to 30 or selecting the relevant frames.
\end{itemize}

Comparing results for the Action Units with the papers of the background like 'Deep Learning the Dynamic Appearance and Shape of Facial Action Units' \cite{RefWorks:doc:5af1cc28e4b0011b0775e7ee} or 'Transfer Learning for Action Unit Recognition' \cite{RefWorks:doc:5af48017e4b0f7bd1fabb8de} is not possible because these papers worked on different databases : BP4D and SEMAINE for \cite{RefWorks:doc:5af1cc28e4b0011b0775e7ee} and DISFA for \cite{RefWorks:doc:5af48017e4b0f7bd1fabb8de}. As these Action Unit annotations has been added for this project, there is no way of comparing these results with a rigorous method.

On the other part of the GAN, the Generator is able to produce good images for each type of specialized GAN : for the GAN customized for Action Units the Figure \ref{fig:gen_au_img_2} show the best possible face-looking images, for the GAN dedicated to Valence Arousal the Figures \ref{fig:gen_mse_img_1}, \ref{fig:gen_mse_img_3} and \ref{fig:gen_ccc_img_1} are the best pictures created and for the GAN gathering both models the Figures \ref{fig:gen_all_mse_img_1}, \ref{fig:gen_all_mse_img_4} and \ref{fig:gen_all_mse_img_5} are the best pictures of faces generated. The results are decent taking into account the size of the images taken as inputs and generated (28*28 pixels) and the number of layers for the Generator (4). 

\section{Improvements \& possible future work}

The improvements to bring would be to the GAN gathering the Action Units model and the Valence Arousal model. Indeed, even though good results were obtained for Valence Arousal, the results for Action Units were too low. As a result, different ways of improving this GAN would be :
\begin{itemize}
    \item Train the architectures tested for more than 1,000,000 iterations. We noticed that some of the best results were achieved at late iterations, so training the GANs longer may improve its performance;
    \item Trying to remove the label smoothing (i.e. put $alpha=1$) which was a default parameter as best results were achieved without label smoothing for Valence Arousal;
    \item Put even more weight on the Action Unit loss function in the final loss function so as to give more importance to the Action Units loss function and as a result improve the scores for this model.
\end{itemize}

The possible future work on this project could be :
\begin{itemize}
    \item Expand the dataset so as to have better generalization properties and prevent overfitting. A more important dataset could improve the scores;
    \item In the meantime, layers could be added to the Generator and the Discriminator neural networks so that they can learn the more vast and complex dataset;
    \item Increase the resolution of images to 64*64 first, then to 96*96. It would be interesting to see what images the Generator would produce with a better resolution. Moreover images with a better resolution may improve the Discriminator predictions especially for the Action Units combined with Valence Arousal provided that its number of layers is increased;
    \item Finally, it would be interesting to incorporate in the Discriminator a well-known efficient neural network like VGG or ResNet. We might expect an increase in the results for classifying the Action Units and regressing the Valence Arousal as well as predicting if an image is real or fake.
\end{itemize}

\chapter{Conclusion}

This project was challenging because it aimed at assembling two existing models for emotion recognition : Facial Action Units which consists of muscle movements on faces and Valence Arousal where Valence represents how much a person is positive or negative and Arousal depicts how much a person is active or passive. Moreover, this project proposed an original way of approaching this subject using Generative Adversarial Networks (GANs) while the best scores achieved for both models were performed by more classic neural networks like VGG-Face, VGG-16 or ResNet 50 for example (\cite{RefWorks:doc:5af1a263e4b02dbdaeebbecb}, \cite{RefWorks:doc:5b0e689ee4b02f452d8e6de3}, \cite{kollias2} for Valence Arousal and \cite{RefWorks:doc:5af1cc28e4b0011b0775e7ee}, \cite{RefWorks:doc:5af48017e4b0f7bd1fabb8de} for Action Units). \\
\\
In order to assemble the two existing models, i.e. Action Units and Valence Arousal, a new dataset had to be created. The Aff-Wild videos made up of YouTube videos in-the-wild was already annotated for Valence Arousal. Some videos from this dataset were chosen according to the 8 Action Units that have been chosen to study : Action Unit 1 (Inner brow raiser), Action Unit 2 (Outer brow raiser), Action Unit 4 (Brow lowerer), Action Unit 6 (Cheek raiser), Action Unit 12 (Lip corner puller), Action Unit 15 (Lip corner depressor), Action Unit 20 (Lip stretcher) and Action Unit 25 (Lips part). After two stages of annotations, the fusion of the two types of annotations, and some data processing and cleaning, the new dataset with 64 videos was ready. Even though this phase of the project was long, it was also interesting to better understand what is at stake when a dataset is created. \\
\\
Three different architectures of GANs were then implemented : one architecture customized for Action Units only, another for Valence Arousal only, and a last one that gathered together the two models. This approach enabled to better understand how GANs work and test different sets of hyperparameters and loss functions performing best for each model. The particularity of these GANs was that the Discriminator did not have to only classify the images as real or fake but also had to determine which Action Units were present on the pictures and the values for Valence Arousal. This is what we call a categorical GAN. The Discriminator of the GAN performed well when training for Action Units only and for Valence Arousal only with scores even better than the papers taken as reference in the background. However when the two models (Action Units and Valence Arousal) were assembled the scores for Action Units collapsed but the scores for Valence Arousal stayed same at the same level as for the customized VA model. This means that there is still work to do so as to better assemble these two models so that they achieve the same scores as the customized architectures or even better. The Generator was also able to generate good looking images considering the fact that it has only four layers and takes as references images of 28*28 pixels. A future work to carry out would be to try to generate images with better resolution.

\appendix
\chapter{Ethics checklist}

\begin{figure}[h!]
\centering
\includegraphics[page=1, scale = 0.85]{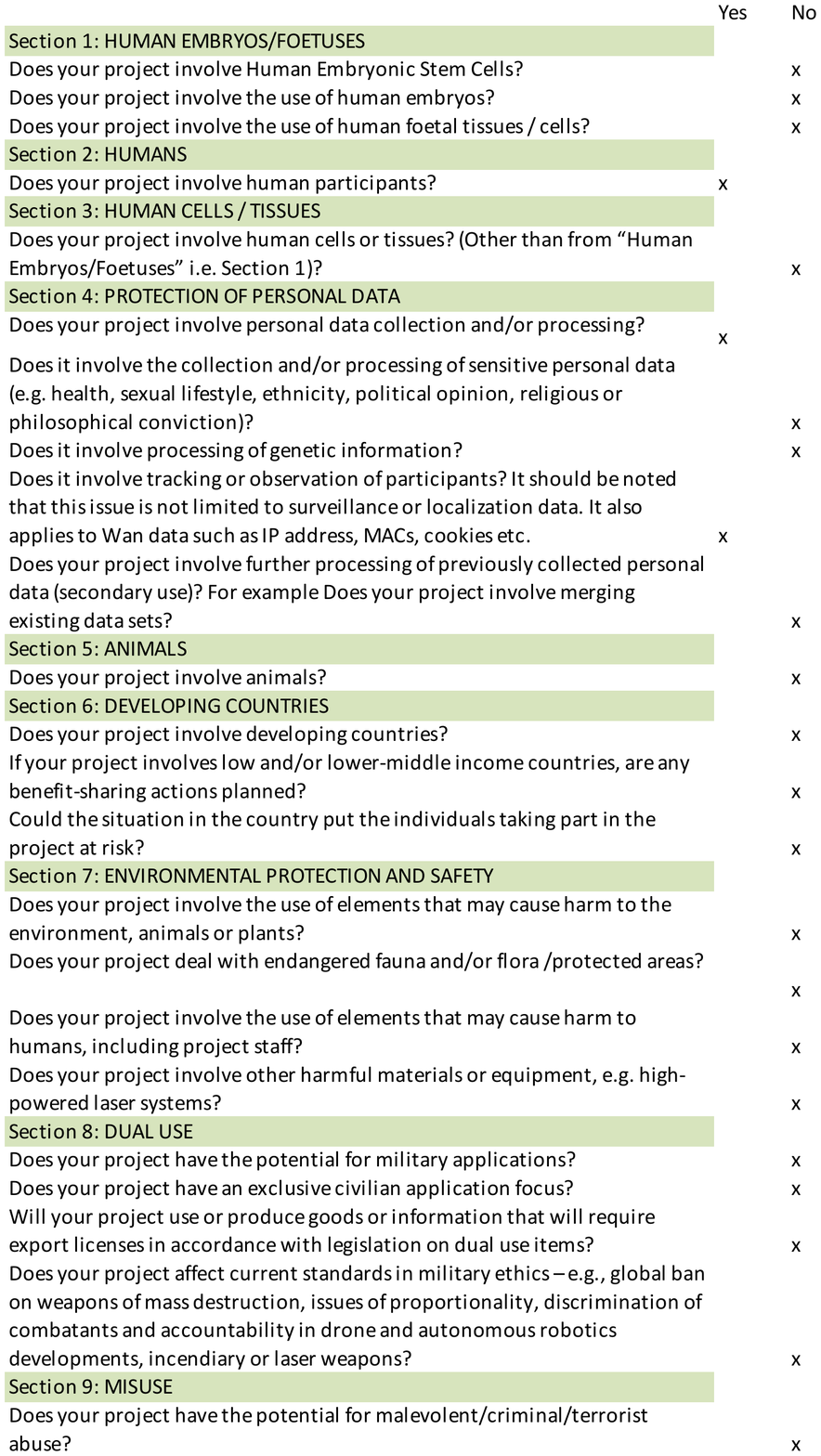}
\captionsetup{labelformat=empty}
\caption{}
\end{figure}
\begin{figure}[h!]
\centering
\includegraphics[page=2, scale = 0.85]{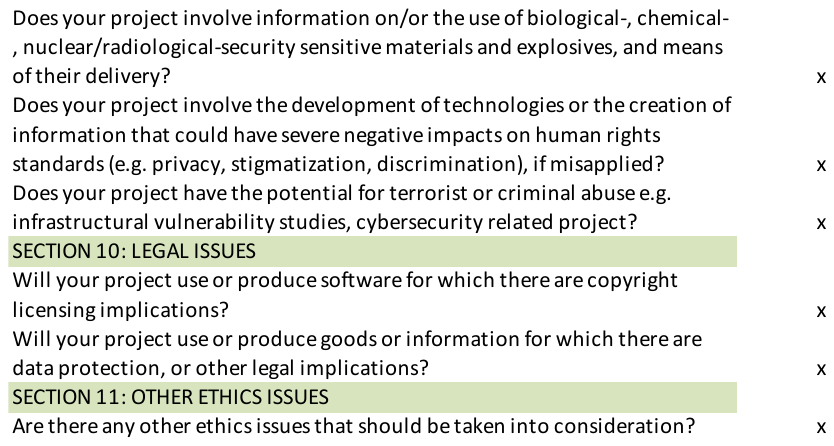}
\captionsetup{labelformat=empty}
\caption{}
\end{figure}

\chapter{Ethical and professional considerations}

This project involves human participants because the Aff-Wild dataset is YouTube videos of people. Thus this project is about the observation of participants, especially their face movements due to their reactions. The annotations like Valence and Arousal and Actions Units made on these videos are then processed for analysis (like training a neural network) and with the goal of detecting these annotations in non-labelled videos. The collection of these videos had already been done previously for the construction of the dataset Aff-Wild. People were asked the permission to use their video for this dataset \cite{kollias1}. Furthermore, the impact this project had on the environment could be taken into account. Indeed, deep learning needs a lot of GPU ressources to train the neural networks. However, the consequences can be played down compared to other industries. \\
No personal data is collected or treated. This project does not involve human embryos/foetuses, human cells/tissues, animals, subjects related to developing countries, environmental or safety issues. Finally, the possible dual use, misuse, legal issues, or other ethics issues are very limited.



{\small
\bibliographystyle{ieee_fullname}
\bibliography{export}

\begin{thebibliography}{10}\itemsep=-1pt

\bibitem{RefWorks:doc:5b8c0195e4b07993ca600a55}
Convolution arithmetic tutorial.

\bibitem{RefWorks:doc:5b7abccce4b00447110ca60d}
Menpo.

\bibitem{RefWorks:doc:5b8aa565e4b036495fd78c2c}
Menpo detect - load ffld2 frontal face detector.

\bibitem{RefWorks:doc:5b8e9cc0e4b0ad32cc6389f3}
Tensorflow.

\bibitem{RefWorks:doc:5b15422ce4b03a529bc77e56}
André~Christoffer Andersen.
\newblock Lstm with n units.

\bibitem{avrithis2000broadcast}
Yannis Avrithis, Nicolas Tsapatsoulis, and Stefanos Kollias.
\newblock Broadcast news parsing using visual cues: A robust face detection
  approach.
\newblock In {\em 2000 IEEE International Conference on Multimedia and Expo.
  ICME2000. Proceedings. Latest Advances in the Fast Changing World of
  Multimedia (Cat. No. 00TH8532)}, volume~3, pages 1469--1472. IEEE, 2000.

\bibitem{RefWorks:doc:5b0c139ae4b05e315bed2039}
M.~S. Bartlett, G. Littlewort, M. Frank, C. Lainscsek, I. Fasel, and J.
  Movellan.
\newblock Fully automatic facial action recognition in spontaneous behavior.
\newblock pages 223--230. IEEE, 2006.

\bibitem{RefWorks:doc:5b082aaae4b0f72b00d4ca73}
Tanja Bänziger and Klaus~R. Scherer.
\newblock Introducing the geneva multimodal emotion portrayal (gemep) corpus.

\bibitem{RefWorks:doc:5af47ed9e4b0bb78c0e4a778}
Junyoung Chung, Caglar Gulcehre, KyungHyun Cho, and Yoshua Bengio.
\newblock Empirical evaluation of gated recurrent neural networks on sequence
  modeling.
\newblock Dec 11, 2014.

\bibitem{RefWorks:doc:5b154169e4b03a529bc77e3c}
Adit Deshpande.
\newblock A beginner's guide to understanding convolutional neural networks,
  July 20, 2016.

\bibitem{RefWorks:doc:5b0ec87be4b0ca9da8b4fa87}
Friesen Wallace V. Hager Joseph~C. Ekman, Paul.
\newblock {\em Facial action coding system}.
\newblock A Human Face, Douglas (Arizona), 2002.
\newblock ID: 932790409.

\bibitem{RefWorks:doc:5b0ec7f2e4b02f452d8e78c0}
Paul Ekman and Wallace~V. Friesen.
\newblock Measuring facial movement.
\newblock {\em Environmental Psychology and Nonverbal Behavior}, 1(1):56--75,
  1976.

\bibitem{glimm2013using}
Birte Glimm, Yevgeny Kazakov, Ilianna Kollia, and Giorgos~B Stamou.
\newblock Using the tbox to optimise sparql queries.
\newblock {\em Description Logics}, 1014:181--196, 2013.

\bibitem{goudelis2013exploring}
Georgios Goudelis, Konstantinos Karpouzis, and Stefanos Kollias.
\newblock Exploring trace transform for robust human action recognition.
\newblock {\em Pattern Recognition}, 46(12):3238--3248, 2013.

\bibitem{RefWorks:doc:5af42248e4b0f7b951da62fe}
Hatice Gunes and Maja Pantic.
\newblock Automatic, dimensional and continuous emotion recognition.
\newblock {\em International Journal of Synthetic Emotions (IJSE)},
  1(1):68--99, Jan 1, 2010.

\bibitem{RefWorks:doc:5b0e689ee4b02f452d8e6de3}
Kaiming He, Xiangyu Zhang, Shaoqing Ren, and Jian Sun.
\newblock Deep residual learning for image recognition.
\newblock Dec 10, 2015.

\bibitem{RefWorks:doc:5b0fd825e4b0a35b73cb39e4}
Stephanie~C. Herring, Nikolaos Christidi, Andrew Hoell, James~P. Kossin, Carl
  J.~Schreck III, and Peter~A. Stott.
\newblock Understanding the difficulty of training deep feedforward neural
  networks.
\newblock {\em Bulletin of the American Meteorological Society}, 99(1):Sii, Jan
  1, 2018.

\bibitem{horrocks2011answering}
Ilianna Kollia Birte Glimm~Ian Horrocks.
\newblock Answering queries over owl ontologies with sparql.
\newblock 2011.

\bibitem{RefWorks:doc:5af42188e4b07aa41af12318}
Mehdi Mirza Bing Xu David Warde-Farley Sherjil Ozair† Aaron Courville
  Yoshua~Bengio‡ Ian J.~Goodfellow, Jean Pouget-Abadie∗.
\newblock Generative adversarial nets, June 2014.

\bibitem{RefWorks:doc:5af1cc28e4b0011b0775e7ee}
Shashank Jaiswal and Michel Valstar.
\newblock Deep learning the dynamic appearance and shape of facial action
  units.
\newblock pages 1--8. IEEE, March 2016.

\bibitem{RefWorks:doc:5af1a263e4b02dbdaeebbecb}
Pooya Khorrami, Tom~Le Paine, Kevin Brady, Charlie Dagli, and Thomas~S. Huang.
\newblock How deep neural networks can improve emotion recognition on video
  data.
\newblock Feb 23, 2016.

\bibitem{RefWorks:doc:5b0d2daae4b0c6f743c5ef95}
Davis~E. King.
\newblock Dlib-ml: A machine learning toolkit.

\bibitem{kollia2009interweaving}
Ilianna Kollia, Nikolaos Simou, Giorgos Stamou, and Andreas Stafylopatis.
\newblock Interweaving knowledge representation and adaptive neural networks.
\newblock In {\em Workshop on Inductive Reasoning and Machine Learning on the
  Semantic Web}, 2009.

\bibitem{kollias8}
Dimitrios Kollias, Shiyang Cheng, Maja Pantic, and Stefanos Zafeiriou.
\newblock Photorealistic facial synthesis in the dimensional affect space.
\newblock In {\em Proceedings of the European Conference on Computer Vision
  (ECCV)}, pages 0--0, 2018.

\bibitem{kollias9}
Dimitrios Kollias, Shiyang Cheng, Evangelos Ververas, Irene Kotsia, and
  Stefanos Zafeiriou.
\newblock Generating faces for affect analysis.
\newblock {\em arXiv preprint arXiv:1811.05027}, 2018.

\bibitem{kollias10}
Dimitris Kollias, George Marandianos, Amaryllis Raouzaiou, and Andreas-Georgios
  Stafylopatis.
\newblock Interweaving deep learning and semantic techniques for emotion
  analysis in human-machine interaction.
\newblock In {\em 2015 10th International Workshop on Semantic and Social Media
  Adaptation and Personalization (SMAP)}, pages 1--6. IEEE, 2015.

\bibitem{kollias2}
Dimitrios Kollias, Mihalis~A Nicolaou, Irene Kotsia, Guoying Zhao, and Stefanos
  Zafeiriou.
\newblock Recognition of affect in the wild using deep neural networks.
\newblock In {\em Proceedings of the IEEE Conference on Computer Vision and
  Pattern Recognition Workshops}, pages 26--33, 2017.

\bibitem{kollias11}
Dimitrios Kollias, Athanasios Tagaris, and Andreas Stafylopatis.
\newblock On line emotion detection using retrainable deep neural networks.
\newblock In {\em 2016 IEEE Symposium Series on Computational Intelligence
  (SSCI)}, pages 1--8. IEEE, 2016.

\bibitem{kollias13}
Dimitrios Kollias, Athanasios Tagaris, Andreas Stafylopatis, Stefanos Kollias,
  and Georgios Tagaris.
\newblock Deep neural architectures for prediction in healthcare.
\newblock {\em Complex \& Intelligent Systems}, 4(2):119--131, 2018.

\bibitem{kollias3}
Dimitrios Kollias, Panagiotis Tzirakis, Mihalis~A Nicolaou, Athanasios
  Papaioannou, Guoying Zhao, Bj{\"o}rn Schuller, Irene Kotsia, and Stefanos
  Zafeiriou.
\newblock Deep affect prediction in-the-wild: Aff-wild database and challenge,
  deep architectures, and beyond.
\newblock {\em International Journal of Computer Vision}, 127(6-7):907--929,
  2019.

\bibitem{kollias12}
Dimitrios Kollias, Miao Yu, Athanasios Tagaris, Georgios Leontidis, Andreas
  Stafylopatis, and Stefanos Kollias.
\newblock Adaptation and contextualization of deep neural network models.
\newblock In {\em 2017 IEEE Symposium Series on Computational Intelligence
  (SSCI)}, pages 1--8. IEEE, 2017.

\bibitem{kollias4}
Dimitrios Kollias and Stefanos Zafeiriou.
\newblock Aff-wild2: Extending the aff-wild database for affect recognition.
\newblock {\em arXiv preprint arXiv:1811.07770}, 2018.

\bibitem{kollias7}
Dimitrios Kollias and Stefanos Zafeiriou.
\newblock A multi-component cnn-rnn approach for dimensional emotion
  recognition in-the-wild.
\newblock {\em arXiv preprint arXiv:1805.01452}, 2018.

\bibitem{kollias5}
Dimitrios Kollias and Stefanos Zafeiriou.
\newblock A multi-task learning \& generation framework: Valence-arousal,
  action units \& primary expressions.
\newblock {\em arXiv preprint arXiv:1811.07771}, 2018.

\bibitem{kollias6}
Dimitrios Kollias and Stefanos Zafeiriou.
\newblock Training deep neural networks with different datasets in-the-wild:
  The emotion recognition paradigm.
\newblock In {\em 2018 International Joint Conference on Neural Networks
  (IJCNN)}, pages 1--8. IEEE, 2018.

\bibitem{kollias14}
Dimitrios Kollias and Stefanos Zafeiriou.
\newblock Exploiting multi-cnn features in cnn-rnn based dimensional emotion
  recognition on the omg in-the-wild dataset.
\newblock {\em arXiv preprint arXiv:1910.01417}, 2019.

\bibitem{kollias15}
Dimitrios Kollias and Stefanos Zafeiriou.
\newblock Expression, affect, action unit recognition: Aff-wild2, multi-task
  learning and arcface.
\newblock {\em arXiv preprint arXiv:1910.04855}, 2019.

\bibitem{RefWorks:doc:5b0c082ee4b0562e289ce43d}
Jean Kossaifi, Georgios Tzimiropoulos, Sinisa Todorovic, and Maja Pantic.
\newblock Afew-va database for valence and arousal estimation in-the-wild.
\newblock {\em Image and Vision Computing}, 65:23--36, Sep 2017.

\bibitem{RefWorks:doc:5b116947e4b006165bc77480}
Slawomir Koziel.
\newblock Computational optimization, methods and algorithms, 2011.

\bibitem{RefWorks:doc:5b0e6651e4b01f2c3e37bcbb}
Alex Krizhevsky, Ilya Sutskever, and Geoffrey Hinton.
\newblock Imagenet classification with deep convolutional neural networks, May
  24, 2017.

\bibitem{RefWorks:doc:5af48017e4b0f7bd1fabb8de}
Yen~Khye Lim, Zukang Liao, Stavros Petridis, and Maja Pantic.
\newblock Transfer learning for action unit recognition.
\newblock 2016.

\bibitem{RefWorks:doc:5b07d52ce4b0262a90ba52d7}
Patrick Lucey, Jeffrey~F. Cohn, Takeo Kanade, Jason Saragih, Zara Ambadar, and
  Iain Matthews.
\newblock The extended cohn-kanade dataset (ck+): A complete dataset for action
  unit and emotion-specified expression.
\newblock pages 94--101, 2010.

\bibitem{RefWorks:doc:5b083834e4b0357468f0b292}
Patrick Lucey, Jeffrey~F. Cohn, Kenneth~M. Prkachin, Patricia~E. Solomon, and
  Iain Matthews.
\newblock Painful data: The unbc-mcmaster shoulder pain expression archive
  database.
\newblock pages 57--64, 2011.

\bibitem{RefWorks:doc:5af42217e4b0cebb7adfb215}
Brais Martinez, Michel~F. Valstar, Bihan Jiang, and Maja Pantic.
\newblock Automatic analysis of facial actions: A survey.
\newblock {\em IEEE Transactions on Affective Computing}, PP(99):1, September
  2014.

\bibitem{RefWorks:doc:5b8aa62ce4b01735413d84df}
Markus Mathias, Rodrigo Benenson, Marco Pedersoli, and Luc~Van Gool.
\newblock Face detection without bells and whistles.

\bibitem{RefWorks:doc:5b08331be4b0353e399a5f01}
S.~M. Mavadati, M.~H. Mahoor, K. Bartlett, P. Trinh, and J.~F. Cohn.
\newblock Disfa: A spontaneous facial action intensity database.
\newblock {\em IEEE Transactions on Affective Computing}, 4(2):151--160, 2013.

\bibitem{RefWorks:doc:5b0c15b9e4b08a9fe9335907}
Daniel McDuff, Rana el Kaliouby, Thibaud Senechal, May Amr, Jeffrey~F. Cohn,
  and Rosalind Picard.
\newblock Affectiva-mit facial expression dataset (am-fed): Naturalistic and
  spontaneous facial expressions collected in-the-wild.
\newblock In {\em Proceedings of the 2013 IEEE Conference on Computer Vision
  and Pattern Recognition Workshops}, CVPRW '13, pages 881--888, Washington,
  DC, USA, 2013. IEEE Computer Society.

\bibitem{RefWorks:doc:5b07ebbce4b0f7c767179557}
G. McKeown, M. Valstar, R. Cowie, M. Pantic, and M. Schroder.
\newblock The semaine database: Annotated multimodal records of emotionally
  colored conversations between a person and a limited agent.
\newblock {\em IEEE Transactions on Affective Computing}, 3(1):5--17, 2012.

\bibitem{RefWorks:doc:5b0c04c9e4b08a9fe9335675}
Hongying Meng, Di Huang, Heng Wang, Hongyu Yang, Mohammed AI-Shuraifi, and
  Yunhong Wang.
\newblock Depression recognition based on dynamic facial and vocal expression
  features using partial least square regression.
\newblock AVEC '13, pages 21--30. ACM, Oct 21, 2013.

\bibitem{mylonas2009using}
Phivos Mylonas, Evaggelos Spyrou, Yannis Avrithis, and Stefanos Kollias.
\newblock Using visual context and region semantics for high-level concept
  detection.
\newblock {\em IEEE Transactions on Multimedia}, 11(2):229--243, 2009.

\bibitem{RefWorks:doc:5af1a0ffe4b0ac09013ce59e}
Augustus Odena.
\newblock Semi-supervised learning with generative adversarial networks.
\newblock Jun 5, 2016.

\bibitem{RefWorks:doc:5b154078e4b014266afb5926}
Christopher Olah.
\newblock Understanding lstm networks, Aug 27, 2015.

\bibitem{RefWorks:doc:5b0c0ca2e4b0542f750169d4}
M. Pantic, M. Valstar, R. Rademaker, and L. Maat.
\newblock Web-based database for facial expression analysis.
\newblock page 5 pp. IEEE, 2005.

\bibitem{RefWorks:doc:5b0ff4e6e4b01f2c3e37e5ea}
Omkar Parkhi, Andrew Zisserman, and Andrea Vedaldi.
\newblock Deep face recognition.
\newblock {\em International Journal of Multimedia Information Retrieval},
  4(2):75--93, Jun 2015.

\bibitem{raftopoulos2018beneficial}
Konstantinos~A Raftopoulos, Stefanos~D Kollias, Dionysios~D Sourlas, and Marin
  Ferecatu.
\newblock On the beneficial effect of noise in vertex localization.
\newblock {\em International Journal of Computer Vision}, 126(1):111--139,
  2018.

\bibitem{RefWorks:doc:5b07f529e4b0407f82f771b2}
Fabien Ringeval, Andreas Sonderegger, Juergen Sauer, and Denis Lalanne.
\newblock Introducing the recola multimodal corpus of remote collaborative and
  affective interactions.
\newblock pages 1--8. IEEE, 2013.

\bibitem{RefWorks:doc:5b0e67fce4b0a35b73cb168b}
Karen Simonyan and Andrew Zisserman.
\newblock Very deep convolutional networks for large-scale image recognition.
\newblock Sep 4, 2014.

\bibitem{simou2008image}
Nikos Simou, Th Athanasiadis, Giorgos Stoilos, and Stefanos Kollias.
\newblock Image indexing and retrieval using expressive fuzzy description
  logics.
\newblock {\em Signal, Image and Video Processing}, 2(4):321--335, 2008.

\bibitem{simou2007fire}
Nikolaos Simou and Stefanos Kollias.
\newblock Fire: A fuzzy reasoning engine for impecise knowledge.
\newblock In {\em K-Space PhD Students Workshop, Berlin, Germany}, volume~14.
  Citeseer, 2007.

\bibitem{RefWorks:doc:5af1a0d1e4b0155360c56aa1}
Jost~Tobias Springenberg.
\newblock Unsupervised and semi-supervised learning with categorical generative
  adversarial networks.
\newblock Nov 19, 2015.

\bibitem{RefWorks:doc:5b8924f2e4b036495fd77f4d}
Shao-Hua Sun.
\newblock Semi-supervised learning gan in tensorflow.

\bibitem{RefWorks:doc:5af47bc8e4b0ab1e65d9dd33}
Christian Szegedy, Wei Liu, Yangqing Jia, Pierre Sermanet, Scott Reed, Dragomir
  Anguelov, Dumitru Erhan, Vincent Vanhoucke, and Andrew Rabinovich.
\newblock Going deeper with convolutions.
\newblock Sep 16, 2014.

\bibitem{tagaris1}
Athanasios Tagaris, Dimitrios Kollias, and Andreas Stafylopatis.
\newblock Assessment of parkinson’s disease based on deep neural networks.
\newblock In {\em International Conference on Engineering Applications of
  Neural Networks}, pages 391--403. Springer, 2017.

\bibitem{tagaris2}
Athanasios Tagaris, Dimitrios Kollias, Andreas Stafylopatis, Georgios Tagaris,
  and Stefanos Kollias.
\newblock Machine learning for neurodegenerative disorder diagnosis—survey of
  practices and launch of benchmark dataset.
\newblock {\em International Journal on Artificial Intelligence Tools},
  27(03):1850011, 2018.

\bibitem{RefWorks:doc:5af421b6e4b02dfcb38d52a0}
Y.~I Tian, T. Kanade, and J.~F. Cohn.
\newblock Recognizing action units for facial expression analysis.
\newblock {\em IEEE Transactions on Pattern Analysis and Machine Intelligence},
  23(2):97--115, Feb 2001.

\bibitem{RefWorks:doc:5b081f42e4b085f70ca5255f}
Michel~F. Valstar and Maja Pantic.
\newblock Induced disgust, happiness and surprise: an addition to the mmi
  facial expression database.

\bibitem{RefWorks:doc:5b0be2a4e4b02f452d8e2352}
9.00 Welcome.
\newblock Programme of the workshop on corpora for research on emotion and
  affect.

\bibitem{RefWorks:doc:5b0811f6e4b009b94b6ca272}
James Williamson, Thomas Quatieri, Brian Helfer, Gregory Ciccarelli, and
  Daryush Mehta.
\newblock Vocal and facial biomarkers of depression based on motor
  incoordination and timing.
\newblock AVEC '14, pages 65--72. ACM, Nov 7, 2014.

\bibitem{RefWorks:doc:5af1a199e4b0b5a912ed9e0a}
Jianwei Yang, Devi Parikh, and Dhruv Batra.
\newblock Joint unsupervised learning of deep representations and image
  clusters.
\newblock pages 5147--5156. IEEE, 2016.

\bibitem{kollias1}
Stefanos Zafeiriou, Dimitrios Kollias, Mihalis~A Nicolaou, Athanasios
  Papaioannou, Guoying Zhao, and Irene Kotsia.
\newblock Aff-wild: Valence and arousal'in-the-wild'challenge.
\newblock In {\em Proceedings of the IEEE Conference on Computer Vision and
  Pattern Recognition Workshops}, pages 34--41, 2017.

\bibitem{RefWorks:doc:5af1aedee4b04597bddec2c7}
Stefanos Zafeiriou, Athanasios Papaioannou, Irene Kotsia, Mihalis Nicolaou, and
  Guoying Zhao.
\newblock Facial affect "in-the-wild": A survey and a new database.
\newblock pages 1487--1498. IEEE, 2016.

\bibitem{RefWorks:doc:5b0e6a23e4b0dc9b736d309b}
Matthew~D. Zeiler and Rob Fergus.
\newblock Visualizing and understanding convolutional networks.
\newblock Nov 12, 2013.

\end{thebibliography}
}


\end{document}